\documentclass{article}

\usepackage{microtype}
\usepackage{graphicx}
\usepackage{subfigure}
\usepackage{booktabs} 

\usepackage[breaklinks]{hyperref}
\usepackage{url}

\usepackage{breakurl}

\usepackage[accepted]{icml2023}

\usepackage{soulutf8}

\usepackage{tikz}
\usetikzlibrary{shapes.geometric, arrows.meta}
\usetikzlibrary{positioning}

\tikzstyle{boxes} = [rectangle, rounded corners, minimum width=4cm, minimum height=1.5cm,text centered, text width = 2.5cm, draw=black, fill=orange!30]
\tikzstyle{arrow} = [thick,->,>=Latex]
\tikzstyle{arrow2} = [thick,<->,>=Latex]

\usepackage{amsmath}
\usepackage{amssymb}
\usepackage{mathtools}
\usepackage{amsthm}

\usepackage[capitalize,noabbrev]{cleveref}

\theoremstyle{plain}
\newtheorem{theorem}{Theorem}[section]

\newtheorem{lemma}[theorem]{Lemma}

\theoremstyle{definition}

\newtheorem{assumption}[theorem]{Assumption}
\theoremstyle{remark}

\usepackage{tikz}
\newcommand*\term[1]{\;  \; \tikz[baseline=(char.base)]{
            \node[shape=circle,draw,inner sep=1pt] (char) {#1};}}
\newcommand*\termw[1]{\tikz[baseline=(char.base)]{
            \node[shape=circle,draw,inner sep=1pt] (char) {#1};}}

\usepackage[textsize=tiny]{todonotes}

\icmltitlerunning{Average Reward Actor-Critic}

\begin{document}

\twocolumn[
\icmltitle{ Off-Policy Average Reward Actor-Critic with \\ Deterministic Policy Search}



\icmlsetsymbol{equal}{*}

\begin{icmlauthorlist}
\icmlauthor{Naman Saxena}{yyy}
\icmlauthor{Subhojyoti Khastagir}{yyy}
\icmlauthor{Shishir Kolathaya}{yyy,xxx}
\icmlauthor{Shalabh Bhatnagar}{yyy}
\end{icmlauthorlist}

\icmlaffiliation{yyy}{Department of Computer Science and Automation, Indian Institute of Science, Bangalore, India}
\icmlaffiliation{xxx}{Robert Bosch Centre for Cyber-Physical Systems, Indian Institute of Science, Bangalore, India}


\icmlcorrespondingauthor{Naman Saxena}{namansaxena@iisc.ac.in}

\icmlkeywords{reinforcement learning, actor critic algorithm, deterministic policy,
off-policy, target network, average reward, finite time analysis,
asymptotic convergence, three time scale stochastic approximation,
DeepMind control suite}

\vskip 0.3in
]



\printAffiliationsAndNotice{}  

\begin{abstract}
The average reward criterion is relatively less studied as most existing works in the Reinforcement Learning literature consider the discounted reward criterion. There are few recent works that present on-policy average reward actor-critic algorithms, but average reward off-policy actor-critic is relatively less explored. In this work, we present both on-policy and off-policy deterministic policy gradient theorems for the average reward performance criterion. Using these theorems, we also present an Average Reward Off-Policy Deep Deterministic Policy Gradient (ARO-DDPG) Algorithm. We first show asymptotic convergence analysis using the ODE-based method. Subsequently, we provide a finite time analysis of the resulting stochastic approximation scheme with linear function approximator and obtain an $\epsilon$-optimal stationary policy with a sample complexity of $\Omega(\epsilon^{-2.5})$. We compare the average reward performance of our proposed ARO-DDPG algorithm and observe better empirical performance compared to state-of-the-art on-policy average reward actor-critic algorithms over MuJoCo-based environments.
\end{abstract}

\section{Introduction}
\label{sc:1}
The reinforcement learning (RL) paradigm has shown significant promise for finding solutions to decision making problems that rely on a reward-based feedback from the environment. 
Here one is mostly concerned with the long-term reward acquired by the algorithm. In the case of infinite horizon problems, the discounted reward criterion has largely been studied because of its simplicity. 
Major recent development in the context of RL in continuous state-action spaces has considered the discounted reward criterion \citep{1,2,3,4}. However, there are very few works which focus on the average reward performance criterion in the continuous state-action  setting \citep{5,6}. 

The average reward criterion has started receiving attention in recent times and there are papers that discuss the benefits of using this criterion over the discounted reward \citep{19,20}. One of the reasons being, average reward criteria only considers recurrent states and it happens to be the most selective optimization criterion in recurrent Markov Decision Processes (MDPs) according to n-discount optimality criterion. Please refer \citet{18} for more details on n-discount optimality criterion. Further, optimization in average reward setting is not dependent on the initial state distribution. Moreover, the discrepancy between the objective function and the evaluation metric, that  exists for discounted reward setting, is resolved by opting for the average reward criterion. We encourage the readers to go through \citet{19,20} for better understanding of the benefits mentioned.

There are very few algorithms in literature that optimize the average reward and all of them happen to be on-policy algorithms \citep{5,6}. It has been demonstrated several times that on-policy algorithms are less sample efficient than off-policy algorithms \cite{3,4,7} for the discounted reward criterion. In this paper we try to find whether the same is true for the average reward criterion. We try to overcome the research gap in development of off-policy average reward algorithms for continuous state and action spaces by proposing an Average Reward Off-Policy Deep Deterministic Policy Gradient (ARO-DDPG) Algorithm.

Average reward algorithms suffers from few challenges. The policy evaluation step in the case of the average reward algorithm is equivalent to finding the solution to the Poisson equation (i.e., the Bellman equation for a given policy). Poisson equation, because of its form, does not admit a unique solution but only solutions that are unique up to a constant term. Further, the policy evaluation step in this case consists of finding not just the Differential Q-value function but also the average reward. Thus, because of the required estimation of two quantities instead of one, the role of the optimization algorithm and the target network increases here. Therefore we implement the proposed ARO-DDPG algorithm by using target network and by carefully selecting the optimization algorithm. 

The following are the broad contributions of our paper:
\begin{itemize}
  \item We provide both on-policy and off-policy deterministic policy gradient theorems for the average reward performance metric.
  \item We present our Average Reward
Off-Policy Deep Deterministic Policy Gradient (ARO-DDPG) algorithm.
\item We show a comparison of our algorithm on several environments with other state-of-the-art average reward algorithms in the literature.
\item We perform asymptotic convergence analysis using ODE-based method and also provide a finite time analysis of our three timescale stochastic approximation based actor-critic algorithm using a linear function approximator.
\end{itemize}

\citet{10,3} and \citet{21} individually address one of the aspects of discounted reward performance criteria for deterministic policies such as policy gradient theorem, implementation of practical algorithm and convergence analysis. In this paper we provide a comprehensive treatment of average reward performance criteria for deterministic policies by covering policy gradient theorem, implementation of practical algorithm and convergence analysis. The rest of the paper is structured as follows: In Section \ref{sc:2}, we present the preliminaries on the MDP framework, the basic setting as well as the policy gradient algorithm. Section \ref{sc:3} presents the deterministic policy gradient theorem and our proposed ARO-DDPG algorithm. Section \ref{sc:4} then presents the main theoretical results related to the convergence analysis. Section \ref{sc:5} presents the experimental results. In Section \ref{sc:6}, we discuss other related work and Section \ref{sc:7} presents the conclusions. The detailed proofs for the convergence analysis are available in the Appendix.

\section{Preliminaries}\label{sc:2}

Consider a Markov Decision Process (MDP) $M = \{S,A,R,P,\pi\}$ where $S \subset \mathbb{R}^{n}$ is the (continuous) state space, $A \subset \mathbb{R}^{m}$ is the (continuous) action space, $R: S \times A \mapsto \mathbb{R}$ denotes the reward function with $R(s,a)$ being the reward obtained under state $s$ and action $a$.  Further, $P(\cdot|s,a)$ denotes the state transition function defined as $P: S \times A \mapsto \mu(\cdot)$, where $\mu: \mathcal{B}(S) \mapsto [0,1]$ is a probability measure. Deterministic policy $\pi$ is defined as $\pi: S \mapsto A$. 
In the above, $\mathcal{B}(S)$ represents the Borel sigma algebra on $S$. Stochastic policy $\pi_r$ is defined as $\pi_r: S \mapsto \mu'(\cdot)$, where $\mu' : \mathcal{B}(A) \mapsto [0,1]$ and $\mathcal{B}(A)$ is the Borel sigma algebra on $A$. 

\begin{assumption}\label{as:1}
The Markov process obtained under any policy $\pi$ is ergodic.
\end{assumption}
Assumption \ref{as:1} is necessary to ensure existence of a unique steady state distribution of the Markov process.

\subsection{Discounted Reward MDPs}\label{sc:2.1}

In discounted reward MDPs, discounting is controlled by $\gamma \in (0,1)$. The following performance metric is optimized with respect to the policy:
\begin{equation}\label{eq:1}
 \eta(\pi) =  \mathbb{E}^{\pi}[ \sum_{t = 0}^{\infty} \gamma^{t}R(s_{t},a_{t})] = \int_{S} \rho_{0}(s) V^{\pi}(s) \,ds. 
\end{equation}
Here, $\rho_{0}$ is the initial state distribution and $V^{\pi}$ is the value function. $V^{\pi}(s)$ denotes the long term discounted reward acquired when starting in the state $s$.
\begin{equation}\label{eq:2}
 V^{\pi}(s_{t}) = \mathbb{E}^{\pi}\Big[ R(s_{t},a_{t}) + \gamma V^{\pi}(s_{t+1})|s_{t}\Big]. 
\end{equation}
\subsection{Average reward MDPs}\label{sc:2.2}

The performance metric in the case of average reward MDPs is the long-run average reward $\rho(\pi)$ defined as follows: 
\begin{equation}\label{eq:3}
 \rho(\pi) = \lim_{N \to \infty} \frac{1}{N} \mathbb{E}^{\pi}\Big[ \sum_{t = 0}^{N-1} R(s_{t},a_{t})\Big] = \int_{S} d^{\pi}(s) R^{\pi}(s) \,ds, 
\end{equation}
where $R^\pi(s) \stackrel{\triangle}{=} R(s,\pi(s))$.
The limit in the first equality in \eqref{eq:3} exists because of Assumption \ref{as:1}. The quantity $d^\pi(s)$ in the second equality in \eqref{eq:3} corresponds to the steady state probability of the Markov process being in state $s\in S$ and it exists and is unique given $\pi$ from Assumption \ref{as:1} as well.

$V_{diff}^{\pi}$ is the differential value function corresponding to the policy $\pi$ and is defined in (\ref{eq:6}). Further, the differential Q-value or action-value function $Q^\pi_{diff}$ is defined in (\ref{eq:7}).
\begin{equation}\label{eq:6}
 V_{diff}^{\pi}(s_{t}) =  \mathbb{E}^{\pi}[ \sum_{i = t}^{\infty} R(s_{i},a_{i}) - \rho(\pi) | s_{t}]. 
\end{equation}
\begin{equation}\label{eq:7}
 Q_{diff}^{\pi}(s_{t}, a_{t}) =  \mathbb{E}^{\pi}[ \sum_{i = t}^{\infty} R(s_{i},a_{i}) - \rho(\pi) | s_{t}, a_{t}]. 
\end{equation}
\begin{lemma}\label{lm:1}
There exists a unique constant $k(=\rho(\pi))$ which satisfies the following equation for differential value function $V_{diff}$ :  
\begin{equation}\label{eq:5}
 V_{diff}^{\pi}(s_{t}) = \mathbb{E}^{\pi}[ R(s_{t},a_{t}) - k +  V_{diff}^{\pi}(s_{t+1})|s_{t}] 
\end{equation}
\end{lemma}

\begin{proof}
See Lemma \ref{lm:a1} in the appendix for the proof.
\end{proof}

\subsection{Policy Gradient Theorem}\label{sc:2.3}

Unlike in Q-learning where we try to find the optimal Q-value function and then infer the policy from it, the policy gradient theorem \citep{9,10,11} allows us to directly optimize the performance metric via its gradient with respect to the policy parameters. Q-learning can be visualized to be a value iteration scheme while an algorithm based on the policy gradient theorem can be seen as mimicking policy iteration. \citet{9} provided the policy gradient theorem for on-policy optimization of both the discounted reward and the average reward algorithms, see (\ref{eq:8})-(\ref{eq:9}), respectively.
\begin{equation}\label{eq:8}
\nabla_{\theta} \eta(\pi) = \int_{S}\omega^{\pi}(s)\int_{A}\nabla_{\theta}\pi_r(a|s,\theta)Q^{\pi_{r}}(s,a) \,da \,ds.
\end{equation}
\begin{equation}\label{eq:9}
\nabla_{\theta} \rho(\pi) = \int_{S}d^{\pi}(s)\int_{A}\nabla_{\theta}\pi_r(a|s,\theta)Q_{diff}^{\pi_{r}}(s,a) \,da \,ds.
\end{equation}
In (\ref{eq:8}) 
$\omega^{\pi}$ denotes the long term discounted state visitation probability density which is defined in \eqref{eq:10} while $d^{\pi}(s)=\lim_{t \to \infty} P^{\pi}_{t}(s)$ is the steady state probability density on states. 
$P^{\pi}$ denotes the transition probability kernel for the Markov chain induced by policy $\pi$ and $P^\pi_t$ is the state distribution at instant $t$ given by (\ref{eq:12}). 
\begin{equation}\label{eq:10}
\omega^{\pi}(s) = (1 - \gamma) \sum_{t = 0}^{\infty} \gamma^{t}P^{\pi}_{t}(s).
\end{equation}
\begin{equation}\label{eq:12}
P^{\pi}_{t}(s) = \int_{S \times S \ldots }\rho_{0}(s_0)\prod_{k=0}^{t-1}P^{\pi}(s_{k+1}|s_{k}) \,ds_{0} \ldots \,ds_{t-1}.
\end{equation}
The policy gradient theorem in 
\citet{9} is only valid for on-policy algorithms. \citet{11} proposed an approximate off-policy policy gradient theorem for stochastic policies, see (\ref{eq:13}), where $d^{\mu}$ stands for the steady state density function corresponding to the policy $\mu$. 
\begin{equation}\label{eq:13}
\nabla_{\theta} \eta(\pi) \approx \int_{S}d^{\mu}(s)\int_{A}\nabla_{\theta}\pi_r(a|s,\theta)Q^{\pi}(s,a) \,da \,ds.
\end{equation}
\citet{10} came up with the deterministic policy gradient theorem for discounted reward setting, see (\ref{eq:14}), which eventually led to the development of very successful Deep Deterministic Policy Gradient (DDPG) \citep{3} algorithm and Twin Delayed DDPG (TD3) algorithm \citep{7}. In the next section we show how we extend the same development for average reward criterion.
\begin{equation}\label{eq:14}
\nabla_{\theta} \eta(\pi) = \int_{S}\omega^{\pi}(s)\nabla_{a}Q^{\pi}(s,a)|_{a =\pi(s)}\nabla_{\theta}\pi(s,\theta) \,ds.
\end{equation}

\section{Proposed Average Reward Algorithm}\label{sc:3}

We now propose the deterministic policy gradient theorem for the average reward criterion. The policy gradient estimator has to be derived separately for both the on-policy and off-policy settings. Obtaining the on-policy deterministic policy gradient estimator is straight forward but dealing with the off-policy gradient estimates involves an approximate gradient \citep{11}.

\subsection{On-Policy Policy Gradient Theorem}\label{sc:3.1}

We cannot directly use the second equality of (\ref{eq:3}) to derive the policy gradient theorem because of the inability to take the derivative of steady state density function. Therefore one needs to use Lemma \ref{lm:1} to obtain the average reward deterministic policy gradient theorem.

\begin{theorem}\label{th:1}
 The gradient of $\rho(\pi)$ with respect to policy parameter $\theta$ is given as follows:
\begin{equation}\label{eq:15}
\nabla_{\theta} \rho(\pi) = \int_{S}d^{\pi}(s)\nabla_{a}Q_{diff}^{\pi}(s,a)|_{a =\pi(s)}\nabla_{\theta}\pi(s,\theta) \,ds.
\end{equation}
\end{theorem}
\begin{proof}
See Theorem \ref{th:a1} in the appendix for the proof.
\end{proof}

\subsection{Compatible Function Approximation}\label{sc:3.2}

The result in this section is mostly inspired from \citet{10}. 
Recall that $Q^\pi_{diff}(s,a)$ is the `true' differential $Q$-value of the state-action tuple $(s,a)$ under the parameterized policy $\pi$. Now let $Q^w_{diff}(s,a)$ denote the approximate differential $Q$-value of the $(s,a)$-tuple when function approximation with parameter $w$ is used. Lemma~\ref{lm:2} says that when the function approximator satisfies a compatibility condition (cf.~(\ref{eq:16},\ref{eq:16.2})), then the gradient expression in (\ref{eq:15}) is also satisfied by $Q^w_{diff}$ in place of $Q^\pi_{diff}$.

\begin{lemma}\label{lm:2}
For on-policy case, assume that the differential Q-value function (\ref{eq:7}) satisfies the following:
\begin{enumerate}
\item 
\begin{equation}\label{eq:16}
\nabla_{w}\nabla_{a} Q^{w}_{diff}(s,a)|_{a=\pi(s)} = \nabla_{\theta} \pi(s,\theta).
\end{equation}
\item
Differential Q-value function parameter $w = w_\epsilon^{*}$ optimizes the following error function:
\begin{equation}\label{eq:16.2}
\begin{split}
\zeta(\theta,w) =& \frac{1}{2}\int_{S}d^{\pi}(s)\|\nabla_{a}Q_{diff}^{\pi}(s,a)|_{a =\pi(s)}\\
&-\nabla_{a} Q_{diff}^{w}(s,a)|_{a =\pi(s)}\|^2 \,ds.      
\end{split}
\end{equation}
\end{enumerate}
Then,
\begin{equation}\label{eq:17}
\begin{split}
&\int_{S}d^{\pi}(s)\nabla_{a}Q_{diff}^{\pi}(s,a)|_{a =\pi(s)}\nabla_{\theta}\pi(s,\theta) \,ds\\ 
&= \int_{S}d^{\pi}(s)\nabla_{a}Q_{diff}^{w}(s,a)|_{a =\pi(s)}\nabla_{\theta}\pi(s,\theta) \,ds.
\end{split}
\end{equation}
Further, in the case when a linear function approximator is used, we obtain
\begin{equation}
\nabla_{a} Q^{w}_{diff}(s,a)|_{a=\pi(s)} = \nabla_{\theta} \pi(s,\theta)^{\intercal}w. 
\end{equation}
\end{lemma}

\begin{proof}
See Lemma \ref{lm:a2} in the appendix for a proof.
\end{proof}

An important implication of \cref{lm:2} also is that the dimension of the matrix on the left hand side and the right hand side of (\ref{eq:16}) should be the same. Hence the dimensions of the parameters $\theta$ (used in the parameterized policy) and $w$ (used to approximate the differential Q-value function) are the same. Lemma \ref{lm:2} shows that the compatible function approximation theorem has the same form in the average reward setting as the discounted reward setting.

\subsection{Off-Policy Policy gradient theorem}\label{sc:3.3}

In order to derive off-policy policy gradient theorem it is not possible to use the direction adopted by \citet{11} for off-policy stochastic policy gradient theorem for the discounted reward setting. We first mention our proposed approximate off-policy deterministic policy gradient theorem and then explain why some alternatives would not have worked.

\begin{assumption}\label{as:2}
For the Markov chain obtained from the policy $\pi$, let $K(\cdot | \cdot)$ be the transition kernel and $S^{\pi}$ the steady state measure. Then there exists $a > 0$ and $\kappa \in(0,1)$ such that 
$$ D_{TV}(K^{t}( \cdot | s), S^{\pi}(\cdot)) \leq a\kappa^{t} , \forall t, \forall s \in S.$$
\end{assumption}

Assumption \ref{as:2}  states that Markov chain generated by a policy $\pi$ follows uniform ergodicity property. This assumption is necessary to get an upper bound on the total variation norm of steady state probability distribution of two policies. Further this assumption allows for fast mixing of markov chain and i.i.d sampling of transitions from buffer for convergence analysis purpose.

\begin{theorem}\label{th:2}
The approximate gradient ($\widehat{\nabla_{\theta} \rho}(\pi)$) of the average reward $\rho(\pi)$ with respect to the policy parameter $\theta$ is given by the following expression:
\begin{equation}\label{eq:18}
\begin{split}
\widehat{\nabla_{\theta} \rho}(\pi) =
 \int_{S}d^{\mu}(s)\nabla_{a}Q_{diff}^{\pi}(s,a)|_{a=\pi(s)}\nabla_{\theta}\pi(s,\theta) \,ds .
\end{split}
\end{equation}
Further, the approximation error is
 $\mathcal{E}(\pi,\mu) = \| \nabla_{\theta} \rho(\pi) - \widehat{\nabla_{\theta} \rho}(\pi) \|,$
where $\mu$ represents the behaviour policy with parameter $\theta^{\mu}$ and $\nabla_{\theta} \rho(\pi)$ is the on-policy policy gradient from Theorem \ref{th:1}. $\mathcal{E}$ satisfies
\begin{equation}
\mathcal{E}(\pi,\mu)     \leq Z \| \theta - \theta^{\mu} \|,
\end{equation}
where, $Z = 2^{n+1}C(\lceil{\log_{\kappa}a^{-1}\rceil} + 1\slash\kappa)L_{t}$ with $L_{t}$ being the Lipchitz constant for the transition probability density function (Assumption \ref{as:9}). Constants $a$ and $\kappa$ are from Assumption \ref{as:2}, $n$ is the dimension of the state space, and $C = \max_s \|\nabla_{a}Q_{diff}^{\pi}(s,a)|_{a=\pi(s)}\nabla_{\theta}\pi(s,\theta)\|.$
\end{theorem}
\begin{proof}
See Theorem \ref{th:a2} in the appendix for a proof.
\end{proof}

Theorem \ref{th:2} suggests that the approximation error in the gradient increases as the difference between the target policy $\pi$ and the behaviour policy $\mu$ increases. 

\subsection{Off-Policy Alternatives}\label{sc:3.4}

In this section we will talk about what alternatives could be thought of in place of what is suggested in \cref{sc:3.3} and why those alternatives would not work.

\begin{enumerate}
    \item One can possibly take inspiration from \citet{11} and define an objective function, $\rho_{new}(\pi)$, as in (\ref{eq:19}), which is a naive off-policy version of (\ref{eq:3}).
    \begin{equation}\label{eq:19}
       \rho_{new}(\pi) = \int_{S} d^{\mu}(s) R^{\pi}(s) \,ds.
    \end{equation}
    
    If, however, we take the derivative of  $\rho_{new}(\pi)$ defined above, we get the policy update rule as in (\ref{eq:20}).
    \begin{equation}\label{eq:20}
    \nabla_{\theta} \rho_{new}(\pi) = \int_{S}d^{\mu}(s)\nabla_{a}R(s,a)|_{a =\pi(s)}\nabla_{\theta}\pi(s,\theta) \,ds.
    \end{equation}
   
   The update rule (\ref{eq:20}) only considers the reward function and  not the transition dynamics of the MDP. In (\ref{eq:18}), the derivative of the objective function includes the differential Q-value function which encapsulates both the information of the reward function and the transition dynamics of the MDP and hence is valid derivative. Therefore we cannot use $\rho_{new}$, given in \eqref{eq:19}.
   
   \item A lot of work in the off-policy setting relies on importance sampling ratios. Recently a few works devised a method to estimate the steady state probability density ratio of the target and behavior policies  \citep{12,13,14,15}. The ratio of steady state densities could be used for deterministic policy optimization but there are certain issues which prohibit its usage, see  (\ref{eq:21}).
   \begin{equation}\label{eq:21}
    \begin{split}
    &\nabla_{\theta} \rho(\pi) \\ 
    &=  \int_{S}d^{\mu}(s)\tau(s)\nabla_{a}Q_{diff}^{\pi}(s,a)|_{a =\pi(s)}\nabla_{\theta}\pi(s,\theta) \,ds.
    \end{split}
    \end{equation}
    Here, $\tau(s)$ is the steady state probability density ratio defined as $d^{\pi}(s)/d^{\mu}(s)$. 
    In order to calculate $\tau(s)$ we need information about ($\pi(a|s),\mu(a|s)$ and $P(s'|s,a)$). 
     We need the ratio $\pi(a|s)/\mu(a|s)$ and for deterministic policies the ratio would be $\delta(a - \pi(s)/\delta(a - \mu(s))$, where $\delta(\cdot)$ is the Dirac-Delta function:
    \begin{equation}\label{eq:22}
    \frac{\delta(a - \pi(s))}{\delta(a - \mu(s))} = \begin{cases}
    0 & \mbox{ if } a = \mu(s) \\
    \infty & \mbox{ if } a = \pi(s) \\
    \frac{0}{0} & \mbox{ otherwise.}
    
    \end{cases}
    \end{equation}
    From (\ref{eq:22}), it is clear that the ratio $\delta(a - \pi(s)/\delta(a - \mu(s))$ will be undefined for almost all actions $a\in A$. Thus, we cannot use this ratio for deterministic policies. Otherwise, we need $P(s'|s,\pi(a))$ and $P(s'|s,\mu(a))$. It is possible to get the information about $P(s'|s,\mu(a))$ by sampling from the Markov process generated by the policy $\mu$ but obtaining this information about $P(s'|s,\pi(a))$ is impossible as in the off-policy setting data from $\pi$ is assumed to be simply unavailable.
    \end{enumerate}
In the next section we will use the policy gradient theorems defined in this section to implement practical actor-critic algorithm.

\subsection{Actor-Critic Update rule}
In our paper actor refers to policy and critic refers to the approximate differential Q-value function and average reward estimate combined.
\begin{assumption}\label{as:2.1}
$\alpha_t, \beta_t,$ and $\gamma_t$ are the step sizes for critic, target critic parameter, and actor parameter updates respectively.
\begin{equation*}
\alpha_t = \frac{C_\alpha}{(1 + t)^\sigma} \quad \beta_t = \frac{C_\beta}{(1 + t)^u} \quad \gamma_t = \frac{C_\gamma}{(1 + t)^v}    
\end{equation*}

Here, $C_\alpha, C_\beta, C_\gamma > 0$ and $0 < \sigma < u < v < 1$. $\alpha_t$ is at the fastest timescale, $\beta_t$ is at slower timescale and $\gamma_t$ is at the slowest timescale. 
\end{assumption}
Please note that target critic parameter refers to copy of main critic parameter that are updated using polyak averaging. The critic parameters are estimated using the TD(0) update rule with target critic parameters. We are using target critic parameters to ensure stability of the iterates of the algorithm. Let $\{s_{i},a_{i},s_{i}'\}_{i=0}^{n-1}$ denote the batch of sampled data from the replay buffer.
\begin{equation}\label{eq:23}
\begin{split}
 \xi_{t}^{j} =& \frac{1}{2} \sum_{i=0}^{n-1}\biggl(R(s_{i},a_{i}) - \overline{\rho_{t}}  - Q_{diff}^{w^{j}}(s_{i}, a_{i})\\ 
 &+ \min(Q_{diff}^{\overline{w^{1}}},Q_{diff}^{\overline{w^{2}}})(s_{i}',\pi(s_{i}',\overline{\theta_{t}})) \biggl)^{2}
 j \in \{1,2\}
\end{split}
\end{equation}
\begin{equation}\label{eq:24}
\begin{split}
\xi_{t}^{3} =& \frac{1}{2} \sum_{i=0}^{n-1}\biggl(R(s_{i},a_{i}) - \rho_{t} - 
\min(Q_{diff}^{\overline{w^{1}}},Q_{diff}^{\overline{w^{2}}})(s_{i},a_{i})\\  
&+\min(Q_{diff}^{\overline{w^{1}}},Q_{diff}^{\overline{w^{2}}})(s_{i}',\pi(s_{i}',\overline{\theta_{t}})) \biggl)^{2}
\end{split}
\end{equation}
Equations (24) and (25) correspond to the Bellman error for the differential Q-value function approximator and the average reward estimator respectively. Note that we are using the double Q-value function approximator. Here $\bar{\rho}_t$ represents the target estimator for average reward at time t, $Q_{diff}^{\overline{w^i}}$ represents the differential Q-value function parameterized by target differential Q-value parameter $\bar{w}_t^i$ and $\bar{\theta}_t$ represents the target parameter for actor at time t, respectively.
\begin{equation}\label{eq:25}
    w_{t+1}^{j} = w_{t}^{j} - \alpha_{t}\nabla_{w^{j}}\xi_{t}^{j} \>\>\>\> j \in \{1,2\} 
\end{equation}
\begin{equation}\label{eq:26}
    \rho_{t+1} = \rho_{t} -\alpha_{t}\nabla_{p}\xi_{t}^{3} 
\end{equation}
Our aim is to find the value of parameters for differential Q-value function and average reward estimator such that the Bellman equation is satisfied. Hence, the Bellman error in (\ref{eq:23}) is used to update the differential Q-value function parameters $w_{t}^{j}$ using (\ref{eq:25}) and the Bellman error in (\ref{eq:24}) is used to update the estimator of the average reward $\rho_{t}$ using (\ref{eq:26}). Our approach is motivated from the update rule for the  differential Q-value function and the average reward parameters given in \citet{25} (equations (3) and (4)) and \citet{28}(Algorithm 2).
\begin{equation}\label{eq:27}
    \nu_{i} = \nabla_{a}min(Q_{diff}^{w^{1}},Q_{diff}^{w^{2}})(s_{i},a)|_{a =\pi(s_{i})}\nabla_{\theta}\pi(s_{i},\theta_{t})
\end{equation}
\begin{equation}\label{eq:28}
    \theta_{t+1} = \theta_{t} + \gamma_{t}\biggl(\sum_{i=0}^{n-1}\nu_{i}\biggl)
\end{equation}
Actor update is performed using Theorem \ref{th:2}. Actor parameter, $\theta_t$, is updated using empirical estimate (\ref{eq:27}) of the gradient in \eqref{eq:18}. 
\begin{equation}\label{eq:29}
    \overline{w_{t+1}^{j}} = \overline{w_{t}^{j}} + \beta_{t}(w_{t+1}^{j} - \overline{w_{t}^{j}}) \>\>\>\> j \in \{1,2\} 
\end{equation}
\begin{equation}\label{eq:30}
    \overline{\rho_{t+1}} = \overline{\rho_{t}} +\beta_{t}(\rho_{t+1} - \overline{\rho_{t}} ) 
\end{equation}
\begin{equation}\label{eq:31}
    \overline{\theta_{t+1}} = \overline{\theta_{t}} + \beta_{t}(\theta_{t+1} - \overline{\theta_{t}} )
\end{equation}
Equation (\ref{eq:29}) - (\ref{eq:31}) are used to update the target Q-value function parameter $\overline{w_{t}^{j}}$, target average reward estimator $\overline{\rho_{t}}$ and target actor parameter $\overline{\theta_{t}}$.

\section{Convergence Analysis}\label{sc:4}
In this section we present the asymptotic convergence analysis and finite time analysis of the on-policy and off-policy average reward actor critic algorithm with linear function approximators. First we mention the assumptions taken to perform the convergence analysis followed by the main results. 

\begin{assumption}\label{as:3}
$\phi^{\pi}(s) \big(=\phi(s,\pi(s)\big)$ denotes the feature vector of state s and satisfies $\|\phi^{\pi}(s)\| \leq 1 $.
\end{assumption}
The assumption above is just taken for the sake of convenience. 
\begin{assumption}\label{as:4}
The reward function is uniformly bounded, viz., $|R^{\pi}(s)| \leq C_{r}<\infty$.
\end{assumption}
Assumption \ref{as:4} is required to make sure that the average reward objective function is bounded from above.
\begin{assumption}\label{as:5}
$Q_{diff}^w(s,a)$ is Lipchitz continuous w.r.t to $a$. Thus, $\forall w \quad \|Q_{diff}^w(s,a_1) - Q_{diff}^w(s,a_2)\| \leq L_a \|a_1 - a_2\|$.
\end{assumption}
Continuity of approximate Q-value function w.r.t action is enforced using Assumption \ref{as:5}. Without the continuity property, the approximate differential Q-values will not generalize for unseen action values.
\begin{assumption}\label{as:6}
Parameterised policy $\pi(s,\theta)$ is Lipchitz continuous w.r.t $\theta$. Thus, $\|\pi(s,\theta_1) - \pi(s,\theta_2)\| \leq L_{\pi}\|\theta_1 - \theta_2\|$.   
\end{assumption}
Assumption \ref{as:6} is a common regularity assumption for convergence of actor. It can be found in \citet{29}, \citet{21} and \citet{30}.
\begin{assumption}\label{as:8}
The state feature mapping ($\phi^\pi(s)=\phi(s,\pi(s)$) defined for a policy $\pi$ with parameter $\theta$ is Lipschitz continuous w.r.t $\theta$. Thus, $\max_s\|\phi^{\pi_1}(s)-\phi^{\pi_2}(s)\| \leq L_\phi\|\theta_1 - \theta_2\|$.
\end{assumption}

Continuity of state action feature w.r.t action is required to ensure generalisation of Q-values to unseen action values. Using this continuity of state action feature with Assumption \ref{as:6} we can satisfy Assumption \ref{as:8}.

\subsection{Asymptotic Convergence}
We prove the asymptotic convergence of the three timescale stochastic approximation on-policy algorithm (Algorithm \ref{alg:4}) using ODE-based method \citep{36,38, 35} in two steps. First we keep the policy parameter $\theta$ fixed and prove the convergence of differential Q-value function parameter $w_t$, average reward estimator $\rho_t$, target differential Q-value function parameter $\bar{w}_t$ and target average reward estimator $\bar{\rho}_t$ in Theorem \ref{th:6} (given below). Later we prove the convergence of policy parameter $\theta_t$ using the point of convergence of critic parameters because the policy parameter are updated at the slowest timescale $\gamma_t$.

\begin{theorem}\label{th:6}
In Algorithm \ref{alg:4}, let policy parameter $\theta_t$ be kept constant at $\theta$. The differential Q-value function parameter $w_t$ and the target differential Q-value function parameter $\bar{w}_t$ converges to $w(\theta)^*$. Also, average reward estimator $\rho_t$ and target average reward estimator $\bar{\rho}_t$ converges to $\rho(\theta)^*$. (Note: The point of convergence $w(\theta)^*$ and $\rho(\theta)^*$ are defined in Theorem \ref{th:a5}.) 
\end{theorem}
\begin{proof}
See Theorem \ref{th:a5} in the appendix for the proof.
\end{proof}
Theorem \ref{th:6} uses the two timescale stochastic approximation stability result from \citet{35}. Later, taking inspiration from \citet{cmdp2} and invoking the Theorem 5.3.1 of \citet{38} we prove the convergence of policy parameter $\theta_t$ in Theorem \ref{th:5}.

\begin{theorem}\label{th:5}
$\Gamma_{C_\theta}: \mathbb{R}^d \to C_\theta$ is a projection operator, where $C_\theta$ is compact convex set and $\hat{\Gamma}_{C_\theta}(\theta)\nabla_{\theta}\rho(\theta)$ refers to directional derivative of $\Gamma_{C_\theta}(\cdot)$ in the direction $\nabla_{\theta}\rho(\theta)$ at $\theta$. Let $K = \{\theta \in C_\theta | \hat{\Gamma}_{C_\theta}(\theta)\nabla_{\theta}\rho(\theta) = 0 \}$ and $K^{\epsilon} = \{\theta' \in C_\theta | \exists\; \theta \in K \; \|\theta'-\theta\|<\epsilon \}$. $\forall \epsilon > 0 \;\exists \delta$ such that if $\sup_{\pi}\|e^{\pi}\| < \delta$ then $\theta_t$ converges to $K^{\epsilon}$ as $t \to \infty$ with probability one. $e^{\pi}$ is the function approximation error defined in Lemma \ref{lm:a20}.
\end{theorem}
\begin{proof}
See Theorem \ref{th:a6} in the appendix for the proof.
\end{proof}
Theorem \ref{th:5} essentially argues that the actor update scheme in Algorithm \ref{alg:4} tracks the ODE $\dot{\theta}(t) = \hat{\Gamma}_{C_\theta}(\theta(t))(\nabla_{\theta}\rho(\theta(t)) + e^{\pi(t)})$ and converges to an $\epsilon-$neighbourhood of the set $K$. Moreover, when $\sup_{\pi}\|e^{\pi}\| \to 0$, the actor update scheme tracks $\dot{\theta}(t) = \hat{\Gamma}_{C_\theta}(\theta(t))(\nabla_{\theta}\rho(\theta(t)))$ and convergence to the set $K$.

Conclusions of Theorem \ref{th:5} will continue to hold for off policy algorithm (Algorithm \ref{alg:5}) by suitably setting the value of l2-regularisation coefficient.

\subsection{Finite Time Analysis}
We perform finite time analysis by finding an upper bound on the expected squared norm of the policy gradient ($\min_{0\leq t\leq T} \mathbb{E}\|\nabla_{\theta}\rho(\theta_t)\|^2$) for both Algorithms \ref{alg:2} and \ref{alg:3}. We first identify error in the parameters of the algorithm and define the dependency graph of error, as shown in Figure \ref{fig:0} for Algorithm \ref{alg:2}. In Figure \ref{fig:0}, an arrow from one error (source) to the other error (destination) shows that an upper bound on the destination error depends on an upper bound on the source error. Exploiting the dependency graph of errors we finally find an upper bound on the expected squared norm of policy gradient ($\min_{0\leq t\leq T} \mathbb{E}\|\nabla_{\theta}\rho(\theta_t)\|^2$) in terms of time $T$.     

\begin{figure}[h]
\centering
\scalebox{0.8}
{
\begin{tikzpicture}[node distance=1cm]
\node (actor) [boxes] {Actor Error $$\frac{1}{T}\sum^{T-1}_{t=0} E \|\nabla_{\theta}\rho(\theta_t)\|^2$$ (Lemma \ref{lm:a3})};
\node (value) [boxes, below = of actor, xshift = -3.0cm] {Differential Q-value Function Error $$\frac{1}{T}\sum_{t=0}^{T-1}\mathbb{E}||\Delta w_t||^2$$ (Lemma \ref{lm:a4})};
\node (tvalue) [boxes, below = of value] {Target Differential Q-value Function Error $$\frac{1}{T}\sum_{t=0}^{T-1}\mathbb{E}\|\Delta\bar{w}_t\|^2$$ (Lemma \ref{lm:a4.8})};
\node (trho) [boxes, below = of actor, xshift = 3.0cm, yshift = -0.2cm] {Target Average Reward Error $$\frac{1}{T}\sum_{t=0}^{T-1}\mathbb{E}|\Delta\bar{\rho}_t|^2$$ (Lemma \ref{lm:a4.9}) };
\node (rho) [boxes, below = of trho, yshift = -0.4cm] {Average Reward Error $$\frac{1}{T}\sum_{t=0}^{T-1}\mathbb{E}|\Delta\rho_t|^2$$ (Lemma \ref{lm:a5})};

\draw [arrow] (value) |- (actor);
\draw [arrow2] (value) -- (tvalue);
\draw [arrow] (trho) -- (value);
\draw [arrow] (tvalue) -- (rho);
\draw [arrow] (rho) -- (trho);
\end{tikzpicture}
}
\caption{Dependency of errors in different types of parameters in Algorithm \ref{alg:2} on one another.}
\label{fig:0}
\end{figure}

\subsubsection{On-Policy Analysis}
In Algorithm \ref{alg:2}, we define the error for policy parameter as the expected squared norm of policy gradient ($\frac{1}{T}\sum^{T-1}_{t=0} E \|\nabla_{\theta}\rho(\theta_t)\|^2$). The error for differential Q-value function parameter $w_t$ and target differential Q-value function parameter $\bar{w}_t$ is defined as $\frac{1}{T}\sum_{t=0}^{T-1}\mathbb{E}||\Delta w_t||^2$ and $\frac{1}{T}\sum_{t=0}^{T-1}\mathbb{E}\|\Delta\bar{w}_t\|^2$ respectively. Here, $\Delta w_t = w_t - w_t^{*}$, $\Delta \bar{w}_t = \bar{w}_t - w_t^{*}$ and $w_t^{*}$ is the optimal differential Q-value function parameter for policy parameter $\theta_t$. The error for target differential Q-value function is defined by taking inspiration from Theorem \ref{th:6}, as Theorem \ref{th:6} says both $w_t$ and $\bar{w}_t$ converge to the same point. The error for average reward estimator $\rho_t$ and target average reward estimator $\bar{\rho}_t$ is defined as $\frac{1}{T}\sum_{t=0}^{T-1}\mathbb{E}|\Delta\rho_t|^2$ and $\frac{1}{T}\sum_{t=0}^{T-1}\mathbb{E}|\Delta\bar{\rho}_t|^2$ respectively. Here, $\Delta \rho_t = \rho_t - \rho_t^{*}$, $\Delta \bar{\rho}_t = \bar{\rho}_t - \rho_t^{*}$ and $\rho_t^{*}$ is the optimal average reward estimate for policy parameter $\theta_t$. Using all the aforementioned errors in the parameter we define a dependency graph in Figure \ref{fig:0} and obtain an upper bound on the expected squared norm of policy gradient ($\min_{0\leq t\leq T} \mathbb{E}\|\nabla_{\theta}\rho(\theta_t)\|^2$) in Theorem \ref{th:3}. 

\begin{theorem}\label{th:3}
The on-policy average reward actor critic algorithm (Algorithm \ref{alg:2}) obtains an $\epsilon$-accurate optimal point with sample complexity of $\Omega(\epsilon^{-2.5})$. We obtain
\begin{equation*}
\begin{split}
\min_{0\leq t \leq T-1} &\mathbb{E}||\nabla_\theta \rho(\theta_t)||^2 \\
& = \mathcal{O}\bigg(\frac{1}{T^{2/5}}\bigg) + 3C_{\pi}^2(C_{a\phi}^2\tau^2 + \frac{4}{M}C_{\pi}^2C_{w_{\epsilon}}^2),  \\   
&\leq \epsilon + \mathcal{O}(1). 
\end{split}
\end{equation*}
Here,$\|\nabla_{\theta} \pi(s)\| \leq C_{\pi}$ (Assumption \ref{as:6}), $\tau = \max_{t} \|w_{t}^{*} - w_{\epsilon,t}^{*}\|$, $w_{\epsilon}^{*}$ is the optimal differential Q-value function parameter according to Lemma \ref{lm:2}. Constant $C_{w_{\epsilon}^{*}}$ is defined in Lemma \ref{lm:a13}. M is the size of batch of samples used to update parameters. $C_{a\phi}$ is the Lipchitz constant defined in Assumption \ref{as:17}. 
\end{theorem}

\begin{proof} 
See Theorem \ref{th:a3} in the appendix for the proof.
\end{proof}

We started the analysis with a three timescale stochastic approximation algorithm but later observed that the best sample complexity is achieved when critic parameter and target critic parameters are updated on the same time-scale, i.e. $u=\sigma$ (Assumption \ref{as:2.1}). The extra terms $3C_{\pi}^2C_{a\phi}^2\tau^2$ and $12C_{\pi}^4C^2_{w_\epsilon^*}/M$ exist in the bound established in Theorem \ref{th:3} because of function approximation error and empirical expectation respectively. $3C_{\pi}^2C_{a\phi}^2\tau^2$ can be reduced using high capacity function approximator such as neural network. $12C_{\pi}^4C^2_{w_\epsilon^*}/M$ can be made small by increasing the size of the batch $M$ used for empirical expectation. The same error terms are also present in the finite time analysis by \citeauthor{21}. Let $K_2 = \{\theta  \;|\; \nabla_{\theta}\rho(\theta) = 0 \}$ and $K_2^{\epsilon} = \{\theta'  | \exists\; \theta \in K_2 \; \|\theta'-\theta\|<\epsilon \}$. $\forall \epsilon > 0 \;\exists \delta$ such that if $|3C_{\pi}^2(C_{a\phi}^2\tau^2 + \frac{4}{M}C_{\pi}^2C_{w_{\epsilon}}^2)| < \delta$ then $\theta_t$ converges to $K_2^{\epsilon}$ with rate $\mathcal{O}(T^{-2/5})$.    

\subsubsection{Off-Policy Analysis}
In Algorithm \ref{alg:3}, we define the error for policy parameter as the expected squared norm of approximate policy gradient ($\frac{1}{T}\sum^{T-1}_{t=0} E \|\widehat{\nabla_{\theta}\rho}(\theta_t)\|^2$). Error in rest of the parameters is defined in the same way as the on-policy case and a similar dependency graph of errors will be obtained. Using the dependency graph of errors in parameters an upper bound in terms of time $T$ on the expected squared norm of approximate policy gradient ($\min_{0\leq t\leq T-1}E \|\widehat{\nabla_{\theta}\rho}(\theta_t)\|^2$) is obtained in Theorem \ref{th:4}. 
\begin{theorem}\label{th:4}
The off-policy average reward actor critic algorithm (Algorithm \ref{alg:3}) with behavior policy $\mu$ obtains an $\epsilon$-accurate optimal point with sample complexity of $\Omega(\epsilon^{-2.5})$. Here $\theta_\mu$ refers to the behavior policy parameter and $\theta_t$ refers to the target or current policy parameter. We obtain
\begin{equation*}
\begin{split}
&\min_{0\leq t \leq T-1} \mathbb{E}\|\widehat{\nabla_\theta \rho}(\theta_t)\|^2 \\
& = \mathcal{O}\biggl(\frac{1}{T^{2/5}}\biggr) + 3C_{\pi}^2(C_{a\phi}^2\tau^2 + \frac{4}{M}C_{\pi}^2C_{w_{\epsilon}}^2) + \mathcal{O}(W_\theta^2) \\
& \leq \epsilon + 3C_{\pi}^2(C_{a\phi}^2\tau^2 + \frac{4}{M}C_{\pi}^2C_{w_{\epsilon}}^2) + \mathcal{O}(W_\theta^2) \\
&\mbox{where } W_\theta  := \sup_t \|\theta^\mu - \theta_t \|.
\end{split}
\end{equation*}
Here,$\|\nabla_{\theta} \pi(s)\| \leq C_{\pi}$ (Assumption \ref{as:6}), $\tau = \max_{t} \|w_{t}^{*} - w_{\epsilon,t}^{*}\|$, $w_{\epsilon}^{*}$ is the optimal differential Q-value function parameter according to Lemma \ref{lm:a2.1}. Constant $C_{w_{\epsilon}^{*}}$ is defined in Lemma \ref{lm:a13}. $C_{a\phi}$ is Lipchitz constant defined in Assumption \ref{as:17}. M is the size of batch of samples used to update parameters.
\end{theorem}
\begin{proof}
See Theorem \ref{th:a4} the appendix for a proof.
\end{proof}

Here also we found that two timescale stochastic approximation algorithm has better sample complexity than three timescale version. We have the same extra term in the bound as established in Theorem \ref{th:3} with an additional term of $\mathcal{O}(W_\theta^2)$. The extra term $\mathcal{O}(W_\theta^2)$ denotes the error induced because of not using the samples from the current policy for performing updates. $W_\theta^2$ will be small when replay buffer is used because replay buffer contains data from policies similar to the current policy. This explains why policy gradient theorem in Theorem \ref{th:2} can be used with replay buffer. Let $K_3 = \{\theta  \;|\; \widehat{\nabla_{\theta}\rho}(\theta) = 0 \}$ and $K_3^{\epsilon} = \{\theta'  | \exists\; \theta \in K_3 \; \|\theta'-\theta\|<\epsilon \}$. $\forall \epsilon > 0 \;\exists \delta$ such that if $|3C_{\pi}^2(C_{a\phi}^2\tau^2 + \frac{4}{M}C_{\pi}^2C_{w_{\epsilon}}^2) + C'W_\theta^2| < \delta$ with $C'>0$ then $\theta_t$ converges to $K_3^{\epsilon}$ with rate $\mathcal{O}(T^{-2/5})$.

\section{Experimental Results}\label{sc:5}
\begin{figure*}[t]
    \centering
    \includegraphics[scale=0.45]{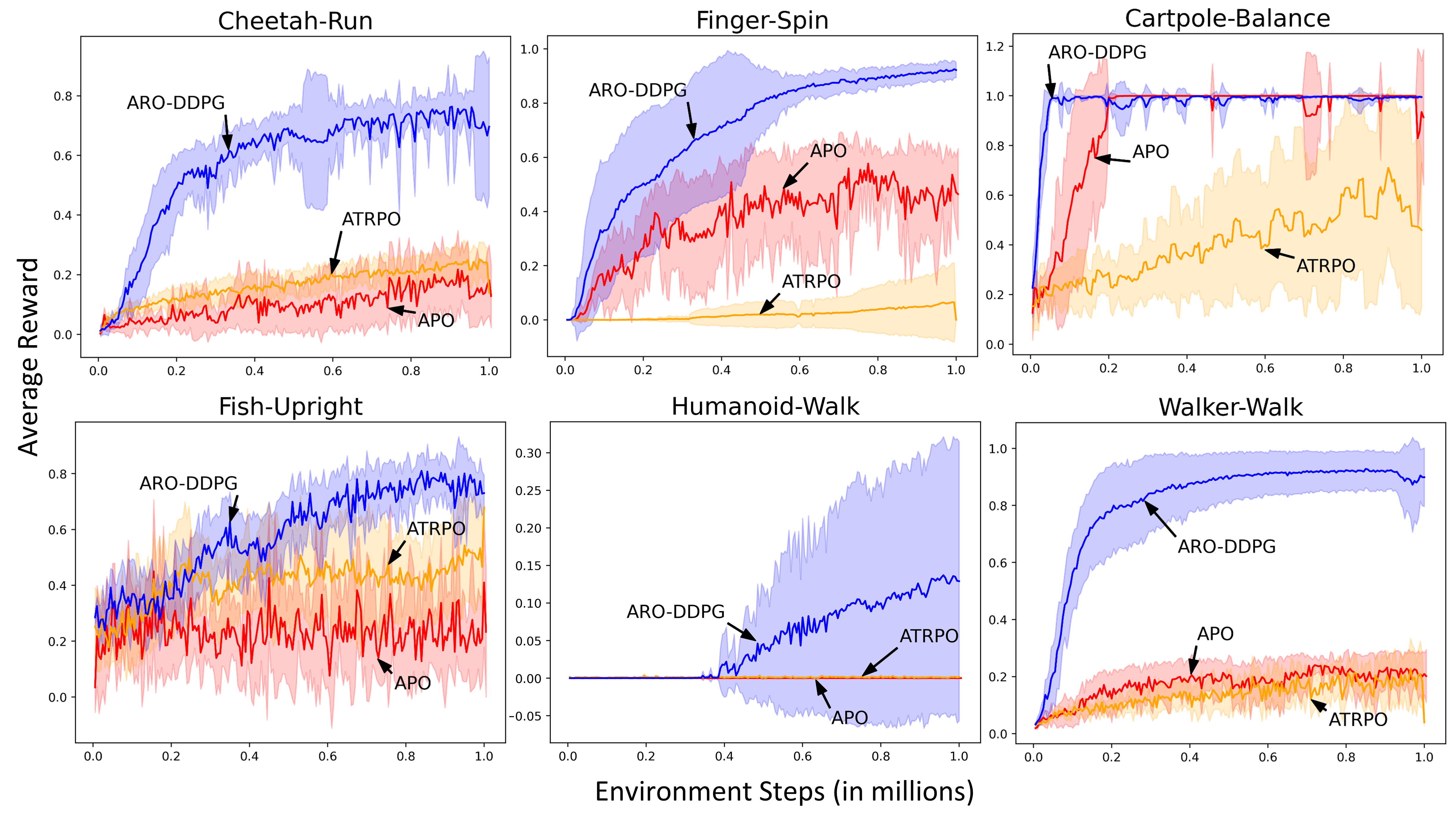}
    \caption{Comparison of performance of different average reward algorithms}
    \label{fig:1}
\end{figure*}

We conducted experiments on six different environments using the DeepMind control suite \citep{16} and found the performance of ARO-DDPG \footnote{ Pytorch implementation of ARO-DDPG could be found at this URL: \url{https://github.com/namansaxena9/ARO-DDPG}} to be superior than the  other algorithms (Figure \ref{fig:1}). All the environments selected are infinite horizon tasks. Maximum reward per time step is 1. None of the tasks have a goal reaching nature. We performed all the experiments using 10 different seeds. 
We show here performance comparisons with two state-of-the-art algorithms: the Average Reward TRPO (ATRPO) \citep{5} and the Average Policy Optimization (APO) \citep{6} respectively. 
In general for the average reward performance, not many algorithms are available in the literature.
We implemented the ATRPO algorithm using the instructions available in the original paper. We performed hyperparameter tuning and found the original hyper-parameters suggested by the author for ATRPO are the best.

For our proposed algorithm we trained the agent for 1 million time steps and evaluated the agent after every 5,000 time steps in the concerned environment. The length of each episode for the training phase was taken to be 1,000 and for the evaluation phase it was taken to be 10,000. The reason for taking longer episode length for evaluation phase was to compare the long term average reward performance of the algorithms. We also tried using episode length of 10,000 for training phase and found that to be giving poor average reward performance. 
We do not reset the agent if before completing 10,000 steps, it lands in a  state from where it is unable to escape of its own. The agent continues to get a reward of zero by default for the remaining length of the episode. That way the cost of failure is high. While training we updated the actor after performing a fixed number of environment steps. We updated the differential Q-value function neural network with more frequency as compared to the actor neural network. We used target actor and differential Q-value networks along with target estimator of the average reward parameter for stability while using bootstrapping updates. We updated the target network using polyak averaging. We tried to enforce multiple timescales in our algorithm by using different update frequency for actor, critic and polyak averaging for target networks. We also borrowed the double Q-network trick from \citet{7}. Complete information regarding the set of hyper-parameters used is provided in the appendix. 

\section{Related Work}\label{sc:6}
Actor-Critic algorithms for average reward performance criterion is much less studied compared to discounted reward performance criterion. One of the earliest works  on the average reward criterion is \citet{18}. In this paper, \citeauthor{18} compares the performance of R-learning with that of Q-learning and concludes that fine tuning is required to get better results from R-learning. R-learning is the average reward version of Q-learning. Later in \citeyear{9}, \citeauthor{9} derived the policy gradient theorem for both discounted and average reward criteria \citep{9}, which formed the bedrock for development of the average reward actor-critic algorithms. The first proof of asymptotic convergence of average reward actor-critic algorithms with function approximation appeared in \citet{23}. A temporal difference learning based off-policy control algorithm has been proposed in \citet{maei2010}. An incremental off-policy search algorithm based on the cross entropy method has been proposed in \cite{Joseph2018ce1}. Further, in \citet{24,B2009automatica}, incremental update natural policy gradient algorithms for the average reward setting have been proposed in the on-policy setting and asymptotic convergence proofs of the same provided. An off-policy variant of the natural actor-critic algorithm has been proposed in \citet{diddigi}. 

Recently, \citeauthor{25} presented a Differential Q-learning algorithm and claimed that their algorithm is able to find the exact differential value function without an offset. Further, \citeauthor{26} provided an extension of the options framework from the discounted setting to the average reward setting and demonstrated the performance of the algorithm in the Four-Room domain task. One of the major contributions in off-policy policy evaluation is made by \citet{27}. Here \citeauthor{27} gave a convergent off-policy evaluation scheme inspired from the gradient temporal difference learning algorithms but involving a primal-dual formulation making the policy evaluation step feasible for a neural network implementation. \citet{28} provided another convergent off-policy evaluation algorithm using target network and $l_2$-regularisation. In our work we use the same policy evaluation update. 

Our work in this paper is actually an extension of the work of \citet{10} from the discounted to the average reward setting. In \citet{21}, a finite time analysis for deterministic policy gradient algorithm was done for the discounted reward setting. We performed the finite time analysis for the average reward deterministic policy gradient algorithm and in particular obtain the same sample complexity for our algorithm as reported by \citet{29} for stochastic policies.

\section{Conclusion and Future Work}\label{sc:7}

In this paper we presented a deterministic policy gradient theorem for both on-policy and off-policy settings considering average reward performance criteria. We then proposed the Average Reward Off-policy Deep Deterministic Policy Gradient(ARO-DDPG) algorithm using neural network and replay buffer for high dimensional MuJoCo based environments. We observed superior performance of ARO-DDPG over existing average reward algorithms (ATRPO and APO). We first showed the asymptotic convergence using ODE-based method. Later we provided finite time analysis for the on-policy and off-policy algorithms based on the proposed policy gradient theorem and obtained the sample complexity of $\Omega(\epsilon^{-2.5})$. Lastly to extend the current line of work, one could try using natural gradient descent based update rule for deterministic policy. Further in the current work we tried optimizing the average reward performance (gain optimality). In the literature, optimizing the differential value function for all the states is mentioned as part of achieving Blackwell optimality. Hence actor-critic algorithms could be designed that not only optimize average reward performance but also differential value function (bias optimality). It would also be interesting to devise similar algorithms for constrained MDPs as with \cite{articlecmdpbhatnagar, cmdp2,bhatnagaretal}.

\section*{Acknowledgements}
N. Saxena was supported by a Ministry of Education (MoE) scholarship and a project from DST-ICPS. S. Khastagir was supported by a Ministry of Education (MoE) scholarship. S. Kolathaya was supported by a Pratiksha Trust Young Investigator Fellowship and the SERB grant no CRG/2021/008115. S. Bhatnagar was supported by the J.C. Bose Fellowship, Project No.~DFTM/02/3125/M/04/AIR-04 from DRDO under the DIA-RCOE scheme, a project from DST-ICPS, and the RBCCPS, IISc. 

\nocite{langley00}

\bibliography{references}
\bibliographystyle{icml2023}

\newpage
\appendix
\onecolumn
\section{Assumptions, Lemmas and Theorems}
\setcounter{equation}{0}
\setcounter{theorem}{0}
\numberwithin{equation}{section}

\subsection{Additional Assumptions}

We make the following additional assumptions.

\begin{assumption}\label{as:9}
The transition probability density function for a policy $\pi$ with parameter $\theta$ is Lipschitz continuous w.r.t $\theta$. Thus, $max_{s',s}|P^{\pi_1}(s'|s)-P^{\pi_2}(s'|s)| \leq L_t\|\theta_1 - \theta_2\|$.
\end{assumption}

The above assumption is a standard assumption in theoretical studies in literature. Reference for those assumptions can be found in \citet{21,31,32} and \citet{33}.

\begin{assumption}\label{as:10}
The reward function for a policy $\pi$ with parameter $\theta$ is Lipschitz continuous w.r.t $\theta$. Thus, $max_{s}|R^{\pi_1}(s)-R^{\pi_2}(s)| \leq L_r\|\theta_1 - \theta_2\|$.
\end{assumption}

The above assumption can be satified by using a well defined reward function to ensure Lipchitz continuity of reward function w.r.t action and then evoking Assumption \ref{as:6}.

\begin{assumption}\label{as:7}
The initial value of target estimators is bounded. Thus,  $\|\bar{w_0}\| \leq C_w$ and $\|\bar{\rho_0}\| \leq (Cr+ 2C_w)$. 
\end{assumption}

Assumption \ref{as:7} is used to enforce the stability of the iterates of target estimators.

\begin{assumption}\label{as:12}
Let $A(\theta) = \int d^{\pi}(s)(\phi^{\pi}(s)(\int P^{\pi}(s'|s) \phi^{\pi}(s') \,ds'-\phi^{\pi}(s))^{\intercal} - \eta I)\,ds$. $\lambda_{min}$ is the lower bound on the minimum eigenvalue of $A(\theta)$ for all values of $\theta$.
\end{assumption}

The assumption above is used in Lemma \ref{lm:a6} to prove the Lipchitz continuity of optimal differential Q-value function parameter $w^{*}$ for a particular value of policy parameter $\theta$ with respect to $\theta$.

\begin{assumption}\label{as:13}
Let $A'(\theta) = \int d^{\pi}(s)(\phi^{\pi}(s)(\int P^{\pi}(s'|s) \phi^{\pi}(s') \,ds'-\phi^{\pi}(s))^{\intercal})\,ds$. $\lambda_{max}^{all}$ is the upper bound on maximum eigenvalue of $(A'(\theta)+A'(\theta)^{\intercal})/2$ for all values of $\theta$. 
\end{assumption}

Assumption \ref{as:13} is used to prove the negative definiteness of the matrix $A_\theta$ (defined in Assumption \ref{as:12}) in Lemma \ref{lm:a11}.

\begin{assumption}\label{as:14}
Let $H_{\theta} = \int_{S}d^{\pi}(s)\nabla_{\theta}\pi(s,\theta)\nabla_{\theta}\pi(s,\theta)^{\intercal}\, ds$. $\lambda_{min}^{\epsilon} > 0$ is the lower bound on the minimum eigenvalues of $H_{\theta}$ for all values of $\theta$.
\end{assumption}

The above assumption is used in Lemma \ref{lm:a13} to make sure $H_{\theta}$ is invertible and optimal differential Q-value function parameter $w_{\epsilon}^{*}$ according to compatible function approximation lemma (Lemma \ref{lm:2}) can be obtained. Similar assumption is present in \citep{21}.

\begin{assumption}\label{as:15}
Let $A_{off}^{{\mu}'}(\theta) = \int d^{\mu}(s)(\phi^{\pi}(s)(\int P^{\pi}(s'|s) \phi^{\pi}(s') \,ds'-\phi^{\pi}(s))^{\intercal})\,ds$. $\chi_{max}^{all}$ is the upper bound on maximum eigenvalue of $(A_{off}^{{\mu}'}(\theta)+A_{off}^{{\mu}'}(\theta)^{\intercal})/2$ for behaviour policy $\mu$ and all values of $\theta$. 
\end{assumption}

Assumption \ref{as:15} is used to prove the negative definiteness of the matrix $A_\theta$ (defined in Lemma \ref{lm:a15}) in Lemma \ref{lm:a15.1}.

\begin{assumption}\label{as:16}
$\forall$ s policy $\pi$ is twice continuously differentiable i.e. $\nabla_{\theta}^2\pi(s)$ exists. 
\end{assumption}

Assumption \ref{as:16} can be satisfied by using neural network to parameterize the policy $\pi$.

\begin{assumption}\label{as:17}
$\phi(s,a)$ is Lipchitiz continuous w.r.t to $a$. Thus, $\forall\;s\;\; \|\phi(s,a_1) - \phi(s,a_2)\| \leq L_{a\phi}\|a_1 - a_2\|$.
\end{assumption}

Continuity of state action feature w.r.t to action is enforced using Assumption \ref{as:17}. Assumption \ref{as:17} is required for better generalisation of differential Q-value function and is in a way implied by Assumption \ref{as:5}.

\begin{assumption}\label{as:18}
The differential Q-value function is uniformly bounded, viz., $|Q^{\pi}_{diff}(s,a)|\leq C_Q < \infty$ 
\end{assumption}
The reason for satisfaction of Assumption \ref{as:18} is given in \citet{23}. 

\begin{assumption}\label{as:19}
$\nabla_{a}\phi(s,a)$ is Lipchitiz continuous w.r.t to $a$. Thus, $\forall\;s\;\; \|\nabla_{a}\phi(s,a_1) - \nabla_{a}\phi(s,a_2)\| \leq L_{a\Phi}\|a_1 - a_2\|$.
\end{assumption}

\subsection{Lemmas and Theorems for Policy Gradient}
\begin{lemma}\label{lm:a1}
There exists a unique constant $k(=\rho(\pi))$ which satisfies the following equation for differential value function $V_{diff}$ :  
\begin{equation*}
 V_{diff}^{\pi}(s_{t}) = \mathbb{E}^{\pi}[ R(s_{t},a_{t}) - k +  V_{diff}^{\pi}(s_{t+1})|s_{t}]. 
\end{equation*}
\end{lemma}

\begin{proof}

\begin{equation*}
\begin{split}
& V_{diff}^{\pi}(s_{t}) = R(s_{t},\pi(s_{t})) - k +  \int_{S} P^{\pi}(s_{t+1}|s_{t})V_{diff}^{\pi}(s_{t+1})\, ds_{t+1} \\
& \implies V_{diff}^{\pi}(s_{t}) -\int_{S} P^{\pi}(s_{t+1}|s_{t})V_{diff}^{\pi}(s_{t+1})\, ds_{t+1} = R(s_{t},\pi(s_{t})) - k \\
& \implies \sum_{t=0}^{T-1}\Bigl(V_{diff}^{\pi}(s_{t}) -\int_{S} P^{\pi}(s_{t+1}|s_{t})V_{diff}^{\pi}(s_{t+1})\, ds_{t+1}\Bigr) = \sum_{t=0}^{T-1}R(s_{t},\pi(s_{t})) - kT \\
\end{split}
\end{equation*}

Integrating w.r.t the stationary distribution $d^{\pi}$ of policy $\pi$ :

\begin{equation*}
\begin{split}
\sum_{t=0}^{T-1}\int_{S} d^{\pi}(s_t)\Bigl(V_{diff}^{\pi}(s_{t}) -\int_{S} P^{\pi}(s_{t+1}|s_{t})&V_{diff}^{\pi}(s_{t+1})\, ds_{t+1}\Bigr) \,ds_t \\& = \sum_{t=0}^{T-1} \int_{S} d^{\pi}(s_t) R(s_{t},\pi(s_{t}))\,ds - kT \\
\end{split}
\end{equation*}

\begin{equation*}
\begin{split}
\sum_{t=0}^{T-1}\Bigl(\int_{S} d^{\pi}(s_t)V_{diff}^{\pi}(s_{t})\,ds_t -\int_{S} d^{\pi}(s_{t+1})&V_{diff}^{\pi}(s_{t+1})\, ds_{t+1}\Bigr)  \\& = \sum_{t=0}^{T-1} \int_{S} d^{\pi}(s_t) R(s_{t},\pi(s_{t}))\,ds_t - kT \\
\end{split}
\end{equation*}

Note: $\Bigl(\int_{S} d^{\pi}(s_t)V_{diff}^{\pi}(s_{t})\,ds_t -\int_{S} d^{\pi}(s_{t+1}) V_{diff}^{\pi}(s_{t+1})\, ds_{t+1}\Bigr) = 0$.

\begin{equation*}
\begin{split}
&\implies k = \frac{1}{T}\sum_{t=0}^{T-1} \int_{S} d^{\pi}(s_t) R(s_{t},\pi(s_{t}))\,ds_t \\ &\implies k = \lim_{T\to \infty} \frac{1}{T}\sum_{t=0}^{T-1} \int_{S} d^{\pi}(s_t) R(s_{t},\pi(s_{t}))\,ds_t \\  
&\implies k = \rho(\pi) \quad (\text{using}\;\; (\ref{eq:3})).
\end{split}
\end{equation*}
\end{proof}

\begin{theorem}\label{th:a1}
 The gradient of $\rho(\pi)$ with respect to the policy parameter $\theta$ is given as follows:
\begin{equation*}
\nabla_{\theta} \rho(\pi) = \int_{S}d^{\pi}(s)\nabla_{a}Q_{diff}^{\pi}(s,a)|_{a =\pi(s)}\nabla_{\theta}\pi(s,\theta) \,ds.
\end{equation*}
\end{theorem}

\begin{proof}
Using Lemma \ref{lm:1}:
\begin{equation*}
\begin{split}
& V_{diff}^{\pi}(s_{t}) = R(s_{t},\pi(s_{t})) - \rho(\pi) +  \int_{S} P^{\pi}(s_{t+1}|s_{t})V_{diff}^{\pi}(s_{t+1})\, ds_{t+1} \\
& \implies Q_{diff}^{\pi}(s_{t},\pi(s_{t})) = R(s_{t},\pi(s_{t})) - \rho(\pi) + \int_{S} P^{\pi}(s_{t+1}|s_{t})Q_{diff}^{\pi}(s_{t+1},\pi(s_{t+1}))\, ds_{t+1}
\end{split}
\end{equation*}

Differentiating w.r.t $\theta$, we obtain

\begin{equation*}
\begin{split}
\nabla_\theta Q_{diff}^{\pi}(s_{t},\pi(s_{t})) & = \nabla_\theta R(s_{t},\pi(s_{t})) - \nabla_\theta \rho(\pi) \\ 
& \quad +\nabla_\theta \Bigl( \int_{S} P^{\pi}(s_{t+1}|s_{t})Q_{diff}^{\pi}(s_{t+1},\pi(s_{t+1}))\, ds_{t+1} \Bigr) \\
& = \nabla_aR(s_{t},a)|_{a = \pi(s_t)}\nabla_\theta \pi(s_t) - \nabla_\theta \rho(\pi) \\
& \quad+ \int_{S} \nabla_aP^{\pi}(s_{t+1}|s_{t},a)|_{a = \pi(s_t)}\nabla_\theta \pi(s_t) Q_{diff}^{\pi}(s_{t+1},\pi(s_{t+1}))\, ds_{t+1}  \\
& \quad+  \int_{S} P^{\pi}(s_{t+1}|s_{t})\nabla_\theta Q_{diff}^{\pi}(s_{t+1},\pi(s_{t+1}))\, ds_{t+1}.  \\
\end{split}
\end{equation*}

Note: $\nabla_a\rho(\pi) = \nabla_a \Bigl(\int_{S} d^{\pi}(s) R^{\pi}(s) \,ds \Bigr) = 0$.
\begin{equation*}
\begin{split}
\implies \nabla_\theta Q_{diff}^{\pi}(s_{t},\pi(s_{t})) 
& = \nabla_{a}Q_{diff}^{\pi}(s_t,a)|_{a =\pi(s_t)}\nabla_{\theta}\pi(s_t) - \nabla_\theta \rho(\pi)  \\
& \quad+  \int_{S} P^{\pi}(s_{t+1}|s_{t})\nabla_\theta Q_{diff}^{\pi}(s_{t+1},\pi(s_{t+1}))\, ds_{t+1}.  
\end{split}
\end{equation*}

Integrating w.r.t stationary distribution $d^{\pi}(\cdot)$ of policy $\pi$:

\begin{equation*}
\begin{split}
\int_{S} d^{\pi}(s_t) \nabla_\theta Q_{diff}^{\pi}(s_{t},\pi(s_{t})) ds_t 
& = \int_{S} d^{\pi}(s_t) \nabla_{a}Q_{diff}^{\pi}(s_t,a)|_{a =\pi(s_t)}\nabla_{\theta}\pi(s_t) ds_t - \nabla_\theta \rho(\pi)  \\
& + \int_{S} d^{\pi}(s_t) \int_{S} P^{\pi}(s_{t+1}|s_{t})\nabla_\theta Q_{diff}^{\pi}(s_{t+1},\pi(s_{t+1}))\, ds_{t+1} \, ds_t. \\
\end{split}
\end{equation*}
Note: $\int_{S} d^{\pi}(s)P^{\pi}(s'|s)\,ds = d^{\pi}(s')$. Thus,
\begin{equation*}
\begin{split}
\nabla_\theta \rho(\pi) 
& = \int_{S} d^{\pi}(s_t) \nabla_{a}Q_{diff}^{\pi}(s_t,a)|_{a =\pi(s_t)}\nabla_{\theta}\pi(s_t) ds_t  \\
& \quad+ \int_{S} d^{\pi}(s_{t+1}) \nabla_\theta Q_{diff}^{\pi}(s_{t+1},\pi(s_{t+1}))\, ds_{t+1} \\
& \quad-\int_{S} d^{\pi}(s_t) \nabla_\theta Q_{diff}^{\pi}(s_{t},\pi(s_{t})) ds_t. 
\end{split}
\end{equation*}

\begin{equation*}
\begin{split}
\nabla_\theta \rho(\pi) 
& = \int_{S} d^{\pi}(s) \nabla_{a}Q_{diff}^{\pi}(s,a)|_{a =\pi(s)}\nabla_{\theta}\pi(s) \,ds.
\end{split}
\end{equation*}
\end{proof}

\begin{lemma}\label{lm:a2}
For on-policy case, assume that the differential Q-value function (\ref{eq:7}) satisfies the following:
\begin{enumerate}
\item\begin{equation*}
\nabla_{w}\nabla_{a} Q^{w}_{diff}(s,a)|_{a=\pi(s)} = \nabla_{\theta} \pi(s,\theta).   
\end{equation*}
\item
The differential Q-value function parameter $w = w_\epsilon^{*}$ optimizes the following error function:
\begin{equation*}
\zeta(\theta,w) = \frac{1}{2}\int_{S}d^{\pi}(s)\|\nabla_{a}Q_{diff}^{\pi}(s,a)|_{a =\pi(s)}-\nabla_{a} Q_{diff}^{w}(s,a)|_{a =\pi(s)}\|^2 \,ds. \end{equation*}
\end{enumerate}
Then,
\begin{equation*}
\begin{split}
\int_{S}d^{\pi}(s)\nabla_{a}Q_{diff}^{\pi}(s,a)|_{a =\pi(s)}\nabla_{\theta}\pi(s,\theta) \,ds = \int_{S}d^{\pi}(s)\nabla_{a}Q_{diff}^{w}(s,a)|_{a =\pi(s)}\nabla_{\theta}\pi(s,\theta) \,ds.
\end{split}
\end{equation*}
Further, 
\begin{equation*}
\nabla_{a} Q^{w}_{diff}(s,a)|_{a=\pi(s)} = \nabla_{\theta} \pi(s,\theta)^{\intercal}w \;\;\;(\text{for linear function approximator}).   
\end{equation*}

\end{lemma}

\begin{proof}
Let $\mathcal{E}(\theta,w,s) = \nabla_{a}Q_{diff}^{\pi}(s,a)|_{a =\pi(s)}-\nabla_{a} Q_{diff}^{w}(s,a)|_{a =\pi(s)}$,

\begin{equation*}
\begin{split}
& \zeta(\theta,w) = \frac{1}{2} \int_{S}d^{\pi}(s) \mathcal{E}(\theta,w,s)^{\intercal}\mathcal{E}(\theta,w,s) \,ds.      
\end{split}
\end{equation*}

Differentiating w.r.t the differential Q-value function parameter $w$, we obtain:
\begin{equation*}
\begin{split}
\nabla_w \zeta(\theta,w) & = \int_{S}d^{\pi}(s) \nabla_w\mathcal{E}(\theta,w,s)\mathcal{E}(\theta,w,s) \,ds \\
& = - \int_{S}d^{\pi}(s) \nabla_w\nabla_{a} Q_{diff}^{w}(s,a)|_{a =\pi(s)}\Bigl(\nabla_{a}Q_{diff}^{\pi}(s,a)|_{a =\pi(s)} \\
& \quad-\nabla_{a} Q_{diff}^{w}(s,a)|_{a =\pi(s)}\Bigr) \,ds = 0.
\end{split}
\end{equation*}
Letting $\nabla_w\nabla_{a} Q_{diff}^{w}(s,a)|_{a =\pi(s)} = \nabla_\theta\pi(s)$, we obtain

\begin{equation*}
\begin{split}
\int_{S}d^{\pi}(s)\nabla_{a}Q_{diff}^{\pi}(s,a)|_{a =\pi(s)}\nabla_{\theta}\pi(s,\theta) \,ds = \int_{S}d^{\pi}(s)\nabla_{a}Q_{diff}^{w}(s,a)|_{a =\pi(s)}\nabla_{\theta}\pi(s,\theta) \,ds.
\end{split}
\end{equation*}

Let us consider the case of linear function approximator with parameter $w$, i.e., $Q_{diff}^{w}(s,\pi(s)) = \phi^{\pi}(s,\pi(s))^{\intercal}w$.

We know from above,

\begin{equation}\label{eq:a0}
\begin{split}
& \nabla_w\nabla_{a} Q_{diff}^{w}(s,a)|_{a =\pi(s)} = \nabla_\theta\pi(s) \\
& \implies \nabla_{a} \phi^{\pi}(s,a)|_{a =\pi(s)} = \nabla_\theta\pi(s).
\end{split}
\end{equation}
Thus, 
\begin{equation*}
\begin{split}
& Q_{diff}^{w}(s,a) = \phi^{\pi}(s,a)^{\intercal}w \\
& \implies \nabla_{a} Q_{diff}^{w}(s,a)|_{a =\pi(s)} = \nabla_{a} \phi^{\pi}(s,a)|_{a =\pi(s)}^{\intercal}w \\
& \implies \nabla_{a} Q_{diff}^{w}(s,a)|_{a =\pi(s)} = \nabla_\theta\pi(s)^{\intercal}w \;\; (\text{using} \;\;(\ref{eq:a0})). \\
\end{split}
\end{equation*}
\end{proof}

\begin{theorem}\label{th:a2}
The approximate gradient ($\widehat{\nabla_{\theta} \rho}(\pi)$) of the average reward $\rho(\pi)$ with respect to the policy parameter $\theta$ is given by the following expression:
\begin{equation}\label{eq:a0.1}
\begin{split}
\widehat{\nabla_{\theta} \rho}(\pi) =
 \int_{S}d^{\mu}(s)\nabla_{a}Q_{diff}^{\pi}(s,a)|_{a=\pi(s)}\nabla_{\theta}\pi(s,\theta) \,ds .
\end{split}
\end{equation}
Further, the approximation error is
 $\mathcal{E}(\pi,\mu) = \| \nabla_{\theta} \rho(\pi) - \widehat{\nabla_{\theta} \rho}(\pi) \|,$
where $\mu$ represents the behaviour policy with parameter $\theta^{\mu}$ and $\nabla_{\theta} \rho(\pi)$ is the on-policy policy gradient from Theorem \ref{th:1}. $\mathcal{E}$ satisfies
\begin{equation}
\mathcal{E}(\pi,\mu)     \leq Z \| \theta - \theta^{\mu} \|,
\end{equation}
where, $Z = 2^{n+1}C(\lceil{\log_{\kappa}a^{-1}\rceil} + 1\slash\kappa)L_{t}$ with $L_{t}$ being the Lipchitz constant for the transition probability density function (Assumption \ref{as:9}). Constants $a$ and $\kappa$ are from Assumption \ref{as:2}, $n$ is the dimension of the state space, and $C = \max_s \|\nabla_{a}Q_{diff}^{\pi}(s,a)|_{a=\pi(s)}\nabla_{\theta}\pi(s,\theta)\|.$

\end{theorem}
\begin{proof}

\begin{equation*}
\begin{split}
\mathcal{E}(\pi,\mu) & = \| \nabla_{\theta} \rho(\pi) - \widehat{\nabla_{\theta} \rho}(\pi) \| \\ 
& = \|\int_{S}d^{\pi}(s)\nabla_{a}Q_{diff}^{\pi}(s,a)|_{a=\pi(s)}\nabla_{\theta}\pi(s,\theta) \,ds \\
& \quad -\int_{S}d^{\mu}(s)\nabla_{a}Q_{diff}^{\pi}(s,a)|_{a=\pi(s)}\nabla_{\theta}\pi(s,\theta) \,ds \| \\
& \leq \int_{S}|d^{\pi}(s)-d^{\mu}(s)| \|\nabla_{a}Q_{diff}^{\pi}(s,a)|_{a=\pi(s)}\nabla_{\theta}\pi(s,\theta)\| \,ds \\
& \leq C \int_{S}|d^{\pi}(s)-d^{\mu}(s)| \,ds.
\end{split}    
\end{equation*}
Here, $C = \max_s \|\nabla_{a}Q_{diff}^{\pi}(s,a)|_{a=\pi(s)}\nabla_{\theta}\pi(s,\theta)\|.$
Thus,

\begin{equation*}
\begin{split}
\mathcal{E}(\pi,\mu) \leq CL_d\| \theta - \theta^{\mu} \| = Z\| \theta - \theta^{\mu} \| \;(\text{using Lemma }\ref{lm:a12}).
\end{split}
\end{equation*}

Here, $Z = 2^{n+1}C(\lceil{\log_{\kappa}a^{-1}\rceil} + 1\slash\kappa)L_{t}$.

\end{proof}

\begin{lemma}\label{lm:a2.1}
Let policy $\pi$ be parameterized by $\theta$ and $\mu$ be the behaviour policy. Assume that the differential Q-value function (\ref{eq:7}) satisfies the following:
\begin{enumerate}
\item\begin{equation*}
\nabla_{w}\nabla_{a} Q^{w}_{diff}(s,a)|_{a=\pi(s)} = \nabla_{\theta} \pi(s,\theta).   
\end{equation*}
\item
The differential Q-value function parameter $w = w_\epsilon^{*}$ optimizes the following error function:
\begin{equation*}
\zeta'(\theta,w,\mu) = \frac{1}{2}\int_{S}d^{\mu}(s)\|\nabla_{a}Q_{diff}^{\pi}(s,a)|_{a =\pi(s)}-\nabla_{a} Q_{diff}^{w}(s,a)|_{a =\pi(s)}\|^2 \,ds. \end{equation*}
\end{enumerate}
Then,
\begin{equation*}
\begin{split}
\int_{S}d^{\mu}(s)\nabla_{a}Q_{diff}^{\pi}(s,a)|_{a =\pi(s)}\nabla_{\theta}\pi(s,\theta) \,ds = \int_{S}d^{\mu}(s)\nabla_{a}Q_{diff}^{w}(s,a)|_{a =\pi(s)}\nabla_{\theta}\pi(s,\theta) \,ds.
\end{split}
\end{equation*}
Further, 
\begin{equation*}
\nabla_{a} Q^{w}_{diff}(s,a)|_{a=\pi(s)} = \nabla_{\theta} \pi(s,\theta)^{\intercal}w \;\;\;(\text{for linear function approximator}).   
\end{equation*}
\end{lemma}

\begin{proof}
The proof will follow in the same way as proof for Lemma \ref{lm:a2}.    
\end{proof}

\subsection{Finite Time Analysis}
Figure \ref{fig:2} shows the relation between different types of error in the on-policy algorithm (Algorithm \ref{alg:2}). Here, arrow from value function error to actor error shows that the actor error is dependent on the value function error. Similar relationship follows for the rest of the error types. Figure \ref{fig:2} intuitively explains which lemma utilizes which lemma to arrive at the final result (Theorem \ref{th:a3}).

\begin{figure}[h]
\centering
\scalebox{0.8}
{
\begin{tikzpicture}[node distance=1cm]
\node (actor) [boxes] {Actor Error $$\frac{1}{T}\sum^{T-1}_{t=0} E \|\nabla_{\theta}\rho(\theta_t)\|^2$$ (Lemma \ref{lm:a3})};
\node (value) [boxes, below = of actor, xshift = -3.0cm] {Differential Q-value Function Error $$\frac{1}{T}\sum_{t=0}^{T-1}\mathbb{E}||\Delta w_t||^2$$ (Lemma \ref{lm:a4})};
\node (tvalue) [boxes, below = of value] {Target Differential Q-value Function Error $$\frac{1}{T}\sum_{t=0}^{T-1}\mathbb{E}\|\Delta\bar{w}_t\|^2$$ (Lemma \ref{lm:a4.8})};
\node (trho) [boxes, below = of actor, xshift = 3.0cm, yshift = -0.2cm] {Target Average Reward Error $$\frac{1}{T}\sum_{t=0}^{T-1}\mathbb{E}|\Delta\bar{\rho}_t|^2$$ (Lemma \ref{lm:a4.9}) };
\node (rho) [boxes, below = of trho, yshift = -0.4cm] {Average Reward Error $$\frac{1}{T}\sum_{t=0}^{T-1}\mathbb{E}|\Delta\rho_t|^2$$ (Lemma \ref{lm:a5})};

\draw [arrow] (value) |- (actor);
\draw [arrow2] (value) -- (tvalue);
\draw [arrow] (trho) -- (value);
\draw [arrow] (tvalue) -- (rho);
\draw [arrow] (rho) -- (trho);
\end{tikzpicture}
}
\caption{Dependency of errors in different types of parameters in Algorithm \ref{alg:2} on one another.}
\label{fig:2}
\end{figure}

\subsubsection{Main Lemmas and Theorems}

\begin{lemma}\label{lm:a3} Let the cumulative error of on-policy actor be $\sum^{T-1}_{t=0} E ||\nabla_{\theta}\rho(\theta_t)||^2$ and cumulative error of differential Q-value function be $\sum^{T-1}_{t=0}E ||\Delta w_{t}||^2$. $\theta_t$ and $w_t$ are the actor and linear differential Q-value function parameter at time t. Bound on the cumulative error of on-policy actor is proven using cumulative error of differential Q-value function as follows:

\begin{equation*}
\begin{split}    
\frac{1}{T}\sum^{T-1}_{t=0} E ||\nabla_{\theta} \rho(\theta_t)||^2 & \leq 4\frac{C_r}{C_\gamma}T^{-1} + 3C_{\pi}^2C_{a\phi}^2(\frac{1}{T}\sum^{T-1}_{t=0}E ||\Delta w_{t}||^2)
+ 3C_{\pi}^2(C_{a\phi}^2\tau^2 + \frac{4}{M}C_{\pi}^2C_{w_{\epsilon}}^2) \\
& \;\;+ \frac{C_{\gamma}L_{J}G_{\theta}^2}{1-v}T^{-v}
\end{split}
\end{equation*}

Here, $C_r$ is the upper bound on rewards (Assumption \ref{as:4}) , $C_\gamma$, v are constants used for step size $\gamma_t$ (Assumption \ref{as:2.1}, $\|\nabla_{\theta} \pi(s)\| \leq C_{\pi}$ (Assumption \ref{as:6}), $\Delta w_{t} = w_{t} - w_{t}^{*}, \tau = \max_{t} \|w_{t}^{*} - w_{\epsilon,t}^{*}\|$, $w_{\epsilon}^{*}$ is the optimal differential Q-value function parameter according to Lemma \ref{lm:2}. $w_t^{*}$ is the optimal parameters given by TD(0) algorithm corresponding to policy parameter $\theta_t$. Constant $C_{w_{\epsilon}^{*}}$ is defined in Lemma \ref{lm:a13}. $L_{J}$ is the coefficient used in smoothness condition of the non convex function $\rho(\theta)$. Constant $G_{\theta}$ is defined in  Lemma \ref{lm:a7}. M is the size of batch of samples used to update parameters. 
\end{lemma}

\begin{proof}

By [$-L_{J},L_{J}$]-smoothness of non-convex function we have: 
\\
\begin{equation}\label{eq:a1}
E[\rho(\theta_{t+1})] \geq E[\rho(\theta_{t})] + E\langle \nabla_{\theta} \rho(\theta_{t}),\theta_{t+1} - \theta_{t} \rangle - \dfrac{L_{J}}{2}E\|\theta_{t+1} - \theta_{t}\|^2.
\end{equation}

Now, $$h(B_{t}, w_{t}, \theta_{t}) = \dfrac{1}{M} \sum_{i} \nabla_{a} Q^{w_t}_{diff}(s_{t,i},a)|_{a = \pi(s_{t,i})} \nabla_{\theta} \pi(s_{t,i}). $$

Here, $B_{t}$ refers to the batch of transitions sampled from the buffer at time $t$ and $\forall \;i\;\; s_{t,i} \in B_t$. 

\begin{equation}\label{eq:a2}
\begin{split}
E\langle \nabla_{\theta} \rho(\theta_{t}),\theta_{t+1} - \theta_{t} \rangle & = \gamma_{t}E\langle \nabla_{\theta} \rho(\theta_{t}),h(B_{t}, w_{t}, \theta_{t}) \rangle \\ 
& =\gamma_{t}E\langle \nabla_{\theta} \rho(\theta_{t}),h(B_{t}, w_{t}, \theta_{t}) - \nabla_{\theta}\rho(\theta_{t}) \rangle + \gamma_{t} E\|\nabla_{\theta}\rho(\theta_{t})\|^2.
\end{split}
\end{equation}

From (\ref{eq:a2}), we have
\begin{equation}\label{eq:a3}
\begin{split}
E\langle \nabla_{\theta} \rho(\theta_{t}),h(B_{t}, w_{t}, \theta_{t}) - \nabla_{\theta}\rho(\theta_{t}) \rangle & \geq -\dfrac{1}{2}E\|\nabla_\theta\rho(\theta_{t})\|^2 -\dfrac{1}{2}E\|h(B_{t}, w_{t}, \theta_{t}) - \nabla_{\theta}\rho(\theta_{t})\|^2 \\
(\because x^\intercal y \geq -\|x\|^2/2-\|y\|^2/2).
\end{split}
\end{equation}

From (\ref{eq:a3}):
\begin{equation} \label{eq:a4}
\begin{split}
& E\|h(B_{t}, w_{t}, \theta_{t}) - \nabla_{\theta}\rho(\theta_{t})\|^2\\& =
E\|h(B_{t}, w_{t}, \theta_{t}) - h(B_{t}, w_{t}^{*}, \theta_{t}) + h(B_{t}, w_{t}^{*}, \theta_{t}) - h(B_{t}, w_{\epsilon,t}^{*}, \theta_{t}) +h(B_{t}, w_{\epsilon,t}^{*}, \theta_{t}) - \nabla_{\theta}\rho(\theta_{t})\|^2 \\ 
& \leq 3(E\|h(B_{t}, w_{t}, \theta_{t}) - h(B_{t}, w_{t}^{*}, \theta_{t})\|^2 \term{1} \\ 
& \;\;\;+ E\|h(B_{t}, w_{t}^{*}, \theta_{t}) - h(B_{t}, w_{\epsilon,t}^{*}, \theta_{t})\|^2 \term{2} \\
& \;\;\;+ E\|h(B_{t}, w_{\epsilon,t}^{*}, \theta_{t}) - \nabla_{\theta}\rho(\theta_{t})\|^2) \term{3}
\end{split}
\end{equation}

In (\ref{eq:a4}), $w_t^{*}$ refers to the point of convergence of on-policy TD(0) algorithm with l2-regularisation and policy $\pi(\theta_t)$ and $w_{\epsilon,t}^{*}$ refers to the optimal value function parameter according to Compatible Function Approximation Lemma \ref{lm:2}.

From (\ref{eq:a4}):

\termw{1}:

\begin{equation*}
\begin{split}
&E||h(B_{t}, w_{t}, \theta_{t}) - h(B_{t}, w_{t}^{*}, \theta_{t})||^2 \\
& = \dfrac{1}{M} || \sum_{i=0}^{M-1} \nabla_{a} Q^{w_{t}}_{diff}(s_{t,i},a)|_{a = \pi(s_{t,i})} \nabla_{\theta} \pi(s_{t,i}) - \sum_{i=0}^{M-1} \nabla_{a} Q^{w_{t}^{*}}_{diff}(s_{t,i},a)|_{a = \pi(s_{t,i})} \nabla_{\theta} \pi(s_{t,i}) ||^2 \\
& = E ||\dfrac{1}{M} \sum_{i=0}^{M-1}  \nabla_{\theta} \pi(s_{t,i}) \nabla_{a} \phi(s_{t,i},a)|_{a = \pi(s_{t,i})}^\intercal (w_t - w_t^{*}) ||^2 \\
& \leq C_{\pi}^{2}C_{a\phi}^2 E ||w_t - w_t^{*}||^2.
\end{split}
\end{equation*}

\termw{2} is similar as \termw{1}:

\begin{equation*}
\begin{split}
E||h(B_{t}, w_{t}^{*}, \theta_{t}) - h(B_{t}, w_{\epsilon,t}^{*}, \theta_{t})||^2 
& \leq C_{\pi}^{2}C_{a\phi}^2 E ||w_t^{*} - w_{\epsilon,t}^{*}||^2 \\
& \leq C_{\pi}^{2}C_{a\phi}^2\tau^2.
\end{split}
\end{equation*}

\termw{3} : \\
\begin{itemize}
\item By Compatible Function Approximation Lemma~\ref{lm:2}: $\nabla_\theta \rho(\theta_{t}) = \int_{S}d^{\pi}(s) \nabla_{\theta} \pi(s) \nabla_{\theta} \pi(s)^\intercal w_{\epsilon,t}^{*} \,ds = E[h(B_t, w_{\epsilon,t}^{*},\theta_t)]$\\

\item By lemma 4 \citep{21}, if $E[\hat{Y}] = \bar{Y}, ||\hat{Y}||,||\bar{Y}|| \leq C_{Y}$ then,
$$ E||\dfrac{1}{M}\sum_{i=0}^{M-1} \hat{Y}_{i} - \bar{Y}|| \leq 4\frac{C_{Y}^2}{M}. $$
\end{itemize}

Using above two bullet points: 
\begin{equation*}
\begin{split}
E||h(B_{t}, w_{\epsilon,t}^{*}, \theta_{t}) - \nabla_{\theta} \rho(\theta_{t})||^2 
& \leq \dfrac{4}{M}||\nabla_{\theta} \pi(s) \nabla_{\theta} \pi(s)^\intercal w_{\epsilon,t}^{*} ||^2 \\
& \leq \dfrac{4C_{\pi}^4C_{w_{\epsilon}}^2}{M}.
\end{split}
\end{equation*}

Combining \termw{1},\termw{2} and \termw{3} and using in (\ref{eq:a4}): 

\begin{equation}\label{eq:a5}
E||h(B_{t}, w_{t}, \theta_{t}) - \nabla_{\theta} \rho(\theta_{t})||^2 \leq 3 C_{\pi}^{2} (C_{a\phi}^2E ||w_t - w_t^{*}||^2 + C_{a\phi}^2\tau^2 + \dfrac{4C_{\pi}^2C_{w_{\epsilon}}^2}{M} ).
\end{equation}

Using (\ref{eq:a5}) in (\ref{eq:a3}):

\begin{equation}\label{eq:a6}
\begin{split}
E\langle \nabla_{\theta} \rho(\theta_{t}),h(B_{t}, w_{t}, \theta_{t}) - \nabla_{\theta} \rho(\theta_{t}) \rangle & \geq -\dfrac{1}{2}E||\nabla_\theta\rho(\theta_{t})||^2 \\ 
& -\dfrac{3}{2} C_{\pi}^{2} (C_{a\phi}^2E ||w_t - w_t^{*}||^2 + C_{a\phi}^2\tau^2 + \dfrac{4C_{\pi}^2C_{w_{\epsilon}}^2}{M} ).
\end{split}
\end{equation}

Using (\ref{eq:a6}) in (\ref{eq:a2}):

\begin{equation}\label{eq:a7}
\begin{split}
E\langle \nabla_{\theta} \rho(\theta_{t}),\theta_{t+1} - \theta_{t} \rangle & \geq \dfrac{\gamma_t}{2}E||\nabla_\theta\rho(\theta_{t})||^2 \\ 
& -\dfrac{3\gamma_t}{2} C_{\pi}^{2} (C_{a\phi}^2E ||w_t - w_t^{*}||^2 + C_{a\phi}^2\tau^2 + \dfrac{4C_{\pi}^2C_{w_{\epsilon}}^2}{M} ).    
\end{split}    
\end{equation}

Using (\ref{eq:a7}) in (\ref{eq:a1}):
\begin{equation*}
\begin{split}
E[\rho(\theta_{t+1})] - E[\rho(\theta_{t})] & \geq
\dfrac{\gamma_t}{2}E||\nabla_\theta\rho(\theta_{t})||^2 - \dfrac{L_{J}}{2}E||\theta_{t+1} - \theta_{t}||^2 \\ 
& -\dfrac{3\gamma_t}{2} C_{\pi}^{2} (C_{a\phi}^2E ||w_t - w_t^{*}||^2 + C_{a\phi}^2\tau^2 + \dfrac{4C_{\pi}^2C_{w_{\epsilon}}^2}{M} )     
\end{split}    
\end{equation*}

\begin{equation*}
\begin{split}
\implies E||\nabla_\theta\rho(\theta_{t})||^2 & \leq \frac{2}{\gamma_t}\Big(E[\rho(\theta_{t+1})] - E[\rho(\theta_{t})]\Big) + 3 C_{\pi}^{2}C_{a\phi}^2 (E ||w_t - w_t^{*}||^2) \\ 
& + 3 C_{\pi}^{2} (C_{a\phi}^2\tau^2 + \dfrac{4C_{\pi}^2C_{w_{\epsilon}}^2}{M}) + L_{J}\gamma_tG_{\theta}^2 \;\;\;( \text{using Lemma \ref{lm:a7}})
\end{split}    
\end{equation*}

\begin{equation}\label{eq:a8}
\begin{split}
\implies \sum_{t=0}^{T-1} E||\nabla_\theta\rho(\theta_{t})||^2 & \leq \sum_{t=0}^{T-1} \frac{2}{\gamma_t}\Big(E[\rho(\theta_{t+1})] - E[\rho(\theta_{t})]\Big) \term{1} \\
& + \sum_{t=0}^{T-1} 3 C_{\pi}^{2}C_{a\phi}^2 (E ||w_t - w_t^{*}||^2) \term{2} \\ 
& + \sum_{t=0}^{T-1} 3 C_{\pi}^{2} (C_{a\phi}^2\tau^2 + \dfrac{4C_{\pi}^2C_{w_{\epsilon}}^2}{M}) \term{3} \\
& + \sum_{t=0}^{T-1} L_{J}\gamma_tG_{\theta}^2 \term{4} \;\;\;( \text{using Lemma \ref{lm:a7}})
\end{split}    
\end{equation}

From \eqref{eq:a8}

\termw{1}:
\begin{equation*}
\begin{split}
\sum_{t=0}^{T-1} \frac{2}{\gamma_t}\Big(E[\rho(\theta_{t+1})] - E[\rho(\theta_{t})]\Big) 
&= 2\biggl(\sum_{t=1}^{T-1}\Bigl( \frac{1}{\gamma_{t-1}} - \frac{1}{\gamma_{t}}\Bigr) E[\rho(\theta_t)] - \frac{ E[\rho(\theta_0)]}{\gamma_0} + \frac{ E[\rho(\theta_T)]}{\gamma_{T-1}}\biggr) \\
& \leq 2\biggl(\sum_{t=1}^{T-1}\Bigl( \frac{1}{\gamma_{t-1}} - \frac{1}{\gamma_{t-1}}\Bigr) E[\rho(\theta_t)] + \frac{ E[\rho(\theta_T)]}{\gamma_{T-1}} +\Big|\frac{ E[\rho(\theta_0)]}{\gamma_0}\Big|\biggr) \\
& \leq 2\biggl(\sum_{t=1}^{T-1}\Bigl( \frac{1}{\gamma_{t-1}} - \frac{1}{\gamma_{t}}\Bigr) + \frac{1}{\gamma_{T-1}} + \frac{1}{\gamma_{0}} \biggr)C_{r} \\
& \leq \frac{4C_r}{\gamma_{0}} = \frac{4C_r}{C_\gamma} 
\end{split}
\end{equation*}

\termw{2}:
\begin{equation*}
\sum_{t=0}^{T-1} 3 C_{\pi}^{2}C_{a\phi} (E ||w_t - w_t^{*}||^2) = \sum_{t=0}^{T-1} 3 C_{\pi}^{2}C_{a\phi}^2 (E ||\Delta w_t||^2)
\end{equation*}

\termw{4}:
\begin{equation*}
\sum_{t=0}^{T-1} L_{J}\gamma_tG_{\theta}^2 \leq L_{J}G_{\theta}^2C_\gamma\frac{T^{1-v}}{1-v} \;\;\Bigl(\because \sum_{t=0}^{T-1}\frac{1}{(1+t)^v} \leq \int_{0}^{T}\frac{1}{t^v} \,dt = \frac{T^{1-v}}{1-v} \Bigr)
\end{equation*}
Using \termw{1}-\termw{4} and dividing \eqref{eq:a8} by T:
\begin{equation*}
\begin{split}    
\frac{1}{T}\sum^{T-1}_{t=0} E ||\nabla_{\theta} \rho(\theta_t)||^2 & \leq 4\frac{C_r}{C_\gamma}T^{-1} + 3C_{\pi}^2C_{a\phi}^2(\frac{1}{T}\sum^{T-1}_{t=0}E ||\Delta w_{t}||^2)
+ 3C_{\pi}^2(C_{a\phi}^2\tau^2 + \frac{4}{M}C_{\pi}^2C_{w_{\epsilon}}^2) \\
& \;\;+ \frac{C_{\gamma}L_{J}G_{\theta}^2}{1-v}T^{-v}
\end{split}
\end{equation*}
\end{proof}

\begin{lemma}\label{lm:a4} Let the cumulative error of linear differential Q-value function be $\sum_{t=0}^{T-1}\mathbb{E}||\Delta w_t||^2$, the cumulative error of target linear differential Q-value function be $\sum_{t=0}^{T-1}\mathbb{E}||\Delta \bar{w_t}||^2$, and the cumulative error of target average reward estimator be $\sum_{t=0}^{T-1}\mathbb{E}||\Delta \bar{\rho}_t||^2$. $w_t$, $\bar{w}_t$ and $\bar{\rho}_t$ are linear differential Q-value function parameter, target linear differential Q-value function parameter and target average reward estimator at time t respectively.
Bound on the cumulative error of differential Q-value function parameter is proven using cumulative error of target average reward estimator and target linear differential Q-value function parameter as follows:
\begin{equation*}
    \begin{split}
        \frac{1}{T}\sum_{t=0}^{T-1}\mathbb{E}||\Delta w_t||^2 &\leq 2\Bigg(\sqrt{\frac{2C_w^2}{(\lambda-1) C_\alpha}T^{\sigma-1}+\frac{C_gC_\alpha}{1-\sigma}T^{-\sigma}} +\frac{L_wG_\theta C_\gamma}{(\lambda-1)C_\alpha}\bigg(\frac{T^{-2(v-\sigma)}}{1-2(v-\sigma)}\bigg)^{\frac{1}{2}}\Bigg)^2\\
        &\quad +\frac{4}{(\lambda-1)^2}\Bigg(\frac{1}{T}\sum_{t=0}^{T-1}\mathbb{E}|\Delta\bar{\rho}_t|^2 + \frac{1}{T}\sum_{t=0}^{T-1}\mathbb{E}||\Delta \bar{w}_t||^2\Bigg)
    \end{split}
\end{equation*}
Here, $\Delta w_t = w_t - w_t^{*}$, $\Delta \bar{w}_t = \bar{w}_t - w_t^{*}$, and $\Delta\bar{\rho}_t = \bar{\rho}_t -\rho_t^{*}$. $w_t^{*}$ and $\rho_t^{*}$ are the optimal parameters given by TD(0) algorithm corresponding to policy parameter $\theta_t$. $C_\alpha$, $C_\gamma$, $\sigma$, $v$ are constants and $\gamma_t, \alpha_t$ are step-sizes  defined in Assumption \ref{as:2.1}, $\|w_t\| \leq C_w$ (Algorithm \ref{alg:2}, step 8), $C_r$ is the upper bound on rewards (Assumption \ref{as:4}), Constant $G_{\theta}$ is defined in  Lemma \ref{lm:a7}, $C_g = \frac{L_w^2}{(\lambda-1)}\max_t\frac{\gamma_t^2}{\alpha_t^2}G_\theta^2 + \frac{C_\delta^2}{(\lambda-1)}$, $C_\delta = 2C_r + (4+\eta)C_w$. $\eta$ is the l2-regularisation coefficient from Algorithm \ref{alg:2} and $\eta > \lambda_{max}^{all}$, where $\lambda_{max}^{all}$ is defined in Lemma \ref{lm:a11}. $\lambda$ is defined in Lemma \ref{lm:a11}. $L_w$ is defined in Lemma \ref{lm:a6}.        
\end{lemma}
\begin{proof}

\begin{equation}\label{eq:a8.2}
    \begin{split}
        w_{t+1} &= w_t + \alpha_t\frac{1}{M}\sum_{i=0}^{M-1}\Bigl(R^\pi(s_{t,i}) - \bar{\rho_t} + \phi^\pi(s_{t,i}')^{\intercal}\bar{w_t} - \phi^\pi(s_{t,i})^{\intercal}w_t\Bigr)\phi^\pi(s_{t,i}) - \alpha_t\eta w_t\\
        \implies w_{t+1} - w_{t+1}^* &= w_t - w_t^* + w_t^* - w_{t+1}^*  \term{1}\\
        &\quad +\alpha_t \frac{1}{M}\sum_{i=0}^{M-1}\Bigl(R^\pi(s_{t,i}) - \rho_t^* + \phi^\pi(s_{t,i}')^{\intercal}\bar{w_t} - \phi^\pi(s_{t,i})^{\intercal}w_t\Bigr)\phi^\pi(s_{t,i}) - \alpha_t\eta w_t \term{2}\\
        &\quad + \alpha_t\frac{1}{M}\sum_{i=0}^{M-1}\Bigl(\rho_t^* - \bar{\rho}_t\Bigr)\phi^\pi(s_{t,i}) \term{3}\\
    \end{split}
\end{equation}

From \eqref{eq:a8.2}:\\
\termw{2}:
\begin{equation}\label{eq:a8.3}
    \begin{split}
        \frac{1}{M}\sum_{i=0}^{M-1}\Bigl(R^\pi(s_{t,i}) &- \rho_t^* + \phi^\pi(s_{t,i}')^{\intercal}\bar{w_t} - \phi^\pi(s_{t,i})^{\intercal}w_t\Bigr)\phi^\pi(s_{t,i}) - \eta w_t\\
        =& \frac{1}{M}\sum_{i=0}^{M-1}\Bigl(R^\pi(s_{t,i}) - \rho_t^* + \phi^\pi(s_{t,i}')^{\intercal}w_t - \phi^\pi(s_{t,i})^{\intercal}w_t\Bigr)\phi^\pi(s_{t,i}) - \eta w_t \\ 
        &+\frac{1}{M}\sum_{i=0}^{M-1}\phi^\pi(s_{t,i})\phi^\pi(s_{t,i}')^{\intercal}(\bar{w_t}-w_t^*) - \frac{1}{M}\sum_{i=0}^{M-1}\phi^\pi(s_{t,i})\phi^\pi(s_{t,i}')^{\intercal}(w_t-w_t^*)\\
        =& \frac{1}{M}\sum_{i=0}^{M-1}\phi^\pi(s_{t,i})\phi^\pi(s_{t,i}')^{\intercal}(\bar{w_t}-w_t^*) - \frac{1}{M}\sum_{i=0}^{M-1}\phi^\pi(s_{t,i})\phi^\pi(s_{t,i}')^{\intercal}(w_t-w_t^*)\\ 
        &+ g(B_t, w_t, \theta_t) - \bar{g}(w_t, \theta_t) +\bar{g}(w_t, \theta_t) - \bar{g}(w_t^*, \theta_t)
    \end{split}
\end{equation}

Let $g(B_t, w_t, \theta_t):=\dfrac{1}{M}\sum_{i=0}^{M-1}\Bigl(R^\pi(s_{t,i}) - \rho_t^*\Bigr)\phi^\pi(s_{t,i}) + \dfrac{1}{M}\sum_{i=0}^{M-1}\Bigl(\phi^\pi(s_{t,i})(\phi^\pi(s_{t,i}') - \phi^\pi(s_{t,i}))^{\intercal} - \eta I\Bigr)w_t$

Let $\bar{g}(w_t, \theta_t) := \int d(s, \pi(\theta_t))\phi^\pi(s)\Bigl(R^\pi(s) - \rho_t^* + \int P^\pi(s'|s)\phi^\pi(s')^{\intercal}w_t\,ds' - \phi^\pi(s)^{\intercal}w_t\Bigr)\,ds - \eta w_t$

Here, $w_t^{*}$ is the differential Q-value function parameter such that $\bar{g}(w_t^{*},\theta_t) = 0$. The existence of $w_t^{*}$ is guaranteed by setting the value of $\eta$ according to Lemma \ref{lm:a11}.

Using \eqref{eq:a8.3} in \eqref{eq:a8.2}:
\begin{equation*}
\begin{split}
w_{t+1} - w_{t+1}^* = & w_t - w_t^* + w_t^* - w_{t + 1}^* + \\ 
&+\alpha_t\frac{1}{M}\sum_{i=0}^{M-1}(\rho_t^* - \bar{\rho}_t)\phi^\pi(s_{t,i})\\ 
&+ \alpha_t\frac{1}{M}\sum_{i=0}^{M-1}\phi^\pi(s_{t,i})\phi^\pi(s_{t,i}')^{\intercal}(\bar{w_t}-w_t^*)\\ 
&+\alpha_t\frac{1}{M}\sum_{i=0}^{M-1}\phi^\pi(s_{t,i})\phi^\pi(s_{t,i}')^{\intercal}(w_t^* - w_t)\\ 
&+ \alpha_t(g(B_t, w_t, \theta_t)-\bar{g}(w_t, \theta_t))\\ 
&+ \alpha_t(\bar{g}(w_t, \theta_t) - \bar{g}(w_t^*, \theta_t))
\end{split}
\end{equation*}

\begin{equation*}
    \begin{split}
        \text{Let,}\quad f(B_t, w_t, \theta_t) := &\frac{1}{M}\sum_{i=0}^{M-1}(\rho_t^* - \bar{\rho}_t)\phi^\pi(s_{t,i})\\ 
        &+ \frac{1}{M}\sum_{i=0}^{M-1}\phi^\pi(s_{t,i})\phi^\pi(s_{t,i}')^{\intercal}(\bar{w_t}-w_t^*)\\ 
        &+ \frac{1}{M}\sum_{i=0}^{M-1}\phi^\pi(s_{t,i})\phi^\pi(s_{t,i}')^{\intercal}(w_t^* - w_t)\\ 
        &+ (g(B_t, w_t, \theta_t)-\bar{g}(w_t, \theta_t))\\ 
        &+ (\bar{g}(w_t, \theta_t) - \bar{g}(w_t^*, \theta_t))
    \end{split}
\end{equation*}

\begin{equation*}
    \begin{split}
        || w_{t+1} - w_{t+1}^* ||^2 
        & = \|\Gamma_{C_w}\big(w_t + \alpha_tf(B_t, w_t, \theta_t)\big) - w_{t+1}^*\| \\
        & \leq \|w_t + \alpha_tf(B_t, w_t, \theta_t)) - w_{t+1}^*\| \quad(\because \text{Projection is a non-expansive operator}) \\
        &\leq ||(w_t - w_t^*) + (w_t^* - w_{t+1}^*) + \alpha_t f(B_t, w_t, \theta_t)||^2\\
        &\leq ||w_t - w_t^*||^2 + ||w_t^* - w_{t+1}^*||^2\\
        &\quad + \alpha_t^2||f(B_t, w_t, \theta_t)||^2\\
        &\quad + 2\langle\Delta w_t, w_t^* - w_{t+1}^*\rangle + 2\alpha_t\langle\Delta w_t, f(B_t, w_t, \theta_t)\rangle\\
        &\quad + 2\alpha_t\langle w_t^*-w_{t+1}^*, f(B_t, w_t, \theta_t)\rangle
    \end{split}
\end{equation*}

We know that $2x^{\intercal}y \leq \|x\|^2 + \|y\|^2$ and hence $2\alpha_t\langle w_t^*-w_{t+1}^*, f(B_t, w_t, \theta_t)\rangle \leq ||w_t^* - w_{t+1}^*||^2 + \alpha_t^2||f(B_t, w_t, \theta_t)||^2$. We get the following equation:

\begin{equation*}
    \begin{split}
        || w_{t+1} - w_{t+1}^* ||^2 
        &\leq ||w_t - w_t^*||^2 + 2||w_t^* - w_{t+1}^*||^2\\
        &\quad + 2\alpha_t^2||f(B_t, w_t, \theta_t)||^2\\
        &\quad + 2\langle\Delta w_t, w_t^* - w_{t+1}^*\rangle + 2\alpha_t\langle\Delta w_t, f(B_t, w_t, \theta_t)\rangle\\
    \end{split}
\end{equation*}

Apply expectation on both sides of the inequality we get :

\begin{equation*}
   \begin{split}
        \mathbb{E}||w_{t+1}-w_{t+1}^*||^2 &\leq \mathbb{E}||\Delta w_t||^2 + 2\mathbb{E}||w_t^*-w_{t+1}^*||^2\\
        &\quad + 2\alpha_t^2\mathbb{E}||f(B_t, w_t, \theta_t)||^2 \\
        &\quad + 2\mathbb{E}\langle\Delta w_t, w_t^* - w_{t+1}^* \rangle \\
        &\quad + 2\alpha_t\mathbb{E}\langle\Delta w_t, f(B_t, w_t, \theta_t)\rangle\\
    \end{split}
\end{equation*}
Expanding the definition of $f(B_t, w_t, \theta_t)$ we get the following:
\begin{equation}\label{eq:a8.4}
   \begin{split}
        \mathbb{E}||w_{t+1}-w_{t+1}^*||^2
        &\leq \mathbb{E}||\Delta w_t||^2 + 2\mathbb{E}||w_t^* - w_{t+1}^*||^2 \term{1}\\
        &\quad + 2\alpha_{t}^2\mathbb{E}||f(B_t, w_t, \theta_t)||^2 \term{2}\\
        &\quad + 2\mathbb{E}\langle\Delta w_t, w_t^* - w_{t+1}^*\rangle \term{3}\\
        &\quad + 2\alpha_t\mathbb{E}\langle\Delta w_t, \frac{1}{M}\sum_{i=0}^{M-1}( \bar{\rho_t} - \rho_t^*)\phi^\pi(s_{t,i})\rangle \term{4}\\
        &\quad+ 2\alpha_t\mathbb{E}\langle\Delta w_t, \frac{1}{M}\sum_{i=0}^{M-1}\phi^\pi(s_{t,i})\phi^\pi(s_{t,i}')^{\intercal}(\bar{w_t} - w_t^*)\rangle \term{5}\\
        &\quad+ 2\alpha_t\mathbb{E}\langle\Delta w_t, \frac{1}{M}\sum_{i=0}^{M-1}\phi^\pi(s_{t,i})\phi^\pi(s_{t,i}')^{\intercal}(w_t^* - w_t)\rangle \term{6}\\
        &\quad + 2\alpha_t\mathbb{E}\langle\Delta w_t, g(B_t, w_t, \theta_t) - \bar{g}(w_t, \theta_t)\rangle \term{7}\\
        &\quad + 2\alpha_t\mathbb{E}\langle\Delta w_t, \bar{g}(w_t, \theta_t) - \bar{g}(w_t^*, \theta_t)\rangle \term{8}
    \end{split}
\end{equation}

From \eqref{eq:a8.4}:\\
\termw{1}:
\begin{equation*}
    \begin{split}
    \mathbb{E}||w_t^* - w_{t+1}^*||^2 &\leq L_w^2\mathbb{E}||\theta_{t+1} - \theta_t||^2 \quad\text{(using Lemma \ref{lm:a6})}\\
    &\leq L_w^2\gamma_t^2G_\theta^2 \quad\text{(using Lemma \ref{lm:a7})}
    \end{split}
\end{equation*}
\termw{2}:
\begin{equation*}
    \begin{split}
    &\mathbb{E}||f(B_t, w_t, \theta_t)||^2\\
    &= \mathbb{E}||\frac{1}{M}\sum_{i=0}^{M-1}(R^\pi(s_{t,i})-\bar{\rho_t} + \phi^\pi(s_{t,i}')^{\intercal}\bar{w_t} - \phi^\pi(s_{t,i})^{\intercal}w_t)\phi^\pi(s_{t,i}) - \eta w_t||^2\\
    &\leq \mathbb{E}\big(||\frac{1}{M}\sum_{i=0}^{M-1}(R^\pi(s_{t,i}) - \bar{\rho_t} + \phi^\pi(s_{t,i}')^{\intercal}\bar{w_t} - \phi^\pi(s_{t,i})^{\intercal}w_t)\phi^\pi(s_{t,i})|| + \eta||w_t||\big)^2\\
    \end{split}
\end{equation*}

Here,
\begin{equation*}
    \begin{split}
        ||\phi^\pi(s)|| &< 1 \quad \text{(Assumption \ref{as:3})}\\
        |R^\pi(s)| &\leq C_r \quad\text{(Assumption \ref{as:4})}\\
        ||w_t|| &\leq C_w \quad\text{(Algorithm \ref{alg:2}, step 8)}\\
        |\rho_t| &\leq C_r + 2C_w \quad\text{(Lemma \ref{lm:a8})}\\
        ||\bar{w_t}|| &\leq C_w \quad \text{(Lemma \ref{lm:a9})}\\
        ||\bar{\rho_t}|| &\leq C_r + 2C_w \quad \text{(Lemma \ref{lm:a10})}\\
    \end{split}
\end{equation*}

\begin{equation*}
    \begin{split}
    &\leq \mathbb{E}\Big(\frac{1}{M}\sum_{i=0}^{M-1}||(R^\pi(s_{t,i}) - \bar{\rho_t}+ \phi^\pi(s_{t,i}')^{\intercal}\bar{w_t} - \phi^\pi(s_{t,i})^{\intercal} w_t)\phi^\pi(s_{t,i})|| + \eta||w_t||\Big)^2\\
    &\leq \mathbb{E}(C_r + C_r + 2C_w + 2C_w + \eta C_w)^2\\
    &\leq \mathbb{E}(C_\delta)^2 \quad ( C_\delta = 2C_r + (4+\eta)C_w )\\
    &\leq C_\delta^2 
    \end{split}
\end{equation*}

\termw{3}:
\begin{equation*}
    \begin{split}
    \mathbb{E}\langle\Delta w_t, w_t^* - w_{t+1}^*\rangle &\leq \mathbb{E}||\Delta w_t||\,||w_t^* - w_{t+1}^*||\\
    &\leq L_w\mathbb{E}||\Delta w_t||\,||\theta_{t+1} - \theta_t|| \quad(\text{using Lemma \ref{lm:a6}})
    \end{split}
\end{equation*}

\termw{4}:
\begin{equation*}
    \begin{split}
    \mathbb{E}[\langle\Delta w_t, \frac{1}{M}\sum_{i=0}^{M-1}(\rho_t^*-\bar{\rho}_t)\phi^\pi(s_{t,i})\rangle] &= \mathbb{E}[\frac{1}{M}\sum_{i=0}^{M-1}\langle\Delta w_t, \phi^\pi(s_{t,i}) \rangle (\rho_t^* - \bar{\rho}_t)]\\
    & \leq \mathbb{E} [\frac{1}{M}\sum_{i=0}^{M-1}||\Delta w_t|| ||\phi^\pi(s_{t,i})|| |(\rho_t^*- \bar{\rho}_t)|]\\
    & \leq \mathbb{E}||\Delta w_t|| |\rho_t^*- \bar{\rho}_t|\\
    &= \mathbb{E}||\Delta w_t|| |\Delta \bar{\rho}_t|
    \end{split}
\end{equation*}

\termw{5}:
\begin{equation*}
    \begin{split}
    \mathbb{E}\langle\Delta w_t, \frac{1}{M}\sum_{i=0}^{M-1}\phi^\pi(s_{t,i})\phi^\pi(s_{t,i}')^{\intercal}(\bar{w_t} - w_t^*)\rangle &\leq \mathbb{E}||\Delta w_t||\,||\frac{1}{M}\sum_{i=0}^{M-1}\phi^\pi(s_{t,i})\phi^\pi(s_{t,i}')^{\intercal}(\bar{w_t} - w_t^*)||\\
    &\leq \mathbb{E}||\Delta w_t||\,||\bar{w_t} - w_t^*|| \\
    &\leq \mathbb{E}||\Delta w_t||\,||\Delta\bar{w_t}|| \\
    \end{split}
\end{equation*}

\termw{6}:
\begin{equation*}
    \begin{split}
    \mathbb{E}\langle\Delta w_t, \frac{1}{M}\sum_{i=0}^{M-1}\phi^\pi(s_{t,i})\phi^\pi(s_{t,i}')^{\intercal}(w_t^* - w_t)\rangle &\leq \mathbb{E}||\Delta w_t||\,||\frac{1}{M}\sum_{i=0}^{M-1}\phi^\pi(s_{t,i})\phi^\pi(s_{t,i}')^{\intercal}(w_t^* - w_t)||\\
    &\leq \mathbb{E}||\Delta w_t||\,||w_t^* - w_t|| \\
    &\leq \mathbb{E}||\Delta w_t||^2
    \end{split}
\end{equation*}

\termw{7}:
\begin{equation*}
    \begin{split}
    &\mathbb{E}[\langle\Delta w_t, g(B_t, w_t, \theta_t) - \bar{g}(w_t, \theta_t)\rangle] = \mathbb{E}[\langle\Delta w_t,\mathbb{E}[g(B_t, w_t, \theta_t) - \bar{g}(w_t, \theta_t)|\Delta w_t]\rangle]\\
    &\text{Note:}\quad\mathbb{E}[g(B_t, w_t, \theta_t) - \bar{g}(w_t, \theta_t)]  = 0\\
    &\text{Hence,}\quad\mathbb{E}[\langle\Delta w_t, g(B_t, w_t, \theta_t) - \bar{g}(w_t, \theta_t)\rangle] = 0
    \end{split}
\end{equation*}

\termw{8}:
\begin{equation*}
    \begin{split}
        \mathbb{E}[\langle\Delta w_t, \bar{g}(w_t, \theta_t) - \bar{g}(w_t^*, \theta_t)\rangle]\\
        A(\theta_t) &= \int_{S} d^\pi(s, \theta_t)\big(\phi^\pi(s)(\mathbb{E}[\phi^\pi(s')] - \phi^\pi(s))^{\intercal} - \eta I\big)ds\\
        b(\theta_t) &= \int_{S} d^\pi(s, \theta_t) r^\pi(s) \phi^\pi(s)ds\\
        \bar{g}(w_t, \theta_t) - \bar{g} (w_t^*, \theta_t) &= b(\theta_t)+A(\theta_t)w_t - b(\theta_t) - A(\theta_t)w_t^*\\
        &=A(\theta_t)(w_t - w_t^*)\\
        \text{Now,}\quad\mathbb{E}[\langle\Delta w_t, \bar{g}(w_t, \theta_t) - \bar{g}(w_t^*, \theta_t)\rangle] &= \mathbb{E}[\langle\Delta w_t, A(\theta_t)\Delta w_t\rangle]\\
        &= \mathbb{E}[\Delta w_t^{\intercal} A(\theta_t)\Delta w_t]\\
        &\leq -\lambda \mathbb{E}||\Delta w_t||^2\qquad\text{(Lemma \ref{lm:a11})}
    \end{split}
\end{equation*}

Combining \termw{1} - \termw{8} into \eqref{eq:a8.4}:
\begin{equation*}
    \begin{split}
        \mathbb{E}||w_{t+1} - w_{t+1}^*||^2 &\leq (1 - 2(\lambda-1)\alpha_t)\mathbb{E}||\Delta w_t||^2 + 2L_w^2\gamma_t^2G_\theta^2 + 2\alpha_t^2C_\delta^2\\
        &\quad + 2L_w\mathbb{E}||\Delta w_t|| ||\theta_{t + 1} - \theta_t|| + 2\alpha_t\mathbb{E}||\Delta w_t|| |\Delta\bar{\rho}_t|\\
        &\quad + 2\alpha_t\mathbb{E}||\Delta w_t||||\Delta \bar{w}_t||  \\
    \end{split}
\end{equation*}

\begin{equation*}
    \begin{split}
        \implies 2(\lambda-1)\alpha_t\mathbb{E}||\Delta w_t||^2 &\leq\mathbb{E}[||\Delta w_t||^2] - \mathbb{E}||\Delta w_{t+1}||^2 + 2L_w^2\gamma_t^2G_\theta^2 + 2\alpha_t^2C_\delta^2\\
        &\quad + 2L_w\gamma_tG_\theta\mathbb{E}||\Delta w_t|| + 2\alpha_t\mathbb{E}||\Delta w_t|| |\Delta\bar{\rho}_t|\\
        &\quad + 2\alpha_t\mathbb{E}||\Delta w_t||||\Delta \bar{w}_t||\\
    \end{split}
\end{equation*}
\begin{equation*}
    \begin{split}
        \implies \mathbb{E}||\Delta w_t||^2 &\leq \frac{1}{2(\lambda-1)\alpha_t}(\mathbb{E}||\Delta w_t||^2 - \mathbb{E}|| w_{t+1}||^2)\\
        &\quad + \Big(\frac{L_w^2\gamma_t^2}{(\lambda-1)\alpha_t}G_\theta^2 + \frac{\alpha_t}{(\lambda-1)}C_\delta^2\Big)\\
        &\quad + \frac{L_w}{(\lambda-1)}\frac{\gamma_t}{\alpha_t}G_\theta\mathbb{E}||\Delta w_t||\\
        &\quad + \frac{\mathbb{E}||\Delta w_t|| |\Delta\bar{\rho}_t|}{(\lambda-1)}\\
        &\quad + \frac{\mathbb{E}||\Delta w_t||||\Delta \bar{w}_t||}{(\lambda-1)}\\
    \end{split}
\end{equation*}
\begin{equation}\label{eq:a8.5}
    \begin{split}
        \implies \sum_{t=0}^{T-1}\mathbb{E}||\Delta w_t||^2 &\leq \sum_{t=0}^{T-1}\frac{1}{2(\lambda-1)\alpha_t}(\mathbb{E}||\Delta w_t||^2 - \mathbb{E}||\Delta w_{t+1}||^2) \term{1}\\
        &\quad + \sum_{t=0}^{T-1}\Big(\frac{L_w}{(\lambda-1)}\frac{\gamma_t^2}{\alpha_t}G_\theta^2 + \frac{\alpha_t}{(\lambda-1)}C_\delta^2\Big) \term{2}\\
        &\quad + \sum_{t=0}^{T-1}\frac{L_w}{(\lambda-1)}\frac{\gamma_t}{\alpha_t}G_\theta\mathbb{E}||\Delta w_t|| \term{3}\\
        &\quad + \sum_{t=0}^{T-1}\frac{\mathbb{E}||\Delta w_t|| \big(|\Delta \bar{\rho}_t|+||\Delta \bar{w}_t||\big)}{(\lambda-1)} \term{4}\\
    \end{split}
\end{equation}

From \eqref{eq:a8.5}:\\
\termw{1}:
\begin{equation*}
    \begin{split}
        \frac{1}{2(\lambda-1)}\sum_{t=0}^{T-1}(\mathbb{E}||\Delta w_t||^2 - \mathbb{E}||\Delta w_{t+1}||^2)\frac{1}{\alpha_t} =& \frac{1}{2(\lambda-1)}\Big(\sum_{t=1}^{T-1}\Big(\frac{1}{\alpha_t}-\frac{1}{\alpha_{t-1}}\Big)\mathbb{E}||\Delta w_t||^2 \\
        &+ \frac{1}{\alpha_0}\mathbb{E}||\Delta w_0||^2 - \frac{1}{\alpha_{T-1}}\mathbb{E}||\Delta w_T||^2\Big)\\
        &\leq \frac{1}{2(\lambda-1)}\Bigg(\sum_{t=1}^{T-1}\Big(\frac{1}{\alpha_t} - \frac{1}{\alpha_{t-1}}\Big)+\frac{1}{\alpha_0}\Bigg)4C_w^2\\
        &\leq \frac{4C_w^2}{2(\lambda-1)\alpha_{T-1}} = \frac{C_w^2}{(\lambda-1) C_\alpha}T^\sigma \quad\text{$(\because\alpha_t = \frac{C_\alpha}{(1 + t)^\alpha})$}
    \end{split}
\end{equation*}

\termw{2}:
\begin{equation*}
    \begin{split}
        \sum_{t=0}^{T-1}\Big(\frac{L_w^2}{(\lambda-1)}\frac{\gamma_t^2}{\alpha_t}G_\theta^2 + \frac{\alpha_t}{(\lambda-1)}C_\delta^2\Big) &= \sum_{t=0}^{T-1}\Big(\frac{L_w^2}{(\lambda-1)}\frac{\gamma_t^2}{\alpha_t^2}G_\theta^2 + \frac{C_\delta^2}{(\lambda-1)}\Big)\alpha_t\\
        &\leq \sum_{t=0}^{T-1}\Big(\frac{L_w^2}{(\lambda-1)}\max_t\frac{\gamma_t^2}{\alpha_t^2}G_\theta^2 + \frac{C_\delta^2}{(\lambda-1)}\Big)\alpha_t\\
        &= \sum_{t=0}^{T-1}C_g\alpha_t = \frac{C_gC_\alpha}{1 - \sigma}T^{1 - \sigma} \quad \Big(C_g = \frac{L_w^2}{(\lambda-1)}\max_t\frac{\gamma_t^2}{\alpha_t^2}G_\theta^2 + \frac{C_\delta^2}{(\lambda-1)}\Big)
    \end{split}
\end{equation*}

\termw{3}:
\begin{equation*}
    \begin{split}
        \sum_{t=0}^{T-1}\frac{L_w}{(\lambda-1)}\frac{\gamma_t}{\alpha_t}G_\theta\mathbb{E}||\Delta w_t|| &= \frac{L_w}{(\lambda-1)}G_\theta\sum_{t=0}^{T-1}\frac{\gamma_t}{\alpha_t}\mathbb{E}||\Delta w_t||\\
        &\leq \frac{L_w}{(\lambda-1)}G_\theta\Bigg(\sum_{t=0}^{T-1}\Big(\frac{\gamma_t}{\alpha_t}\Big)^2\Bigg)^{\frac{1}{2}}\Big(\sum_{t=0}^{T-1}(\mathbb{E}||\Delta w_t||)^2\Big)^{\frac{1}{2}}\\&(\text{Using Cauchy Schwartz inequality})\\
        &\leq\frac{L_w}{(\lambda-1)}G_\theta\Bigg(\sum_{t=0}^{T-1}\Big(\frac{\gamma_t}{\alpha_t}\Big)^2\Bigg)^{\frac{1}{2}}\Big(\sum_{t=0}^{T-1}\mathbb{E}||\Delta w_t||^2\Big)^{\frac{1}{2}}\\&(\text{Using Jensen's inequality})\\
        &\leq\frac{L_wG_\theta C_\gamma}{(\lambda-1)C_\alpha}\Bigg(\frac{T^{1-2(v-\sigma)}}{1-2(v-\sigma)}{}\Bigg)^{\frac{1}{2}}\Big(\sum_{t=0}^{T-1}\mathbb{E}||\Delta w_t||^2\Big)^{\frac{1}{2}}
    \end{split}
\end{equation*}

\termw{4}:
\begin{equation*}
    \begin{split}
        \frac{1}{(\lambda-1)}\sum_{t=0}^{T-1}\mathbb{E}||\Delta w_t||( |\Delta \bar{\rho}_t|+\|\Delta\bar{w}_t\|) &\leq \frac{1}{(\lambda-1)}\Big(\sum_{t=0}^{T-1}(\mathbb{E}||\Delta w_t||)^2\Big)^{\frac{1}{2}}\Big(\sum_{t=0}^{T-1}\big(\mathbb{E}(|\Delta\bar{\rho}_t|+||\Delta \bar{w}_t||)\big)^2\Big)^{\frac{1}{2}}\\
        &\leq \frac{1}{(\lambda-1)}\Big(\sum_{t=0}^{T-1}(\mathbb{E}||\Delta w_t||)^2\Big)^{\frac{1}{2}}\Big(\sum_{t=0}^{T-1}\mathbb{E}(|\Delta\bar{\rho}_t|+||\Delta \bar{w}_t||)^2\Big)^{\frac{1}{2}}\\
        &\leq \frac{1}{(\lambda-1)}\Big(\sum_{t=0}^{T-1}(\mathbb{E}||\Delta w_t||)^2\Big)^{\frac{1}{2}}\Big(2\sum_{t=0}^{T-1}\mathbb{E}(|\Delta\bar{\rho}_t|^2+||\Delta \bar{w}_t||^2)\Big)^{\frac{1}{2}}\\
    \end{split}
\end{equation*}

Combining \termw{1} - \termw{5} into \eqref{eq:a8.5} and dividing by T:
\begin{equation*}
    \begin{split}
       \frac{1}{T} \sum_{t=0}^{T-1}\mathbb{E}||\Delta w_t||^2 \leq& \frac{2C_w^2}{(\lambda-1) C_\alpha}T^{\sigma-1} +\frac{C_gC^\alpha}{1-\sigma}T^{-\sigma}\\
        &+\frac{L_wG_\theta C_\gamma}{(\lambda-1)C_\alpha}\Bigg(\frac{T^{-2(v-\sigma)}}{1-2(v-\sigma)}\Bigg)^{\frac{1}{2}}\Big(\frac{1}{T}\sum_{t=0}^{T-1}\mathbb{E}||\Delta w_t||^2\Big)^{\frac{1}{2}}\\
        &+\frac{2^{1/2}}{(\lambda-1)}\Big(\frac{1}{T}\sum_{t=0}^{T-1}(\mathbb{E}||\Delta w_t||)^2\Big)^{\frac{1}{2}}\Big(\frac{1}{T}\sum_{t=0}^{T-1}\mathbb{E}\big(|\Delta\bar{\rho}_t|^2+||\Delta \bar{w}_t||^2\big)\Big)^{\frac{1}{2}}\\    \end{split}
\end{equation*}

Let,
\begin{equation*}
    \begin{split}
        M(T) &= \frac{1}{T}\sum_{t=0}^{T-1}\mathbb{E}||\Delta w_t||^2\\
        N(T) &= \frac{1}{T}\sum_{t=0}^{T-1}\mathbb{E}|\Delta\bar{\rho}_t|^2 + \frac{1}{T}\sum_{t=0}^{T-1}\mathbb{E}||\Delta \bar{w}_t||^2 
    \end{split}
\end{equation*}

\begin{equation*}
    \begin{split}
        M(T) &\leq K_1 + K_2\sqrt{M(T)} + K_3\sqrt{M(T)}\sqrt{N(T)}\\
        K_1 &:= \frac{2C_w^2}{(\lambda-1) C_\alpha}T^{\sigma-1} + \frac{C_gC_\alpha}{1-\sigma}T^{-\sigma}\\
        K_2 &:= \frac{L_wG_\theta C_\gamma}{(\lambda-1)C_\alpha}\Bigg(\frac{T^{-2(v-\sigma)}}{1-2(v-\sigma)}\Bigg)^{\frac{1}{2}}\\
        K_3 &:= \frac{2^{1/2}}{\lambda-1}
    \end{split}
\end{equation*}

\begin{equation*}
    \begin{split}
        &M(T) - 2\frac{K_2}{2}\sqrt{M(T)}-2\frac{K_3}{2}\sqrt{M(T)}\sqrt{N(T)} +2\frac{K_2}{2}\frac{K_3}{2}\sqrt{N(T)}\\ & + \Big(\frac{K_2}{2}\Big)^2 +\Big(\frac{K_3}{2}\sqrt{N(T)}\Big)^2 \leq K_1 + \Big(\frac{K_2}{2}\Big)^2 + \Big(\frac{K_3}{2}\sqrt{N(T)}\Big)^2 + 2\frac{K_2}{2}\frac{K_3}{2}\sqrt{N(T)}
    \end{split}
\end{equation*}
\begin{equation*}
    \begin{split}
        \implies \Bigl(\sqrt{M(T)} - \frac{K_2}{2} - \frac{K_3}{2}\sqrt{N(T)}\Bigr)^2 &\leq K_1 + \Big(\frac{K_2}{2} + \frac{K_3}{2}\sqrt{N(T)}\Big)^2\\
        \implies \sqrt{M(T)} - \frac{K_2}{2} - \frac{K_3}{2}\sqrt{N(T)} &\leq\sqrt{K_1}+\frac{K_2}{2} + \frac{K_3}{2}\sqrt{N(T)}\\
        \implies \sqrt{M(T)} &\leq \sqrt{K_1} + K_2 + K_3\sqrt{N(T)}\\
        \implies M(T) &\leq 2(\sqrt{K_1} + K_2)^2 + 2K_3^2N(T)
    \end{split}
\end{equation*}

\begin{equation*}
    \begin{split}
        \frac{1}{T}\sum_{t=0}^{T-1}\mathbb{E}||\Delta w_t||^2 &\leq 2\Bigg(\sqrt{\frac{2C_w^2}{(\lambda-1) C_\alpha}T^{\sigma-1}+\frac{C_gC_\alpha}{1-\sigma}T^{-\sigma}} +\frac{L_wG_\theta C_\gamma}{(\lambda-1)C_\alpha}\bigg(\frac{T^{-2(v-\sigma)}}{1-2(v-\sigma)}\bigg)^{\frac{1}{2}}\Bigg)^2\\
        &\quad +\frac{4}{(\lambda-1)^2}\Bigg(\frac{1}{T}\sum_{t=0}^{T-1}\mathbb{E}|\Delta\bar{\rho}_t|^2 + \frac{1}{T}\sum_{t=0}^{T-1}\mathbb{E}||\Delta \bar{w}_t||^2\Bigg)
    \end{split}
\end{equation*}
\end{proof}

\begin{lemma}\label{lm:a4.8}
Let the cumulative error of target linear differential Q-value function parameter be $\sum_{t=0}^{T-1}\mathbb{E}||\Delta \bar{w}_t||^2$ and cumulative error of linear differential Q-value function parameter be $\sum_{t=0}^{T-1}\mathbb{E}||\Delta w_t||^2$. $\bar{w}_t$ and $w_t$ are target linear differential Q-value function parameter and linear differential Q-value function parameter at time t respectively.
Bound on the cumulative error of target linear differential Q-value function parameter is proven using cumulative error of linear differential Q-value function parameter as follows:
    \begin{equation*}
        \begin{split}
            \frac{1}{T}\sum_{t=0}^{T-1}\mathbb{E}\|\Delta\bar{w}_t\|^2&\leq2\Bigg(\sqrt{\frac{2C_w^2T^{u-1}}{C_\beta}+\frac{C_{gt}C_\beta T^{-u}}{1-u}}\\
            &\quad +L_w G_\theta C_\gamma\Big(\frac{T^{-v}}{(1-2v)^{1/2}} + \frac{T^{-(v-u)}}{C_\beta(1-2(v-u))^{1/2}}\Big)\Bigg)^2\\
            &\quad +\frac{2}{T}\sum_{t=0}^{T-1}\mathbb{E}||\Delta w_{t+1}||^2
        \end{split}
    \end{equation*}
Here, $\Delta w_t = w_t -w_t^{*}$, $\Delta \bar{w}_t = \bar{w}_t - w_t^{*}$. $\bar{w}_t^{*}$ and $w_t^{*}$ are the optimal parameters given by TD(0) algorithm corresponding to policy parameter $\theta_t$. $C_\beta$, $C_\gamma$, $u$, and $v$ are constants defined in Assumption \ref{as:2.1}, $\|w_t\| \leq C_w$ (Algorithm \ref{alg:2}, step 8), $C_r$ is the upper bound on rewards (Assumption \ref{as:4}), Constant $G_{\theta}$ is defined in  Lemma \ref{lm:a7}. $C_{gt} = L_w^2G_\theta^2\max_t(\gamma_t^2/\beta_t^2)+4C_w^2$. $L_w$ is Lipchitz constant defined in Lemma \ref{lm:a6}. 
\end{lemma}
\begin{proof}
\begin{equation*}
\begin{split}
\bar{w}_{t+1} =& \bar{w}_{t} + \beta_t(w_{t+1} - \bar{w}_{t}) \\
\implies \bar{w}_{t+1} - w_{t+1}^* =& \bar{w}_{t} - w_{t}^* +w_{t}^* - w_{t+1}^*  + \beta_t(w_{t+1} - \bar{w}_{t}) \\
\implies \|\Delta \bar{w}_{t+1}\|^2 =& \|\Delta\bar{w}_{t} +w_{t}^* - w_{t+1}^*  + \beta_t(w_{t+1} - \bar{w}_{t})\|^2 \\
\leq & \|\Delta\bar{w}_{t}\|^2 + 2\|w_{t}^* - w_{t+1}^*\|^2 + 2\|\beta_t(w_{t+1} - \bar{w}_{t})\|^2 \\
&+ 2\beta_t\langle\Delta\bar{w}_{t},w_{t+1} - \bar{w}_{t}\rangle + 2\langle\Delta\bar{w}_{t},w_{t}^* - w_{t+1}^*\rangle \\
=&\|\Delta\bar{w}_{t}\|^2 + 2\|w_{t}^* - w_{t+1}^*\|^2 + 2\|\beta_t(w_{t+1} - \bar{w}_{t})\|^2 \\
&+ 2\beta_t\langle\Delta\bar{w}_{t},\Delta w_{t+1} - \Delta\bar{w}_{t}\rangle + 2\beta_t\langle\Delta\bar{w}_{t},w_{t+1}^* - w_{t}^*\rangle \\
&+ 2\langle\Delta\bar{w}_{t},w_{t}^* - w_{t+1}^*\rangle \\
\implies \|\Delta \bar{w}_{t+1}\|^2 \leq & (1-2\beta_t)\|\Delta\bar{w}_{t}\|^2 + 2\|w_{t}^* - w_{t+1}^*\|^2 + 2\|\beta_t(w_{t+1} - \bar{w}_{t})\|^2 \\
&+ 2\beta_t\langle\Delta\bar{w}_{t},\Delta w_{t+1}\rangle + 2\beta_t\langle\Delta\bar{w}_{t},w_{t+1}^* - w_{t}^*\rangle \\
&+ 2\langle\Delta\bar{w}_{t},w_{t}^* - w_{t+1}^*\rangle \\
\end{split}
\end{equation*}

\begin{equation*}
\begin{split}
\implies \|\Delta \bar{w}_{t}\|^2 =& \frac{1}{2\beta_t}\big(\|\Delta\bar{w}_{t}\|^2 -\|\Delta\bar{w}_{t+1}\|^2 \big) + \big(\frac{1}{\beta_t} \|w_{t}^* - w_{t+1}^*\|^2 + \beta_t\|(w_{t+1} - \bar{w}_{t})\|^2 \big) \\
&+ \langle\Delta\bar{w}_{t},\Delta w_{t+1}\rangle + \langle\Delta\bar{w}_{t},w_{t+1}^* - w_{t}^*\rangle + \frac{1}{\beta_t}\langle\Delta\bar{w}_{t},w_{t}^* - w_{t+1}^*\rangle  \\
\end{split}
\end{equation*}

\begin{equation}\label{eq:a8.511}
\begin{split}
\implies \sum_{t=0}^{T-1} \mathbb{E}\|\Delta \bar{w}_{t}\|^2 =& \sum_{t=0}^{T-1} \frac{1}{2\beta_t}\big(\mathbb{E}\|\Delta\bar{w}_{t}\|^2 -\mathbb{E}\|\Delta\bar{w}_{t+1}\|^2 \big) \term{1}\\
&+\sum_{t=0}^{T-1}\big(\frac{1}{\beta_t} \mathbb{E}\|w_{t}^* - w_{t+1}^*\|^2 + \beta_t\mathbb{E}\|(w_{t+1} - \bar{w}_{t})\|^2 \big) \term{2} \\
&+ \sum_{t=0}^{T-1}\mathbb{E}\langle\Delta\bar{w}_{t},\Delta w_{t+1}\rangle \term{3} \\
&+ \sum_{t=0}^{T-1}\mathbb{E}\langle\Delta\bar{w}_{t},w_{t+1}^* - w_{t}^*\rangle \term{4} \\
&+ \sum_{t=0}^{T-1}\frac{1}{\beta_t}\mathbb{E}\langle\Delta\bar{w}_{t},w_{t}^* - w_{t+1}^*\rangle \term{5} \\
\end{split}
\end{equation}

From \ref{eq:a8.511}:

\termw{1}:
\begin{equation*}
\begin{split}
&\sum_{t=0}^{T-1} \frac{1}{2\beta_t}\big(\mathbb{E}\|\Delta\bar{w}_{t}\|^2 -\mathbb{E}\|\Delta\bar{w}_{t+1}\|^2 \big)\\
&= \frac{1}{2}\Big(\sum_{t=1}^{T-1}\big(\frac{1}{\beta_{t}} - \frac{1}{\beta_{t-1}}\big)\mathbb{E}\|\Delta\bar{w}_{t}\|^2 + \frac{1}{\beta_0}\mathbb{E}\|\Delta\bar{w}_{0}\|^2 -\frac{1}{\beta_{T-1}}\mathbb{E}\|\Delta\bar{w}_{T}\|^2 \Big) \\
&\leq\frac{1}{2}\Big(\sum_{t=1}^{T-1}\big(\frac{1}{\beta_{t}} - \frac{1}{\beta_{t-1}}\big)\mathbb{E}\|\Delta\bar{w}_{t}\|^2 + \frac{1}{\beta_0}\mathbb{E}\|\Delta\bar{w}_{0}\|^2 \Big)\\
&\leq\frac{1}{2}\Big(\sum_{t=1}^{T-1}\big(\frac{1}{\beta_{t}} - \frac{1}{\beta_{t-1}}\big) + \frac{1}{\beta_0}\Big)4C_w^2 \quad(\text{Using Lemma \ref{lm:a10}})\\
&=\frac{2C_w^2}{\beta_{T-1}} = \frac{2C_w^2T^{u}}{C_\beta} 
\end{split}
\end{equation*}

\termw{2}:
\begin{equation*}
\begin{split}
&\sum_{t=0}^{T-1}\big(\frac{1}{\beta_t} \mathbb{E}\|w_{t}^* - w_{t+1}^*\|^2 + \beta_t\mathbb{E}\|(w_{t+1} - \bar{w}_{t})\|^2 \big) \\ 
& \leq \sum_{t=0}^{T-1}\big(\frac{L_{w}^2}{\beta_t} \mathbb{E}\|\theta_{t} - \theta_{t+1}\|^2 + \beta_t\mathbb{E}\|(w_{t+1} - \bar{w}_{t})\|^2 \big) \quad(\text{Using Lemma \ref{lm:a6} })\\
& \leq \sum_{t=0}^{T-1}\big(L_{w}^2G_{\theta}^2\frac{\gamma_t^2}{\beta_t} + 4\beta_tC_w^2 \big) \quad(\text{Using Lemma \ref{lm:a7},\ref{lm:a8}, and \ref{lm:a11} })\\
& \leq \sum_{t=0}^{T-1}\big(L_{w}^2G_{\theta}^2\max_t\frac{\gamma_t^2}{\beta_t^2} + 4C_w^2 \big)\beta_t \\
& = \sum_{t=0}^{T-1} C_{gt}\beta_t = \frac{C_{gt}C_\beta T^{1-u}}{1-u} \quad(C_{gt} = L_{w}^2G_{\theta}^2\max_t\frac{\gamma_t^2}{\beta_t^2} + 4C_w^2)
\end{split}
\end{equation*}

\termw{3}:
\begin{equation*}
\begin{split}
&\sum_{t=0}^{T-1}\mathbb{E}\langle\Delta\bar{w}_{t},\Delta w_{t+1}\rangle \\
&\leq \Big(\sum_{t=0}^{T-1}\mathbb{E}\|\Delta\bar{w}_{t}\|^2\Big)^{1/2}\Big(\sum_{t=0}^{T-1}\mathbb{E}\|\Delta w_{t+1}\|^2\Big)^{1/2} \quad(\text{Using Cauchy-Schwarz inequality})
\end{split}
\end{equation*}

\termw{4}:
\begin{equation*}
\begin{split}
&\sum_{t=0}^{T-1}\mathbb{E}\langle\Delta\bar{w}_{t},w_{t+1}^* - w_{t}^*\rangle \\
&\leq \sum_{t=0}^{T-1}\mathbb{E}\|\Delta\bar{w}_{t}\|\|w_{t+1}^* - w_{t}^*\| \\
&\leq \sum_{t=0}^{T-1}L_wG_{\theta}\gamma_t\mathbb{E}\|\Delta\bar{w}_{t}\| \quad (\text{Using Lemma \ref{lm:a6},\ref{lm:a7}})\\
&\leq L_wG_{\theta}\Big(\sum_{t=0}^{T-1}\gamma_t^2\Big)^{1/2}\Big(\sum_{t=0}^{T-1}\mathbb{E}\|\Delta\bar{w}_{t}\|^2\Big)^{1/2} \quad (\text{Using Cauchy-Schwarz inequality})\\
&\leq L_wG_{\theta}C_{\gamma}\frac{T^{-v}}{(1-2v)^2}T^{1/2}\Big(\sum_{t=0}^{T-1}\mathbb{E}\|\Delta\bar{w}_{t}\|^2\Big)^{1/2}
\end{split}
\end{equation*}

\termw{5}:
\begin{equation*}
\begin{split}
&\sum_{t=0}^{T-1}\frac{1}{\beta_t}\mathbb{E}\langle\Delta\bar{w}_{t},w_{t+1}^* - w_{t}^*\rangle \\
&\leq \sum_{t=0}^{T-1}\frac{1}{\beta_t}\mathbb{E}\|\Delta\bar{w}_{t}\|\|w_{t+1}^* - w_{t}^*\| \\
&\leq \sum_{t=0}^{T-1}L_wG_{\theta}\frac{\gamma_t}{\beta_t}\mathbb{E}\|\Delta\bar{w}_{t}\| \quad (\text{Using Lemma \ref{lm:a6},\ref{lm:a7}})\\
&\leq L_wG_{\theta}\Big(\sum_{t=0}^{T-1}\frac{\gamma_t^2}{\beta_t^2}\Big)^{1/2}\Big(\sum_{t=0}^{T-1}\mathbb{E}\|\Delta\bar{w}_{t}\|^2\Big)^{1/2} \quad (\text{Using Cauchy-Schwarz inequality})\\
&\leq \frac{L_wG_{\theta}C_{\gamma}}{C_{\beta}}\frac{T^{-(v-u)}}{(1-2(v-u))^2}T^{1/2}\Big(\sum_{t=0}^{T-1}\mathbb{E}\|\Delta\bar{w}_{t}\|^2\Big)^{1/2}
\end{split}
\end{equation*}
Combining \termw{1}-\termw{5} into \ref{eq:a8.511}: 

\begin{equation}
\begin{split}
\frac{1}{T}\sum_{t=0}^{T-1} \mathbb{E}\|\Delta \bar{w}_{t}\|^2 =& \frac{2C_w^2T^{u-1}}{C_\beta} + \frac{C_{gt}C_\beta T^{-u}}{1-u}\\
&+\Big(\frac{1}{T}\sum_{t=0}^{T-1}\mathbb{E}\|\Delta\bar{w}_{t}\|^2\Big)^{1/2}\Big(\frac{1}{T}\sum_{t=0}^{T-1}\mathbb{E}\|\Delta w_{t+1}\|^2\Big)^{1/2}\\
&+L_wG_{\theta}C_{\gamma}\frac{T^{-v}}{(1-2v)^2}\Big(\frac{1}{T}\sum_{t=0}^{T-1}\mathbb{E}\|\Delta\bar{w}_{t}\|^2\Big)^{1/2}\\
&+ \frac{L_wG_{\theta}C_{\gamma}}{C_{\beta}}\frac{T^{-(v-u)}}{(1-2(v-u))^2}\Big(\frac{1}{T}\sum_{t=0}^{T-1}\mathbb{E}\|\Delta\bar{w}_{t}\|^2\Big)^{1/2}\\
\end{split}
\end{equation}

\begin{equation*}
\begin{split}
M(T) &= \frac{1}{T}\sum_{t=0}^{T-1} \mathbb{E}\|\Delta \bar{w}_{t}\|^2 \\
N(T) &= \frac{1}{T}\sum_{t=0}^{T-1} \mathbb{E}\|\Delta w_{t+1}\|^2 \\
M(T) &\leq K_1 + K_2\sqrt{M(T)} + K_3\sqrt{M(T)N(T)}
\end{split}    
\end{equation*}

Here,
\begin{equation*}
\begin{split}
K_1 &= \frac{2C_w^2T^{u-1}}{C_\beta} + \frac{C_{gt}C_\beta T^{1-u}}{1-u}\\
K_2 &= L_wG_{\theta}C_{\gamma}\frac{T^{-v}}{(1-2v)^2} + \frac{L_wG_{\theta}C_{\gamma}}{C_{\beta}}\frac{T^{-(v-u)}}{(1-2(v-u))^2} \\
K_3 &= 1
\end{split}
\end{equation*}

From Lemma \ref{lm:a4}, we know that
        \begin{equation*}
            \begin{split}
                M(T) &\leq 2(\sqrt{K_1}+K_2)^2+2K_3^2N(T)\\
            \end{split}
        \end{equation*}

Hence,
\begin{equation}
\begin{split}
\frac{1}{T}\sum_{t=0}^{T-1} \mathbb{E}\|\Delta \bar{w}_{t}\|^2 \leq& 2\Bigg(\sqrt{\frac{2C_w^2T^{u-1}}{C_\beta} + \frac{C_{gt}C_\beta T^{1-u}}{1-u}}\\ 
&+L_wG_{\theta}C_{\gamma}\frac{T^{-v}}{(1-2v)^2} + \frac{L_wG_{\theta}C_{\gamma}}{C_{\beta}}\frac{T^{-(v-u)}}{(1-2(v-u))^2}\Bigg)^2\\
&+\frac{2}{T}\sum_{t=0}^{T-1}\mathbb{E}\|\Delta w_{t+1}\|^2
\end{split}
\end{equation}

\end{proof}

\begin{lemma}\label{lm:a4.9}
Let the cumulative error of target average reward estimator be $\sum_{t=0}^{T-1}\mathbb{E}||\Delta \bar{\rho}_t||^2$ and cumulative error of average reward estimator be $\sum_{t=0}^{T-1}\mathbb{E}||\Delta \rho_t||^2$. $\bar{\rho}_t$ and $\rho_t$ are target average reward estimator and average reward estimator at time t respectively.
Bound on the cumulative error of target average reward estimator is proven using cumulative error of average reward estimator as follows:
    \begin{equation*}
        \begin{split}
            \frac{1}{T}\sum_{t=0}^{T-1}\mathbb{E}\|\Delta\bar{\rho}_t\|^2&\leq2\Bigg(\sqrt{\frac{2(C_r+C_w)^2T^{u-1}}{C_\beta}+\frac{C_{st}C_\beta T^{-u}}{1-u}}\\
            &\quad +L_p G_\theta C_\gamma\Big(\frac{T^{-v}}{(1-2v)^{1/2}} + \frac{T^{-(v-u)}}{C_\beta(1-2(v-u))^{1/2}}\Big)\Bigg)^2\\
            &\quad +\frac{2}{T}\sum_{t=0}^{T-1}\mathbb{E}||\Delta \rho_{t+1}||^2
        \end{split}
    \end{equation*}
Here, $\Delta \rho_t = \rho_t -\rho_t^{*}$, $\Delta \bar{\rho}_t = \bar{\rho}_t - \rho_t^{*}$. $\bar{\rho}_t^{*}$ and $\rho_t^{*}$ are the optimal parameters given by TD(0) algorithm corresponding to policy parameter $\theta_t$. $C_\beta$, $C_\gamma$, $u$, and $v$ are constants defined in Assumption \ref{as:2.1}, $\|w_t\| \leq C_w$ (Algorithm \ref{alg:2}, step 8), $C_r$ is the upper bound on rewards (Assumption \ref{as:4}), Constant $G_{\theta}$ is defined in  Lemma \ref{lm:a7}. $C_{st} = L_p^2G_\theta^2\max_t(\gamma_t^2/\beta_t^2)+4(C_r+2C_w)^2$. $L_p$ is Lipchitz constant defined in Lemma \ref{lm:a14}. 
\end{lemma}
\begin{proof}
\begin{equation*}
\begin{split}
\bar{\rho}_{t+1} =& \bar{\rho}_{t} + \beta_t(\rho_{t+1} - \bar{\rho}_{t}) \\
\implies \bar{\rho}_{t+1} - \rho_{t+1}^* =& \bar{\rho}_{t} - \rho_{t}^* +\rho_{t}^* - \rho_{t+1}^*  + \beta_t(\rho_{t+1} - \bar{\rho}_{t}) \\
\implies \|\Delta \bar{\rho}_{t+1}\|^2 =& \|\Delta\bar{\rho}_{t} +\rho_{t}^* - \rho_{t+1}^*  + \beta_t(\rho_{t+1} - \bar{\rho}_{t})\|^2 \\
\leq & \|\Delta\bar{\rho}_{t}\|^2 + 2\|\rho_{t}^* - \rho_{t+1}^*\|^2 + 2\|\beta_t(\rho_{t+1} - \bar{\rho}_{t})\|^2 \\
&+ 2\beta\langle\Delta\bar{\rho}_{t},\rho_{t+1} - \bar{\rho}_{t}\rangle + 2\langle\Delta\bar{\rho}_{t},\rho_{t}^* - \rho_{t+1}^*\rangle \\
=&\|\Delta\bar{\rho}_{t}\|^2 + 2\|\rho_{t}^* - \rho_{t+1}^*\|^2 + 2\|\beta_t(\rho_{t+1} - \bar{\rho}_{t})\|^2 \\
&+ 2\beta_t\langle\Delta\bar{\rho}_{t},\Delta \rho_{t+1} - \Delta\bar{\rho}_{t}\rangle + 2\beta_t\langle\Delta\bar{\rho}_{t},\rho_{t+1}^* - \rho_{t}^*\rangle \\
&+ 2\langle\Delta\bar{\rho}_{t},\rho_{t}^* - \rho_{t+1}^*\rangle \\
\implies \|\Delta \bar{\rho}_{t+1}\|^2 \leq & (1-2\beta_t)\|\Delta\bar{\rho}_{t}\|^2 + 2\|\rho_{t}^* - \rho_{t+1}^*\|^2 + 2\|\beta_t(\rho_{t+1} - \bar{\rho}_{t})\|^2 \\
&+ 2\beta_t\langle\Delta\bar{\rho}_{t},\Delta \rho_{t+1}\rangle + 2\beta_t\langle\Delta\bar{\rho}_{t},\rho_{t+1}^* - \rho_{t}^*\rangle \\
&+ 2\langle\Delta\bar{\rho}_{t},\rho_{t}^* - \rho_{t+1}^*\rangle \\
\end{split}
\end{equation*}

\begin{equation*}
\begin{split}
\implies \|\Delta \bar{\rho}_{t}\|^2 =& \frac{1}{2\beta_t}\big(\|\Delta\bar{\rho}_{t}\|^2 -\|\Delta\bar{\rho}_{t+1}\|^2 \big) + \big(\frac{1}{\beta_t} \|\rho_{t}^* - \rho_{t+1}^*\|^2 + \beta_t\|(\rho_{t+1} - \bar{\rho}_{t})\|^2 \big) \\
&+ \langle\Delta\bar{\rho}_{t},\Delta \rho_{t+1}\rangle + \langle\Delta\bar{\rho}_{t},\rho_{t+1}^* - \rho_{t}^*\rangle + \frac{1}{\beta_t}\langle\Delta\bar{\rho}_{t},\rho_{t}^* - \rho_{t+1}^*\rangle  \\
\end{split}
\end{equation*}

\begin{equation}\label{eq:a8.51}
\begin{split}
\implies \sum_{t=0}^{T-1} \mathbb{E}\|\Delta \bar{\rho}_{t}\|^2 =& \sum_{t=0}^{T-1} \frac{1}{2\beta_t}\big(\mathbb{E}\|\Delta\bar{\rho}_{t}\|^2 -\mathbb{E}\|\Delta\bar{\rho}_{t+1}\|^2 \big) \term{1}\\
&+\sum_{t=0}^{T-1}\big(\frac{1}{\beta_t} \mathbb{E}\|\rho_{t}^* - \rho_{t+1}^*\|^2 + \beta_t\mathbb{E}\|(\rho_{t+1} - \bar{\rho}_{t})\|^2 \big) \term{2} \\
&+ \sum_{t=0}^{T-1}\mathbb{E}\langle\Delta\bar{\rho}_{t},\Delta \rho_{t+1}\rangle \term{3} \\
&+ \sum_{t=0}^{T-1}\mathbb{E}\langle\Delta\bar{\rho}_{t},\rho_{t+1}^* - \rho_{t}^*\rangle \term{4} \\
&+ \sum_{t=0}^{T-1}\frac{1}{\beta_t}\mathbb{E}\langle\Delta\bar{\rho}_{t},\rho_{t}^* - \rho_{t+1}^*\rangle \term{5} \\
\end{split}
\end{equation}

From \ref{eq:a8.51}:

\termw{1}:
\begin{equation*}
\begin{split}
&\sum_{t=0}^{T-1} \frac{1}{2\beta_t}\big(\mathbb{E}\|\Delta\bar{\rho}_{t}\|^2 -\mathbb{E}\|\Delta\bar{\rho}_{t+1}\|^2 \big)\\
&= \frac{1}{2}\Big(\sum_{t=1}^{T-1}\big(\frac{1}{\beta_{t}} - \frac{1}{\beta_{t-1}}\big)\mathbb{E}\|\Delta\bar{\rho}_{t}\|^2 + \frac{1}{\beta_0}\mathbb{E}\|\Delta\bar{\rho}_{0}\|^2 -\frac{1}{\beta_{T-1}}\mathbb{E}\|\Delta\bar{\rho}_{T}\|^2 \Big) \\
&\leq\frac{1}{2}\Big(\sum_{t=1}^{T-1}\big(\frac{1}{\beta_{t}} - \frac{1}{\beta_{t-1}}\big)\mathbb{E}\|\Delta\bar{\rho}_{t}\|^2 + \frac{1}{\beta_0}\mathbb{E}\|\Delta\bar{\rho}_{0}\|^2 \Big)\\
&\leq\frac{1}{2}\Big(\sum_{t=1}^{T-1}\big(\frac{1}{\beta_{t}} - \frac{1}{\beta_{t-1}}\big) + \frac{1}{\beta_0}\Big)4(C_r + C_w)^2 \\
&\text{( Using Lemma \ref{lm:a10} and Assumption \ref{as:4}, we have $|\rho_t|\leq C_r + 2C_w $and $|\rho_t^{*}|\leq C_r$ )}\\
&\\
&=\frac{2(C_r+C_w)^2}{\beta_{T-1}} = \frac{2(C_r+C_w)^2T^{u}}{C_\beta} 
\end{split}
\end{equation*}

\termw{2}:
\begin{equation*}
\begin{split}
&\sum_{t=0}^{T-1}\big(\frac{1}{\beta_t} \mathbb{E}\|\rho_{t}^* - \rho_{t+1}^*\|^2 + \beta_t\mathbb{E}\|(\rho_{t+1} - \bar{\rho}_{t})\|^2 \big) \\ 
& \leq \sum_{t=0}^{T-1}\big(\frac{L_{p}^2}{\beta_t} \mathbb{E}\|\theta_{t} - \theta_{t+1}\|^2 + \beta_t\mathbb{E}\|(\rho_{t+1} - \bar{\rho}_{t})\|^2 \big) \quad(\text{Using Lemma \ref{lm:a14} })\\
& \leq \sum_{t=0}^{T-1}\big(L_{p}^2G_{\theta}^2\frac{\gamma_t^2}{\beta_t} + \beta_t4(C_r + 2C_w)^2 \big) \quad(\text{Using Lemma \ref{lm:a7}, \ref{lm:a8}, and \ref{lm:a11} })\\
& \leq \sum_{t=0}^{T-1}\big(L_{p}^2G_{\theta}^2\max_t\frac{\gamma_t^2}{\beta_t^2} + 4(C_r + 2C_w)^2 \big)\beta_t \\
& = \sum_{t=0}^{T-1} C_{st}\beta_t = \frac{C_{st}C_\beta T^{1-u}}{1-u} \quad(C_{st} = L_{p}^2G_{\theta}^2\max_t\frac{\gamma_t^2}{\beta_t^2} + 4(C_r + 2C_w)^2)
\end{split}
\end{equation*}

\termw{3}:
\begin{equation*}
\begin{split}
&\sum_{t=0}^{T-1}\mathbb{E}\langle\Delta\bar{\rho}_{t},\Delta \rho_{t+1}\rangle \\
&\leq \Big(\sum_{t=0}^{T-1}\mathbb{E}\|\Delta\bar{\rho}_{t}\|^2\Big)^{1/2}\Big(\sum_{t=0}^{T-1}\mathbb{E}\|\Delta \rho_{t+1}\|^2\Big)^{1/2} \\
&\text{( Using Cauchy-Schwarz inequality and Jensen's Inequality )}
\end{split}
\end{equation*}

\termw{4}:
\begin{equation*}
\begin{split}
&\sum_{t=0}^{T-1}\mathbb{E}\langle\Delta\bar{\rho}_{t},\rho_{t+1}^* - \rho_{t}^*\rangle \\
&\leq \sum_{t=0}^{T-1}\mathbb{E}\|\Delta\bar{\rho}_{t}\|\|\rho_{t+1}^* - \rho_{t}^*\| \\
&\leq \sum_{t=0}^{T-1}L_pG_{\theta}\gamma_t\mathbb{E}\|\Delta\bar{\rho}_{t}\| \quad (\text{Using Lemma \ref{lm:a7}, \ref{lm:a14}})\\
&\leq L_pG_{\theta}\Big(\sum_{t=0}^{T-1}\gamma_t^2\Big)^{1/2}\Big(\sum_{t=0}^{T-1}\mathbb{E}\|\Delta\bar{\rho}_{t}\|^2\Big)^{1/2} \quad (\text{Using Cauchy-Schwarz and Jensen's inequality})\\
&\leq L_pG_{\theta}C_{\gamma}\frac{T^{-v}}{(1-2v)^2}T^{1/2}\Big(\sum_{t=0}^{T-1}\mathbb{E}\|\Delta\bar{\rho}_{t}\|^2\Big)^{1/2}
\end{split}
\end{equation*}

\termw{5}:
\begin{equation*}
\begin{split}
&\sum_{t=0}^{T-1}\frac{1}{\beta_t}\mathbb{E}\langle\Delta\bar{\rho}_{t},\rho_{t+1}^* - \rho_{t}^*\rangle \\
&\leq \sum_{t=0}^{T-1}\frac{1}{\beta_t}\mathbb{E}\|\Delta\bar{\rho}_{t}\|\|\rho_{t+1}^* - \rho_{t}^*\| \\
&\leq \sum_{t=0}^{T-1}L_pG_{\theta}\frac{\gamma_t}{\beta_t}\mathbb{E}\|\Delta\bar{\rho}_{t}\| \quad (\text{Using Lemma \ref{lm:a7},\ref{lm:a14}})\\
&\leq L_pG_{\theta}\Big(\sum_{t=0}^{T-1}\frac{\gamma_t^2}{\beta_t^2}\Big)^{1/2}\Big(\sum_{t=0}^{T-1}\mathbb{E}\|\Delta\bar{\rho}_{t}\|^2\Big)^{1/2} \quad (\text{Using Cauchy-Schwarz and Jensen's inequality})\\
&\leq \frac{L_pG_{\theta}C_{\gamma}}{C_{\beta}}\frac{T^{-(v-u)}}{(1-2(v-u))^2}T^{1/2}\Big(\sum_{t=0}^{T-1}\mathbb{E}\|\Delta\bar{\rho}_{t}\|^2\Big)^{1/2}
\end{split}
\end{equation*}
Combining \termw{1}-\termw{5} into \ref{eq:a8.51}: 

\begin{equation}\label{eq:a8.52}
\begin{split}
\frac{1}{T}\sum_{t=0}^{T-1} \mathbb{E}\|\Delta \bar{\rho}_{t}\|^2 =& \frac{2(C_r+C_w)^2T^{u-1}}{C_\beta} + \frac{C_{st}C_\beta T^{-u}}{1-u}\\
&+\Big(\frac{1}{T}\sum_{t=0}^{T-1}\mathbb{E}\|\Delta\bar{\rho}_{t}\|^2\Big)^{1/2}\Big(\frac{1}{T}\sum_{t=0}^{T-1}\mathbb{E}\|\Delta \rho_{t+1}\|^2\Big)^{1/2}\\
&+L_pG_{\theta}C_{\gamma}\frac{T^{-v}}{(1-2v)^2}\Big(\frac{1}{T}\sum_{t=0}^{T-1}\mathbb{E}\|\Delta\bar{\rho}_{t}\|^2\Big)^{1/2}\\
&+ \frac{L_pG_{\theta}C_{\gamma}}{C_{\beta}}\frac{T^{-(v-u)}}{(1-2(v-u))^2}\Big(\frac{1}{T}\sum_{t=0}^{T-1}\mathbb{E}\|\Delta\bar{\rho}_{t}\|^2\Big)^{1/2}\\
\end{split}
\end{equation}

\begin{equation*}
\begin{split}
M(T) &= \frac{1}{T}\sum_{t=0}^{T-1} \mathbb{E}\|\Delta \bar{\rho}_{t}\|^2 \\
N(T) &= \frac{1}{T}\sum_{t=0}^{T-1} \mathbb{E}\|\Delta \rho_{t+1}\|^2 \\
M(T) &\leq K_1 + K_2\sqrt{M(T)} + K_3\sqrt{M(T)N(T)}
\end{split}    
\end{equation*}

Here,
\begin{equation*}
\begin{split}
K_1 &= \frac{2(C_r+C_w)^2T^{u-1}}{C_\beta} + \frac{C_{st}C_\beta T^{1-u}}{1-u}\\
K_2 &= L_pG_{\theta}C_{\gamma}\frac{T^{-v}}{(1-2v)^2} + \frac{L_pG_{\theta}C_{\gamma}}{C_{\beta}}\frac{T^{-(v-u)}}{(1-2(v-u))^2} \\
K_3 &= 1
\end{split}
\end{equation*}

From Lemma \ref{lm:a4}, we know that
        \begin{equation*}
            \begin{split}
                M(T) &\leq 2(\sqrt{K_1}+K_2)^2+2K_3^2N(T)\\
            \end{split}
        \end{equation*}

Hence,
\begin{equation}
\begin{split}
\frac{1}{T}\sum_{t=0}^{T-1} \mathbb{E}\|\Delta \bar{\rho}_{t}\|^2 \leq& 2\Bigg(\sqrt{\frac{2(C_r+C_w)^2T^{u-1}}{C_\beta} + \frac{C_{st}C_\beta T^{1-u}}{1-u}}\\ 
&+L_pG_{\theta}C_{\gamma}\frac{T^{-v}}{(1-2v)^2} + \frac{L_pG_{\theta}C_{\gamma}}{C_{\beta}}\frac{T^{-(v-u)}}{(1-2(v-u))^2}\Bigg)^2\\
&+\frac{2}{T}\sum_{t=0}^{T-1}\mathbb{E}\|\Delta \rho_{t+1}\|^2
\end{split}
\end{equation}

\end{proof}

    \begin{lemma}\label{lm:a5}
    Let the cumulative error of average reward estimator be $\sum_{t=0}^{T-1}\mathbb{E}||\Delta \rho_t||^2$ and cumulative error of target linear differential Q-value function be $\sum_{t=0}^{T-1}\mathbb{E}||\Delta \bar{w}_t||^2$. $\bar{w}_t$ and $\rho_t$ are the target linear differential Q-value function parameter and average reward estimator at time t respectively. Bound on the cumulative error of average reward estimator is proven using cumulative error of target differential Q-value function as follows:
        \begin{equation*}
            \begin{split}
                \frac{1}{T}\sum_{t=0}^{T-1}\mathbb{E}|\Delta\rho_t|^2&\leq 2\Bigg(\sqrt{\frac{2(C_r+C_w)^2}{C_\alpha}T^{\sigma-1}+\frac{C_sC_\alpha }{1-\sigma}T^{-\sigma}} +\frac{L_p G_\theta C_\gamma}{C_\alpha}\Big(\frac{T^{-2(v-\sigma)}}{1-2(v-\sigma)}\Big)^{1/2}\Bigg)^2\\
                &\quad+8\frac{1}{T}\sum_{t=0}^{T-1}\mathbb{E}||\Delta \bar{w}_t||^2
            \end{split}
        \end{equation*}
    Here, $\Delta \rho_t = \rho_t -\rho_t^{*}$, $\Delta \bar{w}_t = \bar{w}_t - w_t^{*}$. $w_t^{*}$ and $\rho_t^{*}$ are the optimal parameters given by TD(0) algorithm corresponding to policy parameter $\theta_t$. $C_\alpha$, $\sigma$ are constants and $\gamma_t, \alpha_t$ are step-sizes  defined in Assumption \ref{as:2.1}, $\|w_t\| \leq C_w$ (Algorithm \ref{alg:2}, step 8), $C_r$ is the upper bound on rewards (Assumption \ref{as:4}), Constant $G_{\theta}$ is defined in  Lemma \ref{lm:a7}. $C_s = L_p^2G_\theta^2\max_t\frac{\gamma_t^2}{\alpha_t^2}+4(C_r+C_w)^2$. $L_p$ is Lipchitz constant defined in Lemma \ref{lm:a14}. 

    \end{lemma}
    \begin{proof}
        \begin{equation*}
            \begin{split}
                \rho_{t+1} &= \rho_t+\alpha_t\frac{1}{M}\sum_{i=0}^{M-1}\Big(R^\pi(s_{t,i})-\rho_t+\phi^\pi(s_{t,i}')^{\intercal}\bar{w_t}-\phi^\pi(s_{t,i})^\intercal\bar{w_t}\Big)\\
                \rho_{t+1}-\rho_{t+1}^*&=\rho_t-\rho_t^*+\rho_t^*-\rho_{t+1}^*\\
                &\quad+\alpha_t\frac{1}{M}\sum_{i=0}^{M-1}\Big(R^\pi(s_{t,i})-\rho_t+\phi^\pi(s_{t,i}')^\intercal\bar{w_t}-\phi^\pi(s_{t,i})^\intercal\bar{w_t}\Big)\\
                &= \rho_t-\rho_t^*+\rho_t^*-\rho_{t+1}^*\\
                &\quad+\alpha_t\frac{1}{M}\sum_{i=0}^{M-1}\Big(R^\pi(s_{t,i})-\rho_t^*+\phi^\pi(s_{t,i}')^\intercal\bar{w_t}-\phi^\pi(s_{t,i})^\intercal\bar{w_t}\Big)\\
                &\quad+\alpha_t(\rho_t^*-\rho_t)\\
                \rho_{t+1}-\rho_{t+1}^*&=\rho_t-\rho_t^*+\rho_t^*-\rho_{t+1}^*\\
                &\quad+\alpha_t(\rho_t^*-\rho_t)\\
                &\quad+\alpha_t\Big(\frac{1}{M}\sum_{i=0}^{M-1}(\phi^\pi(s_{t,i}')-\phi^\pi(s_{t,i}))^\intercal(\bar{w_t}-w_t^*)\Big)\\
                &\quad+\alpha_t\Big(\frac{1}{M}\sum_{i=0}^{M-1}(R^\pi(s_{t,i})-\rho_t^*+\phi^\pi(s_{t,i}')^\intercal w_t^*-\phi^\pi(s_{t,i})^\intercal w_t^*)\Big)\\
            \end{split}
        \end{equation*}
Let, $l(B_t, w_t, \theta_t) := \frac{1}{M}\sum_{i=0}^{M-1}\Big(R^\pi(s_{t,i})-\rho_t^*+\phi^\pi(s_{t,i}')^\intercal w_t-\phi^\pi(s_{t,i})^\intercal w_t\Big)$. We get the following:     
        \begin{equation*}
            \begin{split}
                \rho_{t+1}-\rho_{t+1}^*
                &=\rho_t-\rho_t^*+\rho_t^*-\rho_{t+1}^*\\
                &\quad+\alpha_t(\rho_t^*-\rho_t)\\
                &\quad+\alpha_t\Big(\frac{1}{M}\sum_{i=0}^{M-1}(\phi^\pi(s_{t,i}')-\phi^\pi(s_{t,i}))^\intercal(\bar{w_t}-w_t^*)\Big)\\
                &\quad+\alpha_t(l(B_t,w_t^*,\theta_t) - \bar{l}(w_t^{*},\theta_t))\quad \text{( $\bar{l}(w_t^{*}, \theta_t) =0$, as explained below)} \\
                &=\rho_t-\rho_t^* + \rho_t^* - \rho_{t+1}^* +\alpha_tl(B_t, \rho_t, w_t^*, \theta_t)
            \end{split}
        \end{equation*}
Here,
        \begin{equation*}
            \begin{split}
                \bar{l}(w_t, \theta_t)&:=\int_{S} d^\pi\big(s, \pi(\theta_t)\big)\Big(R^\pi(s)-\rho(\pi(\theta_t))+\int_{S}P^{\pi}(s'|s)\phi^\pi(s')^\intercal w_t\;ds' -\phi^\pi(s)^\intercal w_t\Big)\,ds\\
                l(B_t, \rho_t, w_t, \theta_t)&:=(\rho_t^*-\rho_t) + \Big(\frac{1}{M}\sum_{i=0}^{M-1}(\phi^\pi(s_{t,i}')-\phi^\pi(s_{t,i}))^\intercal(\bar{w_t}-w_t)\Big) + l(B_t,w_t^*,\theta_t) - \bar{l}(w_t^{*},\theta_t) \\
                & = \frac{1}{M}\sum_{i=0}^{M-1}\Big(R^\pi(s_{t,i})\rho_t+\phi^\pi(s_{t,i}')^{\intercal}\bar{w_t}-\phi^\pi(s_{t,i})^\intercal\bar{w_t}\Big)
            \end{split}
        \end{equation*}
Note: $\bar{l}(w_t, \theta_t) = \int_{S} d^\pi\big(s, \pi(\theta_t)\big)R^\pi(s)ds -\rho(\pi(\theta_t))+\int_{S} d^\pi\big(s, \pi(\theta_t)\big)\int_{S}P^{\pi}(s'|s)\phi^\pi(s')^\intercal w_t ds'ds -\int_{S} d^\pi\big(s, \pi(\theta_t)\big)\phi^\pi(s)^\intercal w_t\,ds = \rho(\pi(\theta_t)) - \rho(\pi(\theta_t)) + \int_{S} d^\pi\big(s', \pi(\theta_t)\big)\phi^\pi(s')^\intercal w_t\,ds' - \int_{S} d^\pi\big(s, \pi(\theta_t)\big)\phi^\pi(s)^\intercal w_t\,ds = 0$. Hence in the above equation we were able to add $\bar{l}(w_t^{*}, \theta_t)$.

        \begin{equation*}
            \begin{split}
                ||\Delta \rho_{t+1}||^2&=||\Delta\rho_t+\rho_t^*-\rho_{t+1}^{*}+\alpha_t l(B_t,w_t^*,\rho_t,\theta_t)||^2\\
                &=||\Delta\rho_t||^2+||\rho_t^*-\rho_{t+1}^*||^2+\alpha_t^2||l(B_t,w_t^*,\rho_t,\theta_t)||^2\\
                &\quad+2\langle\Delta\rho_t,\rho_t^*-\rho_{t+1}^*\rangle\\
                &\quad+2\alpha_t\langle\Delta\rho_t,l(B_t,w_t^*,\rho_t,\theta_t)\rangle\\
                &\quad+2\alpha_t\langle\rho_t^*-\rho_{t+1}^*,l(B_t,\rho_t,w_t^*,\theta_t)\rangle\\
                &\leq||\Delta\rho_t||^2+2||\rho_t^*-\rho_{t+1}^*||^2+2\alpha_t^2||l(B_t,w_t^*,\rho_t,\theta_t)||^2\\
                &\quad+2\langle\Delta\rho_t,\rho_t^*-\rho_{t+1}^*\rangle\\
                &\quad+2\alpha_t\langle\Delta\rho_t,l(B_t,w_t^*,\rho_t,\theta_t)\rangle\\
            \end{split}
        \end{equation*}
Expanding the definition of $l(B_t, w_t^{*},\rho_t, \theta_t)$ and taking expectation on both sides we get the following:
        \begin{equation}\label{eq:a8.6}
            \begin{split}
                \mathbb{E}||\Delta\rho_{t+1}||^2&\leq\mathbb{E}||\Delta\rho_t||^2+2\mathbb{E}||\rho_t^*-\rho_{t+1}^*||^2 \term{1}\\
                &\quad+2\alpha_t^2\mathbb{E}||l(B_t,w_t^*,\rho_t,\theta_t)||^2 \term{2}\\
                &\quad+2\mathbb{E}\langle\Delta\rho_t,\rho_t^*-\rho_{t+1}^*\rangle \term{3}\\
                &\quad+2\alpha_t\mathbb{E}\langle\Delta\rho_t,-\Delta\rho_t\rangle \term{4}\\
                &\quad+2\alpha_t\mathbb{E}\langle\Delta\rho_t,\frac{1}{M}\sum_{i=0}^{M-1}(\phi^\pi(s_{t,i}')-\phi^\pi(s_{t,i}))^\intercal(\bar{w_t}-w_t^*)\rangle \term{5}\\
                &\quad+2\alpha_t\mathbb{E}\langle\Delta\rho_t,l(B_t, w_t^*, \theta_t)-\bar{l}(w_t^*,\theta_t)\rangle \term{6}
            \end{split}
        \end{equation}

From \eqref{eq:a8.6}:\\
\termw{1}:
        \begin{equation*}
            \begin{split}
                \mathbb{E}||\rho_t^*-\rho_{t+1}^*||^2 &\leq L_p^2\mathbb{E}||\theta_{t+1}-\theta_t||^2\text{(Lemma \ref{lm:a14})} \\
                &\leq L_p^2\gamma_t^2G_{\theta}^2 \quad(\text{Using Lemma \ref{lm:a7}})
            \end{split}
        \end{equation*}
\termw{2}:
        \begin{equation*}
            \begin{split}
                \mathbb{E}||l(B_t,\rho_t,\bar{w_t},\theta_t)||^2&=\mathbb{E}||\frac{1}{M}\sum_{i=0}^{M-1}\big(R^\pi(s_{t,i})-\rho_t+\big(\phi^\pi(s_{t,i}')-\phi^\pi(s_{t,i})\big)^\intercal\bar{w_t}\big)||^2\\
                &\leq\mathbb{E}\Big(\frac{1}{M}\sum_{i=0}^{M-1}(C_r+C_r+2C_w)\Big)^2\\
                &= 4(C_r+C_w)^2
            \end{split}
        \end{equation*}
\termw{3}:
        \begin{equation*}
            \begin{split}
                \mathbb{E}\langle\Delta\rho_t,\rho_t^*-\rho_{t+1}^*\rangle&\leq\mathbb{E}||\Delta\rho_t||\,|\rho_t^*-\rho_{t+1}^*|\\
                &\leq L_p\mathbb{E}|\Delta\rho_t|\,||\theta_{t+1}-\theta_t||\quad\text{(Using Lemma \ref{lm:a14})} \\
                &\leq L_p\gamma_tG_{\theta}\mathbb{E}|\Delta\rho_t| \quad(\text{Using Lemma \ref{lm:a7})} 
            \end{split}
        \end{equation*}
\termw{4}:
        \begin{equation*}
            \begin{split}
                \mathbb{E}\langle\Delta\rho_t,-\Delta\rho_t\rangle=-\mathbb{E}|\Delta\rho_t|^2
            \end{split}
        \end{equation*}
\termw{5}:
        \begin{equation*}
            \begin{split}
                &\mathbb{E}\langle\Delta\rho_t,\frac{1}{M}\sum_{i=0}^{M-1}\big(\phi^\pi(s_{t,i}')^\intercal-\phi^\pi(s_{t,i})^\intercal\big)(\bar{w_t}-w_t^*)\rangle\\
                &\quad\leq\mathbb{E}\Big[\frac{1}{M}\sum_{i=0}^{M-1}||\phi^\pi(s_{t,i}')-\phi^\pi(s_{t,i})||\,||\bar{w_t}-w_t^*||\,|\Delta\rho_t|\Big]\\
                &\quad\leq 2\mathbb{E}|\Delta\rho_t|\|\Delta \bar{w}_t\|
            \end{split}
        \end{equation*}
\termw{6}:
        \begin{equation*}
            \begin{split}
                \mathbb{E}\langle\Delta\rho_t,l(B_t, w_t^{*}, \theta_t) -\bar{l}(w_t^*,\theta_t)\rangle&= \mathbb{E}\langle\Delta\rho_t,\mathbb{E}[l(B_t, w_t^{*}, \theta_t) -\bar{l}(w_t^*,\theta_t)|\Delta\rho_t]\rangle \\
                &=0 \\
                \text{Note:} \mathbb{E}[l(B_t, w_t^{*}, \theta_t) -\bar{l}(w_t^*,\theta_t)|\Delta\rho_t] &=0
            \end{split}
        \end{equation*}

Combining \termw{1}-\termw{6} into \eqref{eq:a8.6}:

        \begin{equation*}
            \begin{split}
                \mathbb{E}||\Delta\rho_{t+1}||^2&\leq(1-2\alpha_t)\mathbb{E}||\Delta\rho_t||^2+2L_p^2\gamma_t^2G_{\theta}^2\\
                &\quad+8\alpha_t^2(C_r+C_w)^2 + 2L_p\gamma_tG_{\theta}\mathbb{E}|\Delta\rho_t|\\
                &\quad+4\alpha_t\mathbb{E}|\Delta\rho_t|\|\Delta \bar{w}_t\|
            \end{split}
        \end{equation*}

        \begin{equation}\label{eq:a8.7}
            \begin{split}
                \implies \sum_{t=0}^{T-1}\mathbb{E}||\Delta\rho_t||^2 &\leq\sum_{t=0}^{T-1}\frac{1}{2\alpha_t}\Big(\mathbb{E}||\Delta\rho_t||^2-\mathbb{E}||\Delta\rho_{t+1}||^2\Big) \term{1}\\
                &\quad+\sum_{t=0}^{T-1}\Big(\frac{L_p^2\gamma_t^2}{\alpha_t}G_\theta^2+4\alpha_t(C_r+C_w)^2\Big)\term{2}\\
                &\quad+\sum_{t=0}^{T-1}\Big(L_pG_\theta\frac{\gamma_t}{\alpha_t}\Big)\mathbb{E}|\Delta\rho_t|\term{3}\\
                &\quad+\sum_{t=0}^{T-1}2\mathbb{E}||\Delta \bar{w}_t||\,|\Delta\rho_t| \term{4}
            \end{split}
        \end{equation}
From \eqref{eq:a8.7}:\\
\termw{1}:
        \begin{equation*}
            \begin{split}
                \frac{1}{2}\sum_{t=0}^{T-1}\frac{1}{\alpha_t}(\mathbb{E}||\Delta\rho_t||^2-\mathbb{E}||\Delta\rho_{t+1}||^2)&=\frac{1}{2}\Bigg(\sum_{t=0}^{T-1}\Big(\frac{1}{\alpha_t}-\frac{1}{\alpha_{t-1}}\Big)\mathbb{E}|\Delta\rho_t|^2+\frac{1}{\alpha_0}\mathbb{E}|\Delta\rho_0|^2-\frac{1}{\alpha_{T-1}}\mathbb{E}|\Delta\rho_t|^2\Bigg)\\
                &\leq\frac{1}{2}\Bigg(\sum_{t=0}^{T-1}\Big(\frac{1}{\alpha_t}-\frac{1}{\alpha_{t-1}}\Big)+\frac{1}{\alpha_0}\Bigg)4(C_r+C_w)^2\\
                &\leq\frac{2(C_r+C_w)^2}{C_\alpha}T^\sigma
            \end{split}
        \end{equation*}
\termw{2}:
        \begin{equation*}
            \begin{split}
                \sum_{t=0}^{T-1}\Big(L_p^2G_\theta^2\frac{\gamma_t^2}{\alpha_t}+4\alpha_t(C_r+C_w)^2\Big)&\leq\sum_{t=0}^{T-1}\Big(L_p^2G_\theta^2\max_t\frac{\gamma_t^2}{\alpha_t^2}+4(C_r+C_w)^2\Big)\alpha_t\\
                &\leq\sum_{t=0}^{T-1}C_s\alpha_t \quad (C_s = L_p^2G_\theta^2\max_t\frac{\gamma_t^2}{\alpha_t^2}+4(C_r+C_w)^2)\\
                &\leq\frac{C_sC_\alpha}{1-\sigma}T^{1-\sigma}
            \end{split}
        \end{equation*}
\termw{3}:
        \begin{equation*}
            \begin{split}
                \sum_{t=0}^{T-1}\Big(L_p G_\theta\frac{\gamma_t}{\alpha_t}\Big)\mathbb{E}||\Delta\rho_t||&=\sum_{t=0}^{T-1}L_p G_\theta\frac{\gamma_t}{\alpha_t}\mathbb{E}||\Delta\rho_t||\\
                &\leq L_p G_\theta\Bigg(\sum_{t=0}^{T-1}\Big(\frac{\gamma_t}{\alpha_t}\Big)^2\Bigg)^{1/2}\Big(\sum_{t=0}^{T-1}\mathbb{E}|\Delta\rho_t|^2\Big)^{1/2}\\
                &\leq \frac{L_p G_\theta C_\gamma}{C_\alpha}\Big(\frac{T^{1-2(v-\sigma)}}{1-2(v-\sigma)}\Big)^{1/2}\Big(\sum_{t=0}^{T-1}\mathbb{E}|\Delta\rho_t|^2\Big)^{1/2}\\
                &(\text{using Cauchy Schwarz and Jensen's inequality})
            \end{split}
        \end{equation*}
\termw{4}:
        \begin{equation*}
            \begin{split}
                2\sum_{t-0}^{T-1}\mathbb{E}||\Delta \bar{w}_t||\,|\Delta\rho_t|&\leq2(\sum_{t=0}^{T-1}\mathbb{E}||\Delta \bar{w}_t||^2)^{1/2}(\sum_{t=0}^{T-1}\mathbb{E}|\Delta\rho_t|^2)^{1/2}\\
                &(\text{using Cauchy Schwarz and Jensen's inequality})
            \end{split}
        \end{equation*}
Combining \termw{1}-\termw{4} into \eqref{eq:a8.7}
        \begin{equation*}
            \begin{split}
                \frac{1}{T}\sum_{t=0}^{T-1}\mathbb{E}||\Delta\rho_t||^2&\leq\frac{2(C_r+C_w)^2T^{\sigma-1}}{C_\alpha}+\frac{C_sC_\alpha T^{-\sigma}}{1-\sigma}\\
                &\quad+\frac{L_p G_\theta C_\gamma}{C_\alpha}\Big(\frac{T^{-2(v-\sigma)}}{1-2(v-\sigma)}\Big)^{1/2}\Bigg(\frac{1}{T}\sum_{t=0}^{T-1}\mathbb{E}|\Delta\rho_t|^2\Bigg)^{1/2}\\
                &\quad+2\Bigg(\frac{1}{T}\sum_{t=0}^{T-1}\mathbb{E}||\Delta \bar{w}_t||^2\Bigg)^{1/2}\Bigg(\frac{1}{T}\sum_{t=0}^{T-1}\mathbb{E}|\Delta\rho_t|^2\Bigg)^{1/2}
            \end{split}
        \end{equation*}

        \begin{equation*}
            \begin{split}
                M(T) &= \frac{1}{T}\sum_{t=0}^{T-1}\mathbb{E}||\Delta\rho_t||^2\\
                N(T) &= \frac{1}{T}\sum_{t=0}^{T-1}\mathbb{E}||\Delta \bar{w}_t||^2\\
                M(T)&\leq K_1+K_2\sqrt{M(T)}+K_3\sqrt{M(T)}\sqrt{N(T)}
            \end{split}
        \end{equation*}
Here,
        \begin{equation*}
            \begin{split}
                K_1 &= \frac{2(C_r+C_w)^2T^{\sigma-1}}{C_\alpha}+\frac{C_sC_\alpha T^{-\sigma}}{1-\sigma}\\
                K_2 &= \frac{L_p G_\theta C_\gamma}{C_\alpha}\Big(\frac{T^{-2(v-\sigma)}}{1-2(v-\sigma)}\Big)^{1/2}\\
                K_3 &= 2
            \end{split}
        \end{equation*}
From Lemma \ref{lm:a4}, we know that
        \begin{equation*}
            \begin{split}
                M(T) &\leq 2(\sqrt{K_1}+K_2)^2+2K_3^2N(T)\\
            \end{split}
        \end{equation*}
Hence,
        \begin{equation*}
            \begin{split}
                \frac{1}{T}\sum_{t=0}^{T-1}\mathbb{E}|\Delta\rho_t|^2&\leq 2\Bigg(\sqrt{\frac{2(C_r+C_w)^2}{C_\alpha}T^{\sigma-1}+\frac{C_sC_\alpha }{1-\sigma}T^{-\sigma}} +\frac{L_p G_\theta C_\gamma}{C_\alpha}\Big(\frac{T^{-2(v-\sigma)}}{1-2(v-\sigma)}\Big)^{1/2}\Bigg)^2\\
                &\quad+8\frac{1}{T}\sum_{t=0}^{T-1}\mathbb{E}||\Delta \bar{w}_t||^2
            \end{split}
        \end{equation*}
\end{proof}

\begin{theorem}\label{th:a3}
The on-policy average reward actor critic algorithm (Algorithm \ref{alg:2}) obtains an $\epsilon$-accurate optimal point with sample complexity of $\Omega(\epsilon^{-2.5})$. We obtain
\begin{equation*}
\begin{split}
\min_{0\leq t \leq T-1} \mathbb{E}||\nabla_\theta \rho(\theta_t)||^2 & = \mathcal{O}\bigg(\frac{1}{T^{2/5}}\bigg) + 3C_{\pi}^2(C_{a\phi}^2\tau^2 + \frac{4}{M}C_{\pi}^2C_{w_{\epsilon}}^2),  \\   
&\leq \epsilon + \mathcal{O}(1). 
\end{split}
\end{equation*}
Here,$\|\nabla_{\theta} \pi(s)\| \leq C_{\pi}$ (Assumption \ref{as:6}), $\tau = \max_{t} \|w_{t}^{*} - w_{\epsilon,t}^{*}\|$, $w_{\epsilon}^{*}$ is the optimal differential Q-value function parameter according to Lemma \ref{lm:2}. Constant $C_{w_{\epsilon}^{*}}$ is defined in Lemma \ref{lm:a13}. M is the size of batch of samples used to update parameters. $C_{a\phi}$ is the Lipchitz constant defined in Assumption \ref{as:17}. 
\end{theorem}
\begin{proof}
From Lemma \ref{lm:a4} we have:
\begin{equation}\label{eq:9.0}
    \begin{split}
        \frac{1}{T}\sum_{t=0}^{T-1}\mathbb{E}||\Delta w_t||^2 &\leq 2\Bigg(\sqrt{\frac{2C_w^2}{(\lambda-1) C_\alpha}T^{\sigma-1}+\frac{C_gC_\alpha}{1-\sigma}T^{-\sigma}} +\frac{L_wG_\theta C_\gamma}{(\lambda-1)C_\alpha}\bigg(\frac{T^{-2(v-\sigma)}}{1-2(v-\sigma)}\bigg)^{\frac{1}{2}}\Bigg)^2\\
        &\quad +\frac{4}{(\lambda-1)^2}\Bigg(\frac{1}{T}\sum_{t=0}^{T-1}\mathbb{E}|\Delta\bar{\rho}_t|^2 + \frac{1}{T}\sum_{t=0}^{T-1}\mathbb{E}||\Delta \bar{w}_t||^2\Bigg)
    \end{split}
\end{equation}

Using Lemma \ref{lm:a4.8} for $\frac{1}{T}\sum_{t=0}^{T-1}\mathbb{E}||\Delta \bar{w}_t||^2$ and Lemma \ref{lm:a4.9} for $\frac{1}{T}\sum_{t=0}^{T-1}\mathbb{E}||\Delta \bar{\rho}_t||^2$ in (\ref{eq:9.0}):
    \begin{align*}        
        \frac{1}{T}\sum_{t=0}^{T-1}\mathbb{E}||\Delta w_t||^2 \leq& 2\Bigg(\sqrt{\frac{2C_w^2}{(\lambda-1) C_\alpha}T^{\sigma-1}+\frac{C_gC_\alpha}{1-\sigma}T^{-\sigma}} +\frac{L_wG_\theta C_\gamma}{(\lambda-1)C_\alpha}\bigg(\frac{T^{-2(v-\sigma)}}{1-2(v-\sigma)}\bigg)^{\frac{1}{2}}\Bigg)^2\\
        &+\frac{4}{(\lambda-1)^2}\Bigg(
        2\Bigg(\sqrt{\frac{2C_w^2T^{u-1}}{C_\beta}+\frac{C_{gt}C_\beta T^{-u}}{1-u}}\\
        &+L_p G_\theta C_\gamma\Big(\frac{T^{-v}}{(1-2v)^{1/2}} + \frac{T^{-(v-u)}}{C_\beta(1-2(v-u))^{1/2}}\Big)\Bigg)^2\Bigg)\\
        &+\frac{4}{(\lambda-1)^2}\Bigg(
        2\Bigg(\sqrt{\frac{2(C_r+C_w)^2T^{u-1}}{C_\beta}+\frac{C_{st}C_\beta T^{-u}}{1-u}}\\
        &+L_p G_\theta C_\gamma\Big(\frac{T^{-v}}{(1-2v)^{1/2}} + \frac{T^{-(v-u)}}{C_\beta(1-2(v-u))^{1/2}}\Big)\Bigg)^2\Bigg)\\
        &+\frac{4}{(\lambda-1)^2}\Bigg(\frac{2}{T}\sum_{t=0}^{T-1}\mathbb{E}||\Delta \rho_{t+1}||^2 +\frac{2}{T}\sum_{t=0}^{T-1}\mathbb{E}||\Delta w_{t+1}||^2\Bigg)
    \end{align*}

\begin{equation*}
    \begin{split}
        \frac{1}{T}\sum_{t=0}^{T-1}\mathbb{E}||\Delta w_t||^2 \leq& 2\Bigg(\sqrt{\frac{2C_w^2}{(\lambda-1) C_\alpha}T^{\sigma-1}+\frac{C_gC_\alpha}{1-\sigma}T^{-\sigma}} +\frac{L_wG_\theta C_\gamma}{(\lambda-1)C_\alpha}\bigg(\frac{T^{-2(v-\sigma)}}{1-2(v-\sigma)}\bigg)^{\frac{1}{2}}\Bigg)^2\\
        &+\frac{4}{(\lambda-1)^2}\Bigg(
        2\Bigg(\sqrt{\frac{2C_w^2T^{u-1}}{C_\beta}+\frac{C_{gt}C_\beta T^{-u}}{1-u}}\\
        &+L_p G_\theta C_\gamma\Big(\frac{T^{-v}}{(1-2v)^{1/2}} + \frac{T^{-(v-u)}}{C_\beta(1-2(v-u))^{1/2}}\Big)\Bigg)^2\Bigg)\\
        &+\frac{4}{(\lambda-1)^2}\Bigg(
        2\Bigg(\sqrt{\frac{2(C_r+C_w)^2T^{u-1}}{C_\beta}+\frac{C_{st}C_\beta T^{-u}}{1-u}}\\
        &+L_p G_\theta C_\gamma\Big(\frac{T^{-v}}{(1-2v)^{1/2}} + \frac{T^{-(v-u)}}{C_\beta(1-2(v-u))^{1/2}}\Big)\Bigg)^2\Bigg)\\
        &+\frac{8}{(\lambda-1)^2T}\sum_{t=0}^{T-1}\mathbb{E}||\Delta \rho_{t}||^2 +\frac{8}{(\lambda-1)^2T}(\mathbb{E}||\Delta \rho_{T}||^2 -\mathbb{E}||\Delta \rho_{0}||^2  )\\
        &+\frac{8}{(\lambda-1)^2T}\sum_{t=0}^{T-1}\mathbb{E}||\Delta w_{t}||^2 +\frac{8}{(\lambda-1)^2T}(\mathbb{E}||\Delta w_{T}||^2-\mathbb{E}||\Delta w_{0}||^2)
    \end{split}
\end{equation*}

Using Lemma \ref{lm:a5} for $\sum_{t=0}^{T-1}\mathbb{E}||\Delta \rho_{t}||^2 $ in the above equation:

\begin{equation*}
    \begin{split}
        \frac{1}{T}\sum_{t=0}^{T-1}\mathbb{E}||\Delta w_t||^2 \leq& 2\Bigg(\sqrt{\frac{2C_w^2}{(\lambda-1) C_\alpha}T^{\sigma-1}+\frac{C_gC_\alpha}{1-\sigma}T^{-\sigma}} +\frac{L_wG_\theta C_\gamma}{(\lambda-1)C_\alpha}\bigg(\frac{T^{-2(v-\sigma)}}{1-2(v-\sigma)}\bigg)^{\frac{1}{2}}\Bigg)^2\\
        &+\frac{4}{(\lambda-1)^2}\Bigg(
        2\Bigg(\sqrt{\frac{2C_w^2T^{u-1}}{C_\beta}+\frac{C_{gt}C_\beta T^{-u}}{1-u}}\\
        &+L_p G_\theta C_\gamma\Big(\frac{T^{-v}}{(1-2v)^{1/2}} + \frac{T^{-(v-u)}}{C_\beta(1-2(v-u))^{1/2}}\Big)\Bigg)^2\Bigg)\\
        &+\frac{4}{(\lambda-1)^2}\Bigg(
        2\Bigg(\sqrt{\frac{2(C_r+C_w)^2T^{u-1}}{C_\beta}+\frac{C_{st}C_\beta T^{-u}}{1-u}}\\
        &+L_p G_\theta C_\gamma\Big(\frac{T^{-v}}{(1-2v)^{1/2}} + \frac{T^{-(v-u)}}{C_\beta(1-2(v-u))^{1/2}}\Big)\Bigg)^2\Bigg)\\
        &+\frac{16}{(\lambda-1)^2}
        \Bigg(\sqrt{\frac{2(C_r+C_w)^2}{C_\alpha}T^{\sigma-1}+\frac{C_sC_\alpha }{1-\sigma}T^{-\sigma}} +\frac{L_p G_\theta C_\gamma}{C_\alpha}\Big(\frac{T^{-2(v-\sigma)}}{1-2(v-\sigma)}\Big)^{1/2}\Bigg)^2\\                &+\frac{64}{(\lambda-1)^2}\frac{1}{T}\sum_{t=0}^{T-1}\mathbb{E}||\Delta \bar{w}_t||^2 +\frac{8}{(\lambda-1)^2}\frac{1}{T}\sum_{t=0}^{T-1}\mathbb{E}||\Delta w_{t}||^2  \\
        &+\frac{8}{(\lambda-1)^2T}(\mathbb{E}||\Delta \rho_{T}||^2 -\mathbb{E}||\Delta \rho_{0}||^2) + \frac{8}{(\lambda-1)^2T}(\mathbb{E}||\Delta w_{T}||^2-\mathbb{E}||\Delta w_{0}||^2)
    \end{split}
\end{equation*}

Using Lemma \ref{lm:a4.8} for $\sum_{t=0}^{T-1}\mathbb{E}||\Delta \bar{w}_{t}||^2 $:
    \begin{align*}
        \frac{1}{T}\sum_{t=0}^{T-1}\mathbb{E}||\Delta w_t||^2 \leq& 2\Bigg(\sqrt{\frac{2C_w^2}{(\lambda-1) C_\alpha}T^{\sigma-1}+\frac{C_gC_\alpha}{1-\sigma}T^{-\sigma}} +\frac{L_wG_\theta C_\gamma}{(\lambda-1)C_\alpha}\bigg(\frac{T^{-2(v-\sigma)}}{1-2(v-\sigma)}\bigg)^{\frac{1}{2}}\Bigg)^2\\
        &+\frac{136}{(\lambda-1)^2}\Bigg(\sqrt{\frac{2C_w^2T^{u-1}}{C_\beta}+\frac{C_{gt}C_\beta T^{-u}}{1-u}}\\
        &+L_p G_\theta C_\gamma\Big(\frac{T^{-v}}{(1-2v)^{1/2}} + \frac{T^{-(v-u)}}{C_\beta(1-2(v-u))^{1/2}}\Big)\Bigg)^2\\
        &+\frac{8}{(\lambda-1)^2}\Bigg(\sqrt{\frac{2(C_r+C_w)^2T^{u-1}}{C_\beta}+\frac{C_{st}C_\beta T^{-u}}{1-u}}\\
        &+L_p G_\theta C_\gamma\Big(\frac{T^{-v}}{(1-2v)^{1/2}} + \frac{T^{-(v-u)}}{C_\beta(1-2(v-u))^{1/2}}\Big)\Bigg)^2\\
        &+\frac{16}{(\lambda-1)^2}
        \Bigg(\sqrt{\frac{2(C_r+C_w)^2}{C_\alpha}T^{\sigma-1}+\frac{C_sC_\alpha }{1-\sigma}T^{-\sigma}} +\frac{L_p G_\theta C_\gamma}{C_\alpha}\Big(\frac{T^{-2(v-\sigma)}}{1-2(v-\sigma)}\Big)^{1/2}\Bigg)^2\\                &+\frac{136}{(\lambda-1)^2}\frac{1}{T}\sum_{t=0}^{T-1}\mathbb{E}||\Delta w_{t}||^2  \\
        &+\frac{8}{(\lambda-1)^2T}(\mathbb{E}||\Delta \rho_{T}||^2 -\mathbb{E}||\Delta \rho_{0}||^2) + \frac{136}{(\lambda-1)^2T}(\mathbb{E}||\Delta w_{T}||^2-\mathbb{E}||\Delta w_{0}||^2)
    \end{align*}

\begin{equation*}
    \begin{split}
        &\frac{1}{T}\sum_{t=0}^{T-1}\mathbb{E}||\Delta w_t||^2 \\
        \leq& \frac{2(\lambda-1)^2}{(\lambda-1)^2-136}\Bigg(\sqrt{\frac{2C_w^2}{(\lambda-1) C_\alpha}T^{\sigma-1}+\frac{C_gC_\alpha}{1-\sigma}T^{-\sigma}} +\frac{L_wG_\theta C_\gamma}{(\lambda-1)C_\alpha}\bigg(\frac{T^{-2(v-\sigma)}}{1-2(v-\sigma)}\bigg)^{\frac{1}{2}}\Bigg)^2\\
        &+\frac{136}{(\lambda-1)^2-136}\Bigg(\sqrt{\frac{2C_w^2T^{u-1}}{C_\beta}+\frac{C_{gt}C_\beta T^{-u}}{1-u}}\\
        &+L_p G_\theta C_\gamma\Big(\frac{T^{-v}}{(1-2v)^{1/2}} + \frac{T^{-(v-u)}}{C_\beta(1-2(v-u))^{1/2}}\Big)\Bigg)^2\\
        &+\frac{8}{(\lambda-1)^2-136}\Bigg(\sqrt{\frac{2(C_r+C_w)^2T^{u-1}}{C_\beta}+\frac{C_{st}C_\beta T^{-u}}{1-u}}\\
        &+L_p G_\theta C_\gamma\Big(\frac{T^{-v}}{(1-2v)^{1/2}} + \frac{T^{-(v-u)}}{C_\beta(1-2(v-u))^{1/2}}\Big)\Bigg)^2\\
        &+\frac{16}{(\lambda-1)^2-136}
        \Bigg(\sqrt{\frac{2(C_r+C_w)^2}{C_\alpha}T^{\sigma-1}+\frac{C_sC_\alpha }{1-\sigma}T^{-\sigma}} +\frac{L_p G_\theta C_\gamma}{C_\alpha}\Big(\frac{T^{-2(v-\sigma)}}{1-2(v-\sigma)}\Big)^{1/2}\Bigg)^2\\                &+\frac{8}{(\lambda-1)^2-136}\frac{1}{T}(\mathbb{E}||\Delta \rho_{T}||^2 -\mathbb{E}||\Delta \rho_{0}||^2) + \frac{136}{(\lambda-1)^2-136}\frac{1}{T}(\mathbb{E}||\Delta w_{T}||^2-\mathbb{E}||\Delta w_{0}||^2)
    \end{split}
\end{equation*}

\begin{equation}\label{eq:a9}
    \begin{split}
        \frac{1}{T}\sum_{t=0}^{T-1}\mathbb{E}||\Delta w_t||^2 \leq&\mathcal{O}\bigg(\frac{1}{T^{1-\sigma}}\bigg) + \mathcal{O}\bigg(\frac{1}{T^{\sigma}}\bigg) + \mathcal{O}\bigg(\frac{1}{T^{2(v-\sigma)}}\bigg) +  \mathcal{O}\bigg(\frac{1}{T^{1-u}}\bigg) +
        \mathcal{O}\bigg(\frac{1}{T^{u}}\bigg) + 
        \mathcal{O}\bigg(\frac{1}{T^{2v}}\bigg) \\
        &+\mathcal{O}\bigg(\frac{1}{T^{2(v-u)}}\bigg)+
        \frac{8}{(\lambda-1)^2-136}\frac{1}{T}(\mathbb{E}||\Delta \rho_{T}||^2 -\mathbb{E}||\Delta \rho_{0}||^2)\\
        &+ \frac{136}{(\lambda-1)^2-136}\frac{1}{T}(\mathbb{E}||\Delta w_{T}||^2-\mathbb{E}||\Delta w_{0}||^2)
    \end{split}
\end{equation}

From Lemma \ref{lm:a3} we have :
\begin{equation*}
\begin{split}
\frac{1}{T}\sum^{T-1}_{t=0} E ||\nabla_{\theta} \rho(\theta_t)||^2 & \leq 4\frac{C_r}{C_\gamma}T^{-1} + 3C_{\pi}^2C_{a\phi}^2(\frac{1}{T}\sum^{T-1}_{t=0}E ||\Delta w_{t}||^2)
+ 3C_{\pi}^2(C_{a\phi}^2\tau^2 + \frac{4}{M}C_{\pi}^2C_{w_{\epsilon}}^2) \\
& \;\;+ \frac{C_{\gamma}L_{J}G_{\theta}^2}{1-v}T^{-v}
\end{split}
\end{equation*}

Using Lemma \ref{lm:a3} in \ref{eq:a9}:
\begin{equation*}
    \begin{split}
        \frac{1}{T}\sum_{t=0}^{T-1}\mathbb{E}||\nabla_{\theta}\rho(\theta_t)||^2 \leq&\mathcal{O}\bigg(\frac{1}{T^{1-\sigma}}\bigg) + \mathcal{O}\bigg(\frac{1}{T^{\sigma}}\bigg) + \mathcal{O}\bigg(\frac{1}{T^{2(v-\sigma)}}\bigg) +  \mathcal{O}\bigg(\frac{1}{T^{1-u}}\bigg) +
        \mathcal{O}\bigg(\frac{1}{T^{u}}\bigg) + 
        \mathcal{O}\bigg(\frac{1}{T^{2v}}\bigg) \\
        &+\mathcal{O}\bigg(\frac{1}{T^{2(v-u)}}\bigg)
        +\mathcal{O}\bigg(\frac{1}{T}\bigg)
        +\mathcal{O}\bigg(\frac{1}{T^{v}}\bigg)
        +3C_{\pi}^2(C_{a\phi}^2\tau^2 + \frac{4}{M}C_{\pi}^2C_{w_{\epsilon}}^2)\\
        &+\underbrace{\frac{24C_{\pi}^2C_{a\phi}^2(\mathbb{E}||\Delta \rho_{T}||^2 -\mathbb{E}||\Delta \rho_{0}||^2)}{(\lambda-1)^2-136}\frac{1}{T}}_{\termw{I}}+
        \underbrace{\frac{408C_{\pi}^2C_{a\phi}^2(\mathbb{E}||\Delta w_{T}||^2-\mathbb{E}||\Delta w_{0}||^2)}{(\lambda-1)^2-136}\frac{1}{T}}_{\termw{II}}
    \end{split}
\end{equation*}

$\|\Delta\rho_t\|$ is bounded because of Lemma \ref{lm:a8} and because $\rho^{*}_t (= \rho(\theta_t))$ is bounded. $\|\Delta w_t\|$ is bounded because of projection operator $\Gamma_{C_w}$ and Lemma \ref{lm:a6}. Hence, we have, $\termw{I}$ and $\termw{II}$ are $\mathcal{O}\Big(\dfrac{1}{T}\Big)$ terms. \\

\noindent
By setting $\sigma = 2/5, u = 2/5$, and $v = 3/5$, we obtain the following bound :
\begin{equation*}
    \begin{split}
        \min_{0\leq t\leq T-1} \mathbb{E}||\nabla_{\theta}\rho(\theta_t)||^2 \leq \frac{1}{T}\sum_{t=0}^{T-1}\mathbb{E}||\nabla_{\theta}\rho(\theta_t)||^2 \leq &\mathcal{O}\bigg(\frac{1}{T^{2/5}}\bigg) +3C_{\pi}^2(C_{a\phi}^2\tau^2 + \frac{4}{M}C_{\pi}^2C_{w_{\epsilon}}^2)\\
        &\leq \epsilon + \mathcal{O}(1)
    \end{split}
\end{equation*}
\noindent
The sample complexity of the on-policy algorithm (Algorithm \ref{alg:2}) is $\Omega(\epsilon^{-2.5})$.

\end{proof}

\begin{lemma}\label{lm:a16} For off-policy Algorithm \ref{alg:3}, let the cumulative error of off-policy actor be $\sum^{T-1}_{t=0} E ||\widehat{\nabla_{\theta}\rho}(\theta_t)||^2$ and cumulative error of differential Q-value function be $\sum^{T-1}_{t=0}E ||\Delta w_{t}||^2$. $\theta_t$ and $w_t$ are the actor and linear differential Q-value function parameter at time t. Bound on the cumulative error of off-policy actor is proven using cumulative error of differential Q-value function as follows:

\begin{equation*}
\begin{split}    
\frac{1}{T}\sum^{T-1}_{t=0} E ||\widehat{\nabla_{\theta} \rho}(\theta_t)||^2 & \leq 4\frac{C_Q}{C_\gamma}T^{-1} + 3C_{\pi}^2C_{a\phi}^2(\frac{1}{T}\sum^{T-1}_{t=0}E ||\Delta w_{t}||^2)
+ 3C_{\pi}^2(C_{a\phi}^2\tau^2 + \frac{4}{M}C_{\pi}^2C_{w_{\epsilon}}^2) \\
& \;\;+ \frac{C_{\gamma}L_{J}'G_{\theta}^2}{1-v}T^{-v}
\end{split}
\end{equation*}

Here, $C_Q$ is the upper bound on differential Q-value function (Assumption \ref{as:17}) , $C_\gamma$, v are constants used for step size $\gamma_t$ (Assumption \ref{as:2.1}, $\|\nabla_{\theta} \pi(s)\| \leq C_{\pi}$ (Assumption \ref{as:6}), $\Delta w_{t} = w_{t} - w_{t}^{*}, \tau = \max_{t} \|w_{t}^{*} - w_{\epsilon,t}^{*}\|$, $w_{\epsilon}^{*}$ is the optimal differential Q-value function parameter according to Lemma \ref{lm:2}. $w_t^{*}$ is the optimal parameters given by TD(0) algorithm corresponding to policy parameter $\theta_t$. Constant $C_{w_{\epsilon}^{*}}$ is defined in Lemma \ref{lm:a13}. $L_{J}'$ is the coefficient used in smoothness condition of the non convex function $\rho^{\mu}(\theta)$. Constant $G_{\theta}$ is defined in  Lemma \ref{lm:a7}. M is the size of batch of samples used to update parameters. 
\end{lemma}

\begin{proof}
Let us define an objective function $\rho^{\mu}(\theta) \stackrel{\Delta}{=} \int_{S} d^{\mu}(s)Q^{\pi}_{diff}(s,\pi(s))\;ds$. Here policy $\pi$ is parameterized by $\theta$ and $d^{\mu}$ is the steady state distribution of policy $\mu$. We have,

\begin{equation}
\nabla_{\theta} \rho^{\mu}(\theta) = \int_{S} d^{\mu}(s)\nabla_{a}Q^{\pi}_{diff}(s,a)|_{a=\pi(s)}\nabla_{\theta}\pi(s)\;ds =  \widehat{\nabla_{\theta}\rho}(\theta) 
\end{equation}

By [$-L_{J}',L_{J}'$]-smoothness of non-convex function we have: 
\\
\begin{equation}\label{eq:a10}
E[\rho^{\mu}(\theta_{t+1})] \geq E[\rho^{\mu}(\theta_{t})] + E\langle \widehat{\nabla_{\theta} \rho}(\theta_{t}),\theta_{t+1} - \theta_{t} \rangle - \dfrac{L_{J}'}{2}E\|\theta_{t+1} - \theta_{t}\|^2.
\end{equation}

Now, $$h'(B_{t}, w_{t}, \theta_{t}) = \dfrac{1}{M} \sum_{i} \nabla_{a} Q^{w_t}_{diff}(s_{t,i},a)|_{a = \pi(s_{t,i})} \nabla_{\theta} \pi(s_{t,i}). $$

Here, $B_{t}$ refers to the batch of transitions sampled from the buffer at time $t$ and $\forall \;i\;\; s_{t,i} \in B_t$. 

\begin{equation}\label{eq:a10.1}
\begin{split}
E\langle \widehat{\nabla_{\theta} \rho}(\theta_{t}),\theta_{t+1} - \theta_{t} \rangle & = \gamma_{t}E\langle \widehat{\nabla_{\theta} \rho}(\theta_{t}),h'(B_{t}, w_{t}, \theta_{t}) \rangle \\ 
& =\gamma_{t}E\langle \widehat{\nabla_{\theta} \rho}(\theta_{t}),h'(B_{t}, w_{t}, \theta_{t}) - \widehat{\nabla_{\theta} \rho}(\theta_{t}) \rangle + \gamma_{t} E\|\widehat{\nabla_{\theta} \rho}(\theta_{t})\|^2.
\end{split}
\end{equation}

From (\ref{eq:a10.1}), we have
\begin{equation}\label{eq:a10.2}
\begin{split}
E\langle \widehat{\nabla_{\theta} \rho}(\theta_{t}),h'(B_{t}, w_{t}, \theta_{t}) - \widehat{\nabla_{\theta} \rho}(\theta_{t}) \rangle & \geq -\dfrac{1}{2}E\|\widehat{\nabla_\theta\rho}(\theta_{t})\|^2 -\dfrac{1}{2}E\|h'(B_{t}, w_{t}, \theta_{t}) - \widehat{\nabla_{\theta} \rho}(\theta_{t})\|^2 \\
(\because x^\intercal y \geq -\|x\|^2/2-\|y\|^2/2).
\end{split}
\end{equation}

From (\ref{eq:a10.2}):
\begin{equation} \label{eq:a10.3}
\begin{split}
& E\|h'(B_{t}, w_{t}, \theta_{t}) - \widehat{\nabla_{\theta} \rho}(\theta_{t})\|^2\\& =
E\|h'(B_{t}, w_{t}, \theta_{t}) - h'(B_{t}, w_{t}^{*}, \theta_{t}) + h'(B_{t}, w_{t}^{*}, \theta_{t}) - h'(B_{t}, w_{\epsilon,t}^{*}, \theta_{t}) +h'(B_{t}, w_{\epsilon,t}^{*}, \theta_{t}) - \widehat{\nabla_{\theta} \rho}(\theta_{t})\|^2 \\ 
& \leq 3(E\|h'(B_{t}, w_{t}, \theta_{t}) - h'(B_{t}, w_{t}^{*}, \theta_{t})\|^2 \term{1} \\ 
& \;\;\;+ E\|h'(B_{t}, w_{t}^{*}, \theta_{t}) - h'(B_{t}, w_{\epsilon,t}^{*}, \theta_{t})\|^2 \term{2} \\
& \;\;\;+ E\|h'(B_{t}, w_{\epsilon,t}^{*}, \theta_{t}) - \widehat{\nabla_{\theta} \rho}(\theta_{t})\|^2) \term{3}
\end{split}
\end{equation}

In (\ref{eq:a10.3}), $w_t^{*}$ refers to the point of convergence of off-policy TD(0) algorithm with l2-regularisation with policy $\pi(\theta_t)$ and $w_{\epsilon,t}^{*}$ refers to the optimal value function parameter  according to Compatible Function Approximation Lemma \ref{lm:a2.1}.

From (\ref{eq:a10.3}):

\termw{1}:

\begin{equation*}
\begin{split}
&E||h'(B_{t}, w_{t}, \theta_{t}) - h'(B_{t}, w_{t}^{*}, \theta_{t})||^2 \\
& = \dfrac{1}{M} || \sum_{i=0}^{M-1} \nabla_{a} Q^{w_{t}}_{diff}(s_{t,i},a)|_{a = \pi(s_{t,i})} \nabla_{\theta} \pi(s_{t,i}) - \sum_{i=0}^{M-1} \nabla_{a} Q^{w_{t}^{*}}_{diff}(s_{t,i},a)|_{a = \pi(s_{t,i})} \nabla_{\theta} \pi(s_{t,i}) ||^2 \\
& = E ||\dfrac{1}{M} \sum_{i=0}^{M-1}  \nabla_{\theta} \pi(s_{t,i}) \nabla_{a} \phi(s_{t,i},a)|_{a = \pi(s_{t,i})}^\intercal (w_t - w_t^{*}) ||^2 \\
& \leq C_{\pi}^{2}C_{a\phi}^2 E ||w_t - w_t^{*}||^2.
\end{split}
\end{equation*}

\termw{2} is similar as \termw{1}:

\begin{equation*}
\begin{split}
E||h'(B_{t}, w_{t}^{*}, \theta_{t}) - h'(B_{t}, w_{\epsilon,t}^{*}, \theta_{t})||^2 
& \leq C_{\pi}^{2}C_{a\phi}^2 E ||w_t^{*} - w_{\epsilon,t}^{*}||^2 \\
& \leq C_{\pi}^{2}C_{a\phi}^2\tau^2.
\end{split}
\end{equation*}

\termw{3} : \\
\begin{itemize}
\item By Compatible Function Approximation Lemma~\ref{lm:a2.1}: $\widehat{\nabla_\theta \rho}(\theta_{t}) = \int_{S}d^{\mu}(s) \nabla_{\theta} \pi(s) \nabla_{\theta} \pi(s)^\intercal w_{\epsilon,t}^{*} \,ds = E[h'(B_t, w_{\epsilon,t}^{*},\theta_t)]$\\

\item By lemma 4 \citep{21}, if $E[\hat{Y}] = \bar{Y}, ||\hat{Y}||,||\bar{Y}|| \leq C_{Y}$ then,
$$ E||\dfrac{1}{M}\sum_{i=0}^{M-1} \hat{Y}_{i} - \bar{Y}|| \leq 4\frac{C_{Y}^2}{M}. $$
\end{itemize}

Using above two bullet points: 
\begin{equation*}
\begin{split}
E||h'(B_{t}, w_{\epsilon,t}^{*}, \theta_{t}) - \widehat{\nabla_{\theta} \rho}(\theta_{t})||^2 
& \leq \dfrac{4}{M}||\nabla_{\theta} \pi(s) \nabla_{\theta} \pi(s)^\intercal w_{\epsilon,t}^{*} ||^2 \\
& \leq \dfrac{4C_{\pi}^4C_{w_{\epsilon}}^2}{M}.
\end{split}
\end{equation*}

Combining \termw{1},\termw{2} and \termw{3} and using in (\ref{eq:a10.3}): 

\begin{equation}\label{eq:a10.4}
E||h'(B_{t}, w_{t}, \theta_{t}) - \widehat{\nabla_{\theta} \rho}(\theta_{t})||^2 \leq 3 C_{\pi}^{2} (C_{a\phi}^2E ||w_t - w_t^{*}||^2 + C_{a\phi}^2\tau^2 + \dfrac{4C_{\pi}^2C_{w_{\epsilon}}^2}{M} ).
\end{equation}

Using (\ref{eq:a10.4}) in (\ref{eq:a10.2}):

\begin{equation}\label{eq:a10.5}
\begin{split}
E\langle \widehat{\nabla_{\theta} \rho}(\theta_{t}),h'(B_{t}, w_{t}, \theta_{t}) - \widehat{\nabla_{\theta} \rho}(\theta_{t}) \rangle & \geq -\dfrac{1}{2}E||\widehat{\nabla_\theta\rho}(\theta_{t})||^2 \\ 
& -\dfrac{3}{2} C_{\pi}^{2} (C_{a\phi}^2E ||w_t - w_t^{*}||^2 + C_{a\phi}^2\tau^2 + \dfrac{4C_{\pi}^2C_{w_{\epsilon}}^2}{M} ).
\end{split}
\end{equation}

Using (\ref{eq:a10.5}) in (\ref{eq:a10.1}):

\begin{equation}\label{eq:a10.6}
\begin{split}
E\langle \widehat{\nabla_{\theta} \rho}(\theta_{t}),\theta_{t+1} - \theta_{t} \rangle & \geq \dfrac{\gamma_t}{2}E||\widehat{\nabla_\theta\rho}(\theta_{t})||^2 \\ 
& -\dfrac{3\gamma_t}{2} C_{\pi}^{2} (C_{a\phi}^2E ||w_t - w_t^{*}||^2 + C_{a\phi}^2\tau^2 + \dfrac{4C_{\pi}^2C_{w_{\epsilon}}^2}{M} ).    
\end{split}    
\end{equation}

Using (\ref{eq:a10.6}) in (\ref{eq:a10}):
\begin{equation*}
\begin{split}
E[\rho^{\mu}(\theta_{t+1})] - E[\rho^{\mu}(\theta_{t})] & \geq
\dfrac{\gamma_t}{2}E||\widehat{\nabla_\theta\rho}(\theta_{t})||^2 - \dfrac{L_{J}'}{2}E||\theta_{t+1} - \theta_{t}||^2 \\ 
& -\dfrac{3\gamma_t}{2} C_{\pi}^{2} (C_{a\phi}^2E ||w_t - w_t^{*}||^2 + C_{a\phi}^2\tau^2 + \dfrac{4C_{\pi}^2C_{w_{\epsilon}}^2}{M} )     
\end{split}    
\end{equation*}

\begin{equation*}
\begin{split}
\implies E||\widehat{\nabla_\theta\rho}(\theta_{t})||^2 & \leq \frac{2}{\gamma_t}\Big(E[\rho^{\mu}(\theta_{t+1})] - E[\rho^{\mu}(\theta_{t})]\Big) + 3 C_{\pi}^{2}C_{a\phi}^2 (E ||w_t - w_t^{*}||^2) \\ 
& + 3 C_{\pi}^{2} (C_{a\phi}^2\tau^2 + \dfrac{4C_{\pi}^2C_{w_{\epsilon}}^2}{M}) + L_{J}'\gamma_tG_{\theta}^2 \;\;\;( \text{using Lemma \ref{lm:a7}})
\end{split}    
\end{equation*}

\begin{equation}\label{eq:a10.7}
\begin{split}
\implies \sum_{t=0}^{T-1} E||\widehat{\nabla_\theta\rho}(\theta_{t})||^2 & \leq \sum_{t=0}^{T-1} \frac{2}{\gamma_t}\Big(E[\rho^{\mu}(\theta_{t+1})] - E[\rho^{\mu}(\theta_{t})]\Big) \term{1} \\
& + \sum_{t=0}^{T-1} 3 C_{\pi}^{2}C_{a\phi}^2 (E ||w_t - w_t^{*}||^2) \term{2} \\ 
& + \sum_{t=0}^{T-1} 3 C_{\pi}^{2} (C_{a\phi}^2\tau^2 + \dfrac{4C_{\pi}^2C_{w_{\epsilon}}^2}{M}) \term{3} \\
& + \sum_{t=0}^{T-1} L_{J}'\gamma_tG_{\theta}^2 \term{4} \;\;\;( \text{using Lemma \ref{lm:a7}})
\end{split}    
\end{equation}

From \eqref{eq:a10.7}

\termw{1}:
\begin{equation*}
\begin{split}
\sum_{t=0}^{T-1} \frac{2}{\gamma_t}\Big(E[\rho^{\mu}(\theta_{t+1})] - E[\rho^{\mu}(\theta_{t})]\Big) 
&= 2\biggl(\sum_{t=1}^{T-1}\Bigl( \frac{1}{\gamma_{t-1}} - \frac{1}{\gamma_{t}}\Bigr) E[\rho^{\mu}(\theta_t)] - \frac{ E[\rho^{\mu}(\theta_0)]}{\gamma_0} + \frac{ E[\rho^{\mu}(\theta_T)]}{\gamma_{T-1}}\biggr) \\
& \leq 2\biggl(\sum_{t=1}^{T-1}\Bigl( \frac{1}{\gamma_{t-1}} - \frac{1}{\gamma_{t-1}}\Bigr) E[\rho^{\mu}(\theta_t)] + \frac{ E[\rho^{\mu}(\theta_T)]}{\gamma_{T-1}} +\Big|\frac{ E[\rho^{\mu}(\theta_0)]}{\gamma_0}\Big|\biggr) \\
& \leq 2\biggl(\sum_{t=1}^{T-1}\Bigl( \frac{1}{\gamma_{t-1}} - \frac{1}{\gamma_{t}}\Bigr) + \frac{1}{\gamma_{T-1}} + \frac{1}{\gamma_{0}} \biggr)C_{Q} \\
& \leq \frac{4C_Q}{\gamma_{0}} = \frac{4C_Q}{C_\gamma} 
\end{split}
\end{equation*}
Here, $C_Q$ is an upper bound on the differential Q-value function. Boundedness of differential Q-value function is given in \citet{23}.

\termw{2}:
\begin{equation*}
\sum_{t=0}^{T-1} 3 C_{\pi}^{2}C_{a\phi}^2 (E ||w_t - w_t^{*}||^2) = \sum_{t=0}^{T-1} 3 C_{\pi}^{2}C_{a\phi}^2 (E ||\Delta w_t||^2)
\end{equation*}

\termw{4}:
\begin{equation*}
\sum_{t=0}^{T-1} L_{J}'\gamma_tG_{\theta}^2 \leq L_{J}'G_{\theta}^2C_\gamma\frac{T^{1-v}}{1-v} \;\;\Bigl(\because \sum_{t=0}^{T-1}\frac{1}{(1+t)^v} \leq \int_{0}^{T}\frac{1}{t^v} \,dt = \frac{T^{1-v}}{1-v} \Bigr)
\end{equation*}
Using \termw{1}-\termw{4} and dividing \eqref{eq:a10.7} by T:
\begin{equation*}
\begin{split}    
\frac{1}{T}\sum^{T-1}_{t=0} E ||\widehat{\nabla_{\theta} \rho}(\theta_t)||^2 & \leq 4\frac{C_Q}{C_\gamma}T^{-1} + 3C_{\pi}^2C_{a\phi}^2(\frac{1}{T}\sum^{T-1}_{t=0}E ||\Delta w_{t}||^2)
+ 3C_{\pi}^2(C_{a\phi}^2\tau^2 + \frac{4}{M}C_{\pi}^2C_{w_{\epsilon}}^2) \\
& \;\;+ \frac{C_{\gamma}L_{J}'G_{\theta}^2}{1-v}T^{-v}
\end{split}
\end{equation*}
\end{proof}

    \begin{lemma}\label{lm:a17}
    For off-policy Algorithm \ref{alg:3}, let the cumulative error of average reward estimator be $\sum_{t=0}^{T-1}\mathbb{E}||\Delta \rho_t||^2$ and cumulative error of target linear differential Q-value function be $\sum_{t=0}^{T-1}\mathbb{E}||\Delta \bar{w}_t||^2$. $\bar{w}_t$ and $\rho_t$ are the target linear differential Q-value function parameter and average reward estimator at time t respectively. Bound on the cumulative error of average reward estimator is proven using cumulative error of target differential Q-value function as follows:
        \begin{equation*}
            \begin{split}
                 \frac{1}{T}\sum_{t=0}^{T-1}\mathbb{E}|\Delta\rho_t|^2&\leq 4\Bigg(\sqrt{\frac{2(C_r+C_w)^2}{C_\alpha}T^{\sigma-1}+\frac{C_sC_\alpha }{1-\sigma}T^{-\sigma}} +\frac{L_p G_\theta C_\gamma}{C_\alpha}\Big(\frac{T^{-2(v-\sigma)}}{1-2(v-\sigma)}\Big)^{1/2}\Bigg)^{2}\\
                &\quad+\frac{4L_p^{2}}{T}\sum_{t=0}^{T-1}\|\theta_t - \theta^{\mu}\|^{2} +8\frac{1}{T}\sum_{t=0}^{T-1}\mathbb{E}||\Delta \bar{w}_t||^2           \end{split}
        \end{equation*}
    Here, $\Delta \rho_t = \rho_t -\rho_t^{*}$, $\Delta \bar{w}_t = \bar{w}_t - w_t^{*}$. $w_t^{*}$ and $\rho_t^{*}$ are the optimal parameters given by TD(0) algorithm corresponding to policy parameter $\theta_t$. $C_\alpha$, $\sigma$ are constants and $\gamma_t, \alpha_t$ are step-sizes  defined in Assumption \ref{as:2.1}, $\|w_t\| \leq C_w$ (Algorithm \ref{alg:2}, step 8), $C_r$ is the upper bound on rewards (Assumption \ref{as:4}), Constant $G_{\theta}$ is defined in  Lemma \ref{lm:a7}. $C_s = L_p^2G_\theta^2\max_t\frac{\gamma_t^2}{\alpha_t^2}+4(C_r+C_w)^2$. $L_p$ is Lipchitz constant defined in Lemma \ref{lm:a14}. $\theta_{\mu}$ is the parameter of behaviour policy $\mu$. 

    \end{lemma}
    \begin{proof}
        \begin{equation*}
            \begin{split}
                \rho_{t+1} &= \rho_t+\alpha_t\frac{1}{M}\sum_{i=0}^{M-1}\Big(R^\mu(s_{t,i})-\rho_t+\phi^\pi(s_{t,i}')^{\intercal}\bar{w_t}-\phi^\pi(s_{t,i})^\intercal\bar{w_t}\Big)\\
                \rho_{t+1}-\rho_{t+1}^*&=\rho_t-\rho_t^*+\rho_t^*-\rho_{t+1}^*\\
                &\quad+\alpha_t\frac{1}{M}\sum_{i=0}^{M-1}\Big(R^\mu(s_{t,i})-\rho_t+\phi^\pi(s_{t,i}')^\intercal\bar{w_t}-\phi^\pi(s_{t,i})^\intercal\bar{w_t}\Big)\\
                &= \rho_t-\rho_t^*+\rho_t^*-\rho_{t+1}^*\\
                &\quad+\alpha_t\frac{1}{M}\sum_{i=0}^{M-1}\Big(R^\mu(s_{t,i})-\rho_t^*+\phi^\pi(s_{t,i}')^\intercal\bar{w_t}-\phi^\pi(s_{t,i})^\intercal\bar{w_t}\Big)\\
                &\quad+\alpha_t(\rho_t^*-\rho_t)\\
                \rho_{t+1}-\rho_{t+1}^*&=\rho_t-\rho_t^*+\rho_t^*-\rho_{t+1}^*\\
                &\quad+\alpha_t(\rho_t^*-\rho_t)\\
                &\quad+\alpha_t\Big(\frac{1}{M}\sum_{i=0}^{M-1}(\phi^\pi(s_{t,i}')-\phi^\pi(s_{t,i}))^\intercal(\bar{w_t}-w_t^*)\Big)\\
                &\quad+\alpha_t\Big(\frac{1}{M}\sum_{i=0}^{M-1}(R^\mu(s_{t,i})-\rho_t^*+\phi^\pi(s_{t,i}')^\intercal w_t^*-\phi^\pi(s_{t,i})^\intercal w_t^*)\Big)\\
            \end{split}
        \end{equation*}
Let, $l(B_t, w_t, \theta_t) := \frac{1}{M}\sum_{i=0}^{M-1}\Big(R^\mu(s_{t,i})-\rho_t^*+\phi^\pi(s_{t,i}')^\intercal w_t-\phi^\pi(s_{t,i})^\intercal w_t\Big)$. We get the following:     
        \begin{equation}\label{eq:ax1}
            \begin{split}
                \rho_{t+1}-\rho_{t+1}^*
                &=\rho_t-\rho_t^*+\rho_t^*-\rho_{t+1}^*\\
                &\quad+\alpha_t(\rho_t^*-\rho_t)\\
                &\quad+\alpha_t\Big(\frac{1}{M}\sum_{i=0}^{M-1}(\phi^\pi(s_{t,i}')-\phi^\pi(s_{t,i}))^\intercal(\bar{w_t}-w_t^*)\Big)\\
                &\quad+\alpha_t(l(B_t,w_t^*,\theta_t) - \bar{l}(w_t^{*},\theta_t))\quad \text{( $\bar{l}(w_t^{*}, \theta_t)$ is defined below)} \\
                &\quad+\alpha_t\bar{l}(w_t^{*}, \theta_t)\\
                &=\rho_t-\rho_t^* + \rho_t^* - \rho_{t+1}^* +\alpha_tl(B_t, \rho_t, w_t^*, \theta_t) \\
            \end{split}
        \end{equation}
Here,
        \begin{equation*}
            \begin{split}
                \bar{l}(w_t, \theta_t)&:=\int_{S} d^\mu\big(s)\big)\Big(R^\mu(s)-\rho(\pi(\theta_t))+\int_{S}P^{\mu}(s'|s)\phi^\pi(s')^\intercal w_t\;ds' -\phi^\pi(s)^\intercal w_t\Big)\,ds\\
                l(B_t, \rho_t, w_t, \theta_t)&:=(\rho_t^*-\rho_t) + \Big(\frac{1}{M}\sum_{i=0}^{M-1}(\phi^\pi(s_{t,i}')-\phi^\pi(s_{t,i}))^\intercal(\bar{w_t}-w_t)\Big) + l(B_t,w_t,\theta_t) - \bar{l}(w_t,\theta_t) + \bar{l}(w_t,\theta_t) \\
                & = \frac{1}{M}\sum_{i=0}^{M-1}\Big(R^\mu(s_{t,i})\rho_t+\phi^\pi(s_{t,i}')^{\intercal}\bar{w_t}-\phi^\pi(s_{t,i})^\intercal\bar{w_t}\Big)
            \end{split}
        \end{equation*}

$P^{\mu}(s'|s)$ refers to the transition probability and $d^{\mu}(\cdot)$ refers to the steady state distribution corresponding to the policy $\mu$ and $R^{\mu}(s) = R(s,\mu(s))$.

\begin{equation}\label{eq:ax2}
\begin{split}
   \bar{l}(w_t, \theta_t) =&\int_{S} d^\mu(s)\Big(R^\mu(s)-\rho(\pi(\theta_t))+\int_{S}P^{\pi}(s'|s)\phi^\pi(s')^\intercal w_t\;ds' -\phi^\pi(s)^\intercal w_t\Big)\,ds \\
   =&\int_{S} d^\mu(s)R^\mu(s)ds-\rho(\pi(\theta_t))+\int_{S} d^\mu(s)\int_{S}P^{\mu}(s'|s)\phi^\pi(s')^\intercal w_t\;ds'ds -\int_{S} d^\mu(s)\phi^\pi(s)^\intercal w_t ds \\
   =&\rho(\mu) - \rho(\pi(\theta_t)) + \int_{S} d^\mu(s)\int_{S}P^{\mu}(s'|s)\phi^\pi(s')^\intercal w_t\;ds'ds -\int_{S} d^\mu(s)\phi^\pi(s)^\intercal w_t ds \\
   =&\rho(\mu) - \rho(\pi(\theta_t)) + \int_{S} d^\mu(s')\phi^\pi(s')^\intercal w_t\;ds' -\int_{S} d^\mu(s)\phi^\pi(s)^\intercal w_t ds  \quad\Big(\because d^{\mu}(s') = \int_{S}d^{\mu}(s)P^{\mu}(s'|s)ds\Big)\\
   =& \rho(\mu) - \rho(\pi(\theta_t)) \\
   \|\bar{l}(w_t, \theta_t)\|\leq& \|\rho(\mu) - \rho(\pi(\theta_t))\| \\
   \leq& L_{p}\|\theta_t - \theta^{\mu}\|\quad \text{(Using Lemma \ref{lm:a14})}
\end{split}
\end{equation}

From (\ref{eq:ax1}), taking l2 norm on both sides we get:

        \begin{equation*}
            \begin{split}
                ||\Delta \rho_{t+1}||^2&=||\Delta\rho_t+\rho_t^*-\rho_{t+1}^{*}+\alpha_t l(B_t,w_t^*,\rho_t,\theta_t)||^2\\
                &=||\Delta\rho_t||^2+||\rho_t^*-\rho_{t+1}^*||^2+\alpha_t^2||l(B_t,w_t^*,\rho_t,\theta_t)||^2\\
                &\quad+2\langle\Delta\rho_t,\rho_t^*-\rho_{t+1}^*\rangle\\
                &\quad+2\alpha_t\langle\Delta\rho_t,l(B_t,w_t^*,\rho_t,\theta_t)\rangle\\
                &\quad+2\alpha_t\langle\rho_t^*-\rho_{t+1}^*,l(B_t,\rho_t,w_t^*,\theta_t)\rangle\\
                &\leq||\Delta\rho_t||^2+2||\rho_t^*-\rho_{t+1}^*||^2+2\alpha_t^2||l(B_t,w_t^*,\rho_t,\theta_t)||^2\\
                &\quad+2\langle\Delta\rho_t,\rho_t^*-\rho_{t+1}^*\rangle\\
                &\quad+2\alpha_t\langle\Delta\rho_t,l(B_t,w_t^*,\rho_t,\theta_t)\rangle\\
            \end{split}
        \end{equation*}
Expanding the definition of $l(B_t, w_t^{*},\rho_t, \theta_t)$ and taking expectation on both sides we get the following:
        \begin{equation}\label{eq:a8.53}
            \begin{split}
                \mathbb{E}||\Delta\rho_{t+1}||^2&\leq\mathbb{E}||\Delta\rho_t||^2+2\mathbb{E}||\rho_t^*-\rho_{t+1}^*||^2 \term{1}\\
                &\quad+2\alpha_t^2\mathbb{E}||l(B_t,w_t^*,\rho_t,\theta_t)||^2 \term{2}\\
                &\quad+2\mathbb{E}\langle\Delta\rho_t,\rho_t^*-\rho_{t+1}^*\rangle \term{3}\\
                &\quad+2\alpha_t\mathbb{E}\langle\Delta\rho_t,-\Delta\rho_t\rangle \term{4}\\
                &\quad+2\alpha_t\mathbb{E}\langle\Delta\rho_t,\frac{1}{M}\sum_{i=0}^{M-1}(\phi^\pi(s_{t,i}')-\phi^\pi(s_{t,i}))^\intercal(\bar{w_t}-w_t^*)\rangle \term{5}\\
                &\quad+2\alpha_t\mathbb{E}\langle\Delta\rho_t,l(B_t, w_t^*, \theta_t)-\bar{l}(w_t^*,\theta_t)\rangle \term{6}\\
                &\quad+2\alpha_t\mathbb{E}\langle\Delta\rho_t,\bar{l}(w_t^{*}, \theta_t)\rangle \term{7}
            \end{split}
        \end{equation}

From \eqref{eq:a8.53}:\\
\termw{1}:
        \begin{equation*}
            \begin{split}
                \mathbb{E}||\rho_t^*-\rho_{t+1}^*||^2 &\leq L_p^2\mathbb{E}||\theta_{t+1}-\theta_t||^2\text{(Lemma \ref{lm:a14})} \\
                &\leq L_p^2\gamma_t^2G_{\theta}^2 \quad(\text{Using Lemma \ref{lm:a7}})
            \end{split}
        \end{equation*}
\termw{2}:
        \begin{equation*}
            \begin{split}
                \mathbb{E}||l(B_t,\rho_t,\bar{w_t},\theta_t)||^2&=\mathbb{E}||\frac{1}{M}\sum_{i=0}^{M-1}\big(R^\pi(s_{t,i})-\rho_t+\big(\phi^\pi(s_{t,i}')-\phi^\pi(s_{t,i})\big)^\intercal\bar{w_t}\big)||^2\\
                &\leq\mathbb{E}\Big(\frac{1}{M}\sum_{i=0}^{M-1}(C_r+C_r+2C_w)\Big)^2\\
                &= 4(C_r+C_w)^2
            \end{split}
        \end{equation*}
\termw{3}:
        \begin{equation*}
            \begin{split}
                \mathbb{E}\langle\Delta\rho_t,\rho_t^*-\rho_{t+1}^*\rangle&\leq\mathbb{E}||\Delta\rho_t||\,|\rho_t^*-\rho_{t+1}^*|\\
                &\leq L_p\mathbb{E}|\Delta\rho_t|\,||\theta_{t+1}-\theta_t||\quad\text{(Using Lemma \ref{lm:a14})} \\
                &\leq L_p\gamma_tG_{\theta}\mathbb{E}|\Delta\rho_t| \quad(\text{Using Lemma \ref{lm:a7})} 
            \end{split}
        \end{equation*}
\termw{4}:
        \begin{equation*}
            \begin{split}
                \mathbb{E}\langle\Delta\rho_t,-\Delta\rho_t\rangle=-\mathbb{E}|\Delta\rho_t|^2
            \end{split}
        \end{equation*}
\termw{5}:
        \begin{equation*}
            \begin{split}
                &\mathbb{E}\langle\Delta\rho_t,\frac{1}{M}\sum_{i=0}^{M-1}\big(\phi^\pi(s_{t,i}')^\intercal-\phi^\pi(s_{t,i})^\intercal\big)(\bar{w_t}-w_t^*)\rangle\\
                &\quad\leq\mathbb{E}\Big[\frac{1}{M}\sum_{i=0}^{M-1}||\phi^\pi(s_{t,i}')-\phi^\pi(s_{t,i})||\,||\bar{w_t}-w_t^*||\,|\Delta\rho_t|\Big]\\
                &\quad\leq 2\mathbb{E}|\Delta\rho_t|\|\Delta \bar{w}_t\|
            \end{split}
        \end{equation*}
\termw{6}:
        \begin{equation*}
            \begin{split}
                \mathbb{E}\langle\Delta\rho_t,l(B_t, w_t^{*}, \theta_t) -\bar{l}(w_t^*,\theta_t)\rangle&= \mathbb{E}\langle\Delta\rho_t,\mathbb{E}[l(B_t, w_t^{*}, \theta_t) -\bar{l}(w_t^*,\theta_t)|\Delta\rho_t]\rangle \\
                &=0 \\
                \text{Note:} \mathbb{E}[l(B_t, w_t^{*}, \theta_t) -\bar{l}(w_t^*,\theta_t)|\Delta\rho_t] &=0
            \end{split}
        \end{equation*}
\termw{7}:
\begin{equation*}
\begin{split}
 \mathbb{E}\langle\Delta\rho_t,\bar{l}(w_t^{*}, \theta_t)\rangle &\leq L_p\mathbb{E}|\Delta\rho_t|\|\theta_t - \theta^{\mu}\| \quad\text{(Using (\ref{eq:ax2}))} \\
\end{split}    
\end{equation*}

Combining \termw{1}-\termw{7} into \eqref{eq:a8.53}:

        \begin{equation*}
            \begin{split}
                \mathbb{E}||\Delta\rho_{t+1}||^2&\leq(1-2\alpha_t)\mathbb{E}||\Delta\rho_t||^2+2L_p^2\gamma_t^2G_{\theta}^2\\
                &\quad+8\alpha_t^2(C_r+C_w)^2 + 2L_p\gamma_tG_{\theta}\mathbb{E}|\Delta\rho_t|\\
                &\quad+4\alpha_t\mathbb{E}|\Delta\rho_t|\|\Delta \bar{w}_t\| + 2\alpha_tL_p\mathbb{E}|\Delta\rho_t|\|\theta_t - \theta^{\mu}\|
            \end{split}
        \end{equation*}

        \begin{equation}\label{eq:a8.71}
            \begin{split}
                \implies \sum_{t=0}^{T-1}\mathbb{E}||\Delta\rho_t||^2 &\leq\sum_{t=0}^{T-1}\frac{1}{2\alpha_t}\Big(\mathbb{E}||\Delta\rho_t||^2-\mathbb{E}||\Delta\rho_{t+1}||^2\Big) \term{1}\\
                &\quad+\sum_{t=0}^{T-1}\Big(\frac{L_p^2\gamma_t^2}{\alpha_t}G_\theta^2+4\alpha_t(C_r+C_w)^2\Big)\term{2}\\
                &\quad+\sum_{t=0}^{T-1}\Big(L_pG_\theta\frac{\gamma_t}{\alpha_t}\Big)\mathbb{E}|\Delta\rho_t|\term{3}\\
                &\quad+\sum_{t=0}^{T-1}2\mathbb{E}||\Delta \bar{w}_t||\,|\Delta\rho_t| \term{4}\\
                &\quad+\sum_{t=0}^{T-1}L_p\mathbb{E}|\Delta\rho_t|\|\theta_t - \theta^{\mu}\| \term{5}
            \end{split}
        \end{equation}
From \eqref{eq:a8.71}:\\
\termw{1}:
        \begin{equation*}
            \begin{split}
                \frac{1}{2}\sum_{t=0}^{T-1}\frac{1}{\alpha_t}(\mathbb{E}||\Delta\rho_t||^2-\mathbb{E}||\Delta\rho_{t+1}||^2)&=\frac{1}{2}\Bigg(\sum_{t=0}^{T-1}\Big(\frac{1}{\alpha_t}-\frac{1}{\alpha_{t-1}}\Big)\mathbb{E}|\Delta\rho_t|^2+\frac{1}{\alpha_0}\mathbb{E}|\Delta\rho_0|^2-\frac{1}{\alpha_{T-1}}\mathbb{E}|\Delta\rho_t|^2\Bigg)\\
                &\leq\frac{1}{2}\Bigg(\sum_{t=0}^{T-1}\Big(\frac{1}{\alpha_t}-\frac{1}{\alpha_{t-1}}\Big)+\frac{1}{\alpha_0}\Bigg)4(C_r+C_w)^2\\
                &\leq\frac{2(C_r+C_w)^2}{C_\alpha}T^\sigma
            \end{split}
        \end{equation*}
\termw{2}:
        \begin{equation*}
            \begin{split}
                \sum_{t=0}^{T-1}\Big(L_p^2G_\theta^2\frac{\gamma_t^2}{\alpha_t}+4\alpha_t(C_r+C_w)^2\Big)&\leq\sum_{t=0}^{T-1}\Big(L_p^2G_\theta^2\max_t\frac{\gamma_t^2}{\alpha_t^2}+4(C_r+C_w)^2\Big)\alpha_t\\
                &\leq\sum_{t=0}^{T-1}C_s\alpha_t \quad (C_s = L_p^2G_\theta^2\max_t\frac{\gamma_t^2}{\alpha_t^2}+4(C_r+C_w)^2)\\
                &\leq\frac{C_sC_\alpha}{1-\sigma}T^{1-\sigma}
            \end{split}
        \end{equation*}
\termw{3}:
        \begin{equation*}
            \begin{split}
                \sum_{t=0}^{T-1}\Big(L_p G_\theta\frac{\gamma_t}{\alpha_t}\Big)\mathbb{E}||\Delta\rho_t||&=\sum_{t=0}^{T-1}L_p G_\theta\frac{\gamma_t}{\alpha_t}\mathbb{E}||\Delta\rho_t||\\
                &\leq L_p G_\theta\Bigg(\sum_{t=0}^{T-1}\Big(\frac{\gamma_t}{\alpha_t}\Big)^2\Bigg)^{1/2}\Big(\sum_{t=0}^{T-1}\mathbb{E}|\Delta\rho_t|^2\Big)^{1/2}\\
                &\leq \frac{L_p G_\theta C_\gamma}{C_\alpha}\Big(\frac{T^{1-2(v-\sigma)}}{1-2(v-\sigma)}\Big)^{1/2}\Big(\sum_{t=0}^{T-1}\mathbb{E}|\Delta\rho_t|^2\Big)^{1/2}\\
                &(\text{using Cauchy Schwarz and Jensen's inequality})
            \end{split}
        \end{equation*}
\termw{4}:
        \begin{equation*}
            \begin{split}
                2\sum_{t-0}^{T-1}\mathbb{E}||\Delta \bar{w}_t||\,|\Delta\rho_t|&\leq2(\sum_{t=0}^{T-1}\mathbb{E}||\Delta \bar{w}_t||^2)^{1/2}(\sum_{t=0}^{T-1}\mathbb{E}|\Delta\rho_t|^2)^{1/2}\\
                &(\text{using Cauchy Schwarz and Jensen's inequality})
            \end{split}
        \end{equation*}

\termw{5}:
\begin{equation*}
\begin{split}
\sum_{t=0}^{T-1}L_p\mathbb{E}|\Delta\rho_t|\|\theta_t - \theta^{\mu}\| &\leq L_p\bigg(\sum_{t=0}^{T-1}\mathbb{E}|\Delta\rho_t|^{2}\bigg)^{1/2}\bigg(\sum_{t=0}^{T-1}\|\theta_t - \theta^{\mu}\|^{2}\bigg)^{1/2} \\
\end{split}    
\end{equation*}

Combining \termw{1}-\termw{5} into \eqref{eq:a8.71}
        \begin{equation*}
            \begin{split}
                \frac{1}{T}\sum_{t=0}^{T-1}\mathbb{E}||\Delta\rho_t||^2&\leq\frac{2(C_r+C_w)^2T^{\sigma-1}}{C_\alpha}+\frac{C_sC_\alpha T^{-\sigma}}{1-\sigma}\\
                &\quad+\Bigg(\frac{L_p G_\theta C_\gamma}{C_\alpha}\Big(\frac{T^{-2(v-\sigma)}}{1-2(v-\sigma)}\Big)^{1/2}+L_p\bigg(\frac{1}{T}\sum_{t=0}^{T-1}\|\theta_t - \theta^{\mu}\|^{2}\bigg)^{1/2}\Bigg)\Bigg(\frac{1}{T}\sum_{t=0}^{T-1}\mathbb{E}|\Delta\rho_t|^2+\Bigg)^{1/2}\\
                &\quad+2\Bigg(\frac{1}{T}\sum_{t=0}^{T-1}\mathbb{E}||\Delta \bar{w}_t||^2\Bigg)^{1/2}\Bigg(\frac{1}{T}\sum_{t=0}^{T-1}\mathbb{E}|\Delta\rho_t|^2\Bigg)^{1/2}
            \end{split}
        \end{equation*}

        \begin{equation*}
            \begin{split}
                M(T) &= \frac{1}{T}\sum_{t=0}^{T-1}\mathbb{E}||\Delta\rho_t||^2\\
                N(T) &= \frac{1}{T}\sum_{t=0}^{T-1}\mathbb{E}||\Delta \bar{w}_t||^2\\
                M(T)&\leq K_1+K_2\sqrt{M(T)}+K_3\sqrt{M(T)}\sqrt{N(T)}
            \end{split}
        \end{equation*}
Here,
        \begin{equation*}
            \begin{split}
                K_1 &= \frac{2(C_r+C_w)^2T^{\sigma-1}}{C_\alpha}+\frac{C_sC_\alpha T^{-\sigma}}{1-\sigma}\\
                K_2 &= \frac{L_p G_\theta C_\gamma}{C_\alpha}\Big(\frac{T^{-2(v-\sigma)}}{1-2(v-\sigma)}\Big)^{1/2} + L_p\bigg(\frac{1}{T}\sum_{t=0}^{T-1}\|\theta_t - \theta^{\mu}\|^{2}\bigg)^{1/2}\\
                K_3 &= 2
            \end{split}
        \end{equation*}
From Lemma \ref{lm:a4}, we know that
        \begin{equation*}
            \begin{split}
                M(T) &\leq 2(\sqrt{K_1}+K_2)^2+2K_3^2N(T)\\
            \end{split}
        \end{equation*}
Hence,
        \begin{equation*}
            \begin{split}
                \frac{1}{T}\sum_{t=0}^{T-1}\mathbb{E}|\Delta\rho_t|^2&\leq 2\Bigg(\sqrt{\frac{2(C_r+C_w)^2}{C_\alpha}T^{\sigma-1}+\frac{C_sC_\alpha }{1-\sigma}T^{-\sigma}} +\frac{L_p G_\theta C_\gamma}{C_\alpha}\Big(\frac{T^{-2(v-\sigma)}}{1-2(v-\sigma)}\Big)^{1/2}\\
                &\quad+L_p\bigg(\frac{1}{T}\sum_{t=0}^{T-1}\|\theta_t - \theta^{\mu}\|^{2}\bigg)^{1/2}\Bigg)^2+8\frac{1}{T}\sum_{t=0}^{T-1}\mathbb{E}||\Delta \bar{w}_t||^2 \\
                \frac{1}{T}\sum_{t=0}^{T-1}\mathbb{E}|\Delta\rho_t|^2&\leq 4\Bigg(\sqrt{\frac{2(C_r+C_w)^2}{C_\alpha}T^{\sigma-1}+\frac{C_sC_\alpha }{1-\sigma}T^{-\sigma}} +\frac{L_p G_\theta C_\gamma}{C_\alpha}\Big(\frac{T^{-2(v-\sigma)}}{1-2(v-\sigma)}\Big)^{1/2}\Bigg)^{2}\\
                &\quad+\frac{4L_p^{2}}{T}\sum_{t=0}^{T-1}\|\theta_t - \theta^{\mu}\|^{2} +8\frac{1}{T}\sum_{t=0}^{T-1}\mathbb{E}||\Delta \bar{w}_t||^2                 
            \end{split}
        \end{equation*}
\end{proof}

\begin{theorem}\label{th:a4}
The off-policy average reward actor critic algorithm (Algorithm \ref{alg:3}) with behavior policy $\mu$ obtains an $\epsilon$-accurate optimal point with sample complexity of $\Omega(\epsilon^{-2.5})$. Here $\theta_\mu$ refers to the behavior policy parameter and $\theta_t$ refers to the target or current policy parameter. We obtain
\begin{equation*}
\begin{split}
\min_{0\leq t \leq T-1} \mathbb{E}\|\widehat{\nabla_\theta \rho}(\theta_t)\|^2 
& = \mathcal{O}\biggl(\frac{1}{T^{2/5}}\biggr) + 3C_{\pi}^2(C_{a\phi}^2\tau^2 + \frac{4}{M}C_{\pi}^2C_{w_{\epsilon}}^2) + \mathcal{O}(W_\theta^2) \\
& \leq \epsilon + 3C_{\pi}^2(C_{a\phi}^2\tau^2 + \frac{4}{M}C_{\pi}^2C_{w_{\epsilon}}^2) + \mathcal{O}(W_\theta^2) \\
\mbox{where } W_\theta  &:= \sup_t \|\theta^\mu - \theta_t \|.
\end{split}
\end{equation*}
Here,$\|\nabla_{\theta} \pi(s)\| \leq C_{\pi}$ (Assumption \ref{as:6}), $\tau = \max_{t} \|w_{t}^{*} - w_{\epsilon,t}^{*}\|$, $w_{\epsilon}^{*}$ is the optimal differential Q-value function parameter according to Lemma \ref{lm:2}. Constant $C_{w_{\epsilon}^{*}}$ is defined in Lemma \ref{lm:a13}. $C_{a\phi}$ is Lipchitz constant defined in Assumption \ref{as:17}. M is the size of batch of samples used to update parameters.
\end{theorem}

\begin{proof}

An upper bound on the error of differential Q-value function parameter ($\frac{1}{T}\sum_{t=0}^{T-1}\|\Delta w_t\|^2$) for off-policy case can be proven in the similar fashion as Lemma \ref{lm:a4} by suitably setting the value of $\eta$ (l2-regularisation coefficient in Algorithm \ref{alg:3}) in accordance with Lemma \ref{lm:a15}.    

The upper bound on the error of target differential Q-value function parameter ($\frac{1}{T}\sum_{t=0}^{T-1}\|\Delta \bar{w}_t\|^2$) and error of target average reward estimate ($\frac{1}{T}\sum_{t=0}^{T-1}\|\Delta \bar{\rho}_t\|^2$) for the off-policy case can be proven in the similar manner as Lemma \ref{lm:a4.8} and Lemma \ref{lm:a4.9} respectively. 

Further, the upper bound on the error of average reward estimate ($\frac{1}{T}\sum_{t=0}^{T-1}\|\Delta \rho_t\|^2$) for the off-policy case has been proved in Lemma \ref{lm:a17}.

By combining all the upper bounds mentioned above in the same way as in Theorem \ref{th:a3} we get the following: 

\begin{equation}\label{eq:a38}
    \begin{split}
        \frac{1}{T}\sum_{t=0}^{T-1}\mathbb{E}||\Delta w_t||^2 \leq&\mathcal{O}\bigg(\frac{1}{T^{1-\sigma}}\bigg) + \mathcal{O}\bigg(\frac{1}{T^{\sigma}}\bigg) + \mathcal{O}\bigg(\frac{1}{T^{2(v-\sigma)}}\bigg) +  \mathcal{O}\bigg(\frac{1}{T^{1-u}}\bigg) +
        \mathcal{O}\bigg(\frac{1}{T^{u}}\bigg) + 
        \mathcal{O}\bigg(\frac{1}{T^{2v}}\bigg) \\
        &+\mathcal{O}\bigg(\frac{1}{T^{2(v-u)}}\bigg)+
        \frac{8}{(\lambda-1)^2-136}\frac{1}{T}(\mathbb{E}||\Delta \rho_{T}||^2 -\mathbb{E}||\Delta \rho_{0}||^2)\\
        &+ \frac{136}{(\lambda-1)^2-136}\frac{1}{T}(\mathbb{E}||\Delta w_{T}||^2-\mathbb{E}||\Delta w_{0}||^2) +\frac{32L_p^2}{(\lambda-1)^2}\frac{1}{T}\sum^{T-1}_{t=0}E ||\theta^\mu - \theta_t||^2
    \end{split}
\end{equation}

From Lemma \ref{lm:a16} we have:
\begin{equation*}
\begin{split}
\frac{1}{T}\sum^{T-1}_{t=0} E ||\widehat{\nabla_{\theta} \rho}(\theta_t)||^2 & \leq 4\frac{C_Q}{C_\gamma}T^{-1} + 3C_{\pi}^2C_{a\phi}^2(\frac{1}{T}\sum^{T-1}_{t=0}E ||\Delta w_{t}||^2)
+ 3C_{\pi}^2(C_{a\phi}^2\tau^2 + \frac{4}{M}C_{\pi}^2C_{w_{\epsilon}}^2) \\
& \;\;+ \frac{C_{\gamma}L_{J}'G_{\theta}^2}{1-v}T^{-v}    
\end{split}   
\end{equation*}

Using Lemma \ref{lm:a16} and \eqref{eq:a38}:
\begin{equation*}
    \begin{split}
        \frac{1}{T}\sum_{t=0}^{T-1}\mathbb{E}||\widehat{\nabla_{\theta}\rho}(\theta_t)||^2 \leq&\mathcal{O}\bigg(\frac{1}{T^{1-\sigma}}\bigg) + \mathcal{O}\bigg(\frac{1}{T^{\sigma}}\bigg) + \mathcal{O}\bigg(\frac{1}{T^{2(v-\sigma)}}\bigg) +  \mathcal{O}\bigg(\frac{1}{T^{1-u}}\bigg) +
        \mathcal{O}\bigg(\frac{1}{T^{u}}\bigg) + 
        \mathcal{O}\bigg(\frac{1}{T^{2v}}\bigg) \\
        &+\mathcal{O}\bigg(\frac{1}{T^{2(v-u)}}\bigg)
        +\mathcal{O}\bigg(\frac{1}{T}\bigg)
        +\mathcal{O}\bigg(\frac{1}{T^{v}}\bigg)
        +3C_{\pi}^2(C_{a\phi}^2\tau^2 + \frac{4}{M}C_{\pi}^2C_{w_{\epsilon}}^2)\\
        &+\underbrace{\frac{24C_{\pi}^2C_{a\phi}^2(\mathbb{E}||\Delta \rho_{T}||^2 -\mathbb{E}||\Delta \rho_{0}||^2)}{(\lambda-1)^2-136}\frac{1}{T}}_{\termw{I}}+
        \underbrace{\frac{408C_{\pi}^2C_{a\phi}^2(\mathbb{E}||\Delta w_{T}||^2-\mathbb{E}||\Delta w_{0}||^2)}{(\lambda-1)^2-136}\frac{1}{T}}_{\termw{II}}\\
        &+\frac{96C_{\pi}^2C_{a\phi}^2L_p^2}{(\lambda-1)^2-136}\frac{1}{T}\sum^{T-1}_{t=0}E ||\theta^\mu - \theta_t||^2
    \end{split}
\end{equation*}

We have, $\termw{I}$ and $\termw{II}$ are $\mathcal{O}\Big(\dfrac{1}{T}\Big)$ terms as discussed in Theorem \ref{th:a3}. Let $Z' = \frac{96C_{\pi}^2C_{a\phi}^2L_p^2}{(\lambda-1)^2-136}$ and $W_{\theta} = \sup_t \|\theta_t -\theta^{\mu}\|$ By setting $\sigma = 2/5, u=2/5$ and $v=2/5$, we obtain:

\begin{equation*}
\begin{split}
\min_{0\leq t \leq T-1} E||\widehat{\nabla_\theta \rho}(\theta_t)||^2 & = \mathcal{O}\biggl(\frac{1}{T^{0.4}}\biggr) + 3C_{\pi}^2(C_{a\phi}^2\tau^2 + \frac{4}{M}C_{\pi}^2C_{w_{\epsilon}}^2) + \frac{Z}{T}\sum^{T-1}_{t=0}E ||\theta^\mu - \theta_t||^2 \\
&= \mathcal{O}\biggl(\frac{1}{T^{0.4}}\biggr) + 3C_{\pi}^2(C_{a\phi}^2\tau^2 + \frac{4}{M}C_{\pi}^2C_{w_{\epsilon}}^2) +\mathcal{O}(W_\theta^2). 
\end{split}
\end{equation*}
Further,
\begin{equation*}
    \mathcal{O}\biggl(\frac{1}{T^{0.4}}\biggr) \leq \epsilon.
\end{equation*}
Hence, the sample complexity of off-policy average reward actor-critic algorithm is $\Omega(\epsilon^{-2.5})$.
\end{proof}
\subsubsection{Auxiliary Lemmas}
\begin{lemma}\label{lm:a6}
The optimal differential Q-value function parameter $w(\theta_t)^{*}$ as a function of actor parameter $\theta_t$ is Lipchitz continuous with constant $L_w$. Note: $w_t^{*}: = w(\theta_t)^{*}$.
$$||w_t^{*} - w_{t+1}^{*} || \leq L_w||\theta_{t+1} - {\theta_t}||$$
\end{lemma}

\begin{proof}
$\eta$ is the l2-regularisation coefficient from Algorithm \ref{alg:2} and $\eta > \lambda_{max}^{all}$, where $\lambda_{max}^{all}$ is defined in Lemma \ref{lm:a11}. Because of carefully setting the value of $\eta$, $A(\theta_t)$ is negative definite. Thus, for on-policy TD(0) with l2-regularization and target estimators, the following condition holds true for optimal differential Q-value function parameter $w_t^{*}$:
\begin{equation*}
    E[(R^{\pi}(s)-\rho_t^{*})\phi^{\pi}(s) + (\phi^{\pi}(s)(E[\phi^{\pi}(s')]-\phi^{\pi}(s))^{\intercal} - \eta I)w_t^{*}] = 0
\end{equation*}
\begin{equation*}
\begin{split}
    & b(\theta_t) := E[(R^{\pi}(s)-\rho_t^{*})\phi^{\pi}(s)] \\
    & A(\theta_t) := E[(\phi^{\pi}(s)(E[\phi^{\pi}(s')]-\phi^{\pi}(s))^{\intercal} - \eta I)]
\end{split}
\end{equation*}
\begin{equation*}
    \therefore b(\theta_t) + A(\theta_t)w_t^{*} = 0 \implies w_t^{*} = - A(\theta_t)^{-1}b(\theta_t)
\end{equation*}
Now,
\begin{equation}\label{eq:a11}
\begin{split}
    ||w_t^{*} - w_{t+1}^{*} || & = ||A(\theta_t)^{-1}b(\theta_t) - A(\theta_{t+1})^{-1}b(\theta_{t+1})|| \\
    &\leq  ||A(\theta_t)^{-1}b(\theta_t) -A(\theta_{t+1})^{-1}b(\theta_t) +A(\theta_{t+1})^{-1}b(\theta_t) - A(\theta_{t+1})^{-1}b(\theta_{t+1})|| \\
    & \leq ||A(\theta_t)^{-1} -A(\theta_{t+1})^{-1}||\;||b(\theta_t)|| \term{1} \\  
    & \;\;\;+ ||A(\theta_{t+1})^{-1}||\;||b(\theta_t) - b(\theta_{t+1})|| \termw{2}
\end{split}    
\end{equation}

From \eqref{eq:a11}:

\termw{1}:
\begin{equation}\label{eq:a12}
\begin{split}
||A(\theta_t)^{-1} -A(\theta_{t+1})^{-1}|| & =||A(\theta_t)^{-1}A(\theta_{t+1})A(\theta_{t+1})^{-1} -A(\theta_t)^{-1}A(\theta_t)A(\theta_{t+1})^{-1}|| \\
& \leq ||A(\theta_t)^{-1}||\;||A(\theta_t)-A(\theta_{t+1})||\;||A(\theta_{t+1})^{-1}||
\end{split}
\end{equation}

From \eqref{eq:a12}:

Here, $\pi'$ and $\pi$ represents the policy with parameter $\theta_{t+1}$ and $\theta_t$ respectively. 
\begin{equation}\label{eq:a13}
\begin{split}
||A(\theta_t)-A(\theta_{t+1})|| & \leq ||\int d^{\pi'}(s)(\phi^{\pi'}(s)(\int P^{\pi'}(s'|s) \phi^{\pi'}(s') \,ds'-\phi^{\pi'}(s))^{\intercal} - \eta I)\,ds \\ 
&\qquad - \int d^{\pi}(s)(\phi^{\pi}(s)(\int P^{\pi}(s'|s) \phi^{\pi}(s') \,ds'-\phi^{\pi}(s))^{\intercal} - \eta I)\,ds || \\
&\leq ||\int d^{\pi'}(s)(\phi^{\pi'}(s)(\int P^{\pi'}(s'|s) \phi^{\pi'}(s') \,ds')^{\intercal})\,ds \\
& \qquad - \int d^{\pi}(s)(\phi^{\pi}(s)(\int P^{\pi}(s'|s) \phi^{\pi}(s') \,ds')^{\intercal})\,ds || \term{1} \\ 
& \qquad +||\int d^{\pi}(s)(\phi^{\pi}(s)(\phi^{\pi}(s))^{\intercal} )\,ds - \int d^{\pi'}(s)(\phi^{\pi'}(s)(\phi^{\pi'}(s))^{\intercal} )\,ds || \term{2}
\end{split}
\end{equation}
From \eqref{eq:a13}:\\
\termw{1}:
\begin{equation*}
\begin{split}
& ||\int d^{\pi'}(s)(\phi^{\pi'}(s)(\int P^{\pi'}(s'|s) \phi^{\pi'}(s') \,ds')^{\intercal})\,ds - \int d^{\pi}(s)(\phi^{\pi}(s)(\int P^{\pi}(s'|s) \phi^{\pi}(s') \,ds')^{\intercal})\,ds || \\
& \leq || \int (d^{\pi'}(s)-d^{\pi}(s))\phi^{\pi'}(s)(\int P^{\pi'}(s'|s) \phi^{\pi'}(s') \,ds')^{\intercal}\,ds || \\
& \qquad + ||\int d^{\pi}(s)(\phi^{\pi'}(s)-\phi^{\pi}(s))(\int P^{\pi'}(s'|s) \phi^{\pi'}(s') \,ds')^{\intercal}\,ds|| \\
& \qquad + ||\int d^{\pi}(s)\phi^{\pi}(s)(\int (P^{\pi'}(s'|s)-P^{\pi}(s'|s)) \phi^{\pi'}(s') \,ds')^{\intercal}\,ds|| \\
& \qquad + ||\int d^{\pi}(s)\phi^{\pi}(s)(\int P^{\pi}(s'|s)(\phi^{\pi'}(s')-\phi^{\pi}(s')) \,ds')^{\intercal}\,ds\| \\
&\leq L_d ||\theta_{t+1} - \theta_{t}|| \qquad \text{(Lemma \ref{lm:a12})} \\
& + L_\phi ||\theta_{t+1} - \theta_{t}|| \qquad \text{(Assumption \ref{as:8})} \\
& + L_t ||\theta_{t+1} - \theta_{t}|| \qquad \text{(Assumption \ref{as:9})}  \\
& + L_\phi ||\theta_{t+1} - \theta_{t}|| \qquad \text{(Assumption \ref{as:8})} \\
\end{split}    
\end{equation*}

\begin{equation}\label{eq:a14}
\begin{split}
& ||\int d^{\pi'}(s)(\phi^{\pi'}(s)(\int P^{\pi'}(s'|s) \phi^{\pi'}(s') \,ds')^{\intercal})\,ds - \int d^{\pi}(s)(\phi^{\pi}(s)(\int P^{\pi}(s'|s) \phi^{\pi}(s') \,ds')^{\intercal})\,ds || \\
& \leq (L_d + L_t + 2L_\phi) ||\theta_{t+1} - \theta_{t}||
\end{split}    
\end{equation}

From \eqref{eq:a13}:\\

\termw{2}:
\begin{equation}\label{eq:a15}
\begin{split}    
||\int d^{\pi}(s)&(\phi^{\pi}(s)(\phi^{\pi}(s))^{\intercal} )\,ds - \int d^{\pi'}(s)(\phi^{\pi'}(s)(\phi^{\pi'}(s))^{\intercal} )\,ds || \\
& \leq ||\int (d^{\pi}(s)-d^{\pi'}(s))\phi^{\pi}(s)(\phi^{\pi}(s))^{\intercal} \,ds|| \\
& \qquad + ||\int d^{\pi'}(s)(\phi^{\pi}(s)-\phi^{\pi'}(s))(\phi^{\pi}(s))^{\intercal} \,ds|| \\
& \qquad + ||\int d^{\pi'}(s)\phi^{\pi'}(s)(\phi^{\pi}(s)-\phi^{\pi'}(s))^{\intercal} \,ds|| \\
& \leq (L_d + 2L_\phi) ||\theta_{t+1} - \theta_{t}||
\end{split}
\end{equation}

Using \eqref{eq:a14} and \eqref{eq:a15} in \eqref{eq:a13}
\begin{equation}\label{eq:a16}
||A(\theta_t)-A(\theta_{t+1})|| \leq (2L_d + 4L_\phi +L_t) ||\theta_{t+1} - \theta_{t}||
\end{equation}

From \eqref{eq:a11}:\\

\termw{2}:
\begin{equation*}
\begin{split}
||b(\theta_t) - b(\theta_{t+1})|| &= ||\int d^{\pi'}(s)(R^{\pi'}(s)-\rho_{t+1}^{*})\phi^{\pi'}(s)\,ds - \int d^{\pi}(s)(R^{\pi}(s)-\rho_{t}^{*})\phi^{\pi}(s)\,ds||\\
& \leq ||\int d^{\pi'}(s)R^{\pi'}(s)\phi^{\pi'}(s)\,ds - \int d^{\pi}(s)R^{\pi}(s)\phi^{\pi}(s)\,ds|| \\
& \quad+ ||\int d^{\pi'}(s)\rho_{t+1}^{*}\phi^{\pi'}(s)\,ds - \int d^{\pi}(s)\rho_{t}^{*}\phi^{\pi}(s)\,ds|| \\
& \leq ||\int (d^{\pi'}(s)-d^{\pi}(s))R^{\pi'}(s)\phi^{\pi'}(s)\,ds|| \\
& \quad + ||\int d^{\pi}(s)(R^{\pi'}(s)-R^{\pi}(s))\phi^{\pi'}(s)\,ds|| \\
& \quad + ||\int d^{\pi}(s)R^{\pi}(s)(\phi^{\pi'}(s)-\phi^{\pi}(s))\,ds||\\
& \quad + ||\int (d^{\pi'}(s)-d^{\pi}(s))\rho_{t+1}^{*}\phi^{\pi'}(s)\,ds|| \\
& \quad + ||\int d^{\pi}(s)(\rho_{t+1}^{*}-\rho_{t}^{*})\phi^{\pi'}(s)\,ds|| \\
& \quad + ||\int d^{\pi}(s)\rho_{t}^{*}(\phi^{\pi'}(s)-\phi^{\pi}(s))\,ds||\\
& \leq C_rL_d||\theta_{t+1} - \theta_{t}|| \qquad \text{( Assumption \ref{as:4}, Lemma \ref{lm:a12})}\\
& \quad + L_r||\theta_{t+1} - \theta_{t}|| \qquad \text{( Assumption \ref{as:10})} \\
& \quad + C_rL_\phi||\theta_{t+1} - \theta_{t}||\qquad \text{(Assumption \ref{as:4}, Assumption \ref{as:8})} \\
& \quad + C_rL_d||\theta_{t+1} - \theta_{t}|| \qquad \text{( Assumption \ref{as:4}, Lemma \ref{lm:a12})}\\
& \quad + L_p||\theta_{t+1} - \theta_{t}|| \qquad \text{(Lemma \ref{lm:a14})} \\
& \quad + C_rL_\phi||\theta_{t+1} - \theta_{t}||\qquad \text{(Assumption \ref{as:4}, Assumption \ref{as:8})}
\end{split}
\end{equation*}

\begin{equation}\label{eq:a17}
 \implies ||b(\theta_t) - b(\theta_{t+1})|| \leq (2L_dC_r + 2C_rL_\phi +L_r+L_p)||\theta_{t+1} - \theta_{t}||
\end{equation}

Using \eqref{eq:a12}, \eqref{eq:a16} and \eqref{eq:a17} in \eqref{eq:a11}:

\begin{equation*}
\begin{split}
||w_t^{*} - w_{t+1}^{*} || & \leq ||A(\theta_t)^{-1} -A(\theta_{t+1})^{-1}||\;||b(\theta_t)||  + ||A(\theta_{t+1})^{-1}||\;||b(\theta_t) - b(\theta_{t+1})|| \\
& \leq ||A(\theta_t)^{-1}||\;||A(\theta_t)-A(\theta_{t+1})||\;||A(\theta_{t+1})^{-1}||\;||b(\theta_t)|| \\
& \quad+ ||A(\theta_{t+1})^{-1}||\;||b(\theta_t) - b(\theta_{t+1})|| \\
&\leq (2L_d + 4L_\phi +L_t)||A(\theta_t)^{-1}||\;||A(\theta_{t+1})^{-1}||\;||b(\theta_t)||\;||\theta_{t+1} - \theta_{t}|| \\
& \quad+ (2L_dC_r + 2C_rL_\phi +L_r+L_p)||A(\theta_{t+1})^{-1}||\;||\theta_{t+1} - \theta_{t}||
\end{split}    
\end{equation*}

Note:
\begin{itemize}
    \item $||b(\theta_t)|| = || \int d^{\pi}(s)(R^{\pi}(s)-\rho_{t}^{*})\phi^{\pi}(s)\,ds || \leq 2C_r$ \text{(Using Assumption \ref{as:4})}
    \item From Assumption \ref{as:12}, $\lambda_{min}$ is the lower bound on eigen values of $A(\theta)$ for all $\theta$.
\end{itemize}

\begin{equation*}
\begin{split}
\therefore ||w_t^{*} - w_{t+1}^{*} || & \leq \frac{2C_r(2L_d + 4L_\phi +L_t)}{\lambda_{min}^2} ||\theta_{t+1} - \theta_{t}|| \\
& \quad +\frac{(2L_dC_r + 2C_rL_\phi +L_r+L_p)}{\lambda_{min}} ||\theta_{t+1} - \theta_{t}|| \\
& \leq L_w||\theta_{t+1} - \theta_{t}||
\end{split}    
\end{equation*}

where,
\begin{equation*}
L_w = \frac{2C_r(2L_d + 4L_\phi +L_t)}{\lambda_{min}^2} + \frac{(2L_dC_r + 2C_rL_\phi +L_r+L_p)}{\lambda_{min}} 
\end{equation*}
\end{proof}

\begin{lemma}\label{lm:a7}
$Q^{w}_{diff}$ is the approximate differential Q-value function parameterized by $w$. Policy $\pi$ is parameterized by parameter $\theta$. Then there exist a constant $G_{\theta}$, independent of policy parameter $\theta$, such that:
$$||\frac{1}{M}\sum_{i=0}^{M-1} \nabla_{a} Q^{w}_{diff}(s,a)|_{a=\pi(s)} \nabla_\theta\pi(s)|| \leq G_\theta$$
\end{lemma}

\begin{proof}

\begin{equation}\label{eq:a18}
\begin{split}
&||Q^{w}_{diff}(s,a_1) - Q^{w}_{diff}(s,a_2)|| \leq L_a||a_1 - a_2|| \;\; \text{(Assumption \ref{as:5})} \\
& \implies ||\nabla_{a} Q^{w}_{diff}(s,a)|| \leq L_a \\
& \implies ||\nabla_{a} Q^{w}_{diff}(s,a)|_{a=\pi(s)} || \leq L_a
\end{split}
\end{equation}

\begin{equation}\label{eq:a19}
\begin{split}
&||\pi(s,\theta_1) - \pi(s,\theta_2)|| \leq L_{\pi}||\theta_1 - \theta_2|| \;\; \text{(Assumption \ref{as:6})} \\
& \implies ||\nabla_\theta \pi(s)|| \leq L_\pi
\end{split}
\end{equation}

Using \eqref{eq:a18} and \eqref{eq:a19}:  

\begin{equation*}
\begin{split}
& \|\frac{1}{M}\sum_{i=0}^{M-1} \nabla_{a} Q^{w}_{diff}(s,a)|_{a=\pi(s)} \nabla_\theta\pi(s)|| \\
& \leq \frac{1}{M} \sum_{i=0}^{M-1}||\nabla_{a} Q^{w}_{diff}(s,a)|_{a=\pi(s)} \nabla_\theta\pi(s)|| \\ 
& \leq L_aL_\pi = G_\theta  
\end{split}    
\end{equation*}
\end{proof}

\begin{lemma}\label{lm:a8}
The average reward estimate $\rho_t$ is bounded.
$$\forall t>0\;\; |\rho_t| \leq C_r + 2C_w$$

Here, $C_w$ is the upper bound on differential Q-value function parameter $w_t$ (Algorithm \ref{alg:2}, step 8), $C_r$ is the upper bound on rewards (Assumption \ref{as:4}).
\end{lemma}

\begin{proof}
\begin{equation*}
    |\rho_0| \leq C_r + 2C_w \quad\text{(Assumption \ref{as:7})}
\end{equation*}

For t = 1:

\begin{equation*}
\begin{split}
\rho_1 &= \rho_0 + \alpha_0\bigg(\frac{1}{M}\sum_{i=0}^{M-1} \Big(R_\pi(s_{0,i}) + \phi^{\pi}(s_{0,i}')^{\intercal}\bar{w}_0 - \phi^{\pi}(s_{0,i})^{\intercal}\bar{w}_0\Big) - \rho_0\bigg) \\
& = (1-\alpha_0)\rho_0 + \alpha_0\bigg(\frac{1}{M}\sum_{i=0}^{M-1} \Big(R^\pi(s_{0,i}) + \phi^{\pi}(s_{0,i}')^{\intercal}\bar{w}_0 - \phi^{\pi}(s_{0,i})^{\intercal}\bar{w}_0\Big) \Bigg) 
\end{split}
\end{equation*}

\begin{equation*}
\begin{split}
|\rho_1| &\leq (1-\alpha_0)|\rho_0| + \alpha_0||\Bigl(\frac{1}{M}\sum_{i=0}^{M-1} \Big(R^\pi(s_{0,i}) + \phi^{\pi}(s_{0,i}')^{\intercal}\bar{w}_0 - \phi^{\pi}(s_{0,i})^{\intercal}\bar{w}_0\Big)\Bigr)|| \\
& \leq (1-\alpha_0)|\rho_0| + \alpha_0\Bigl(\frac{1}{M}\sum_{i=0}^{M-1} \Big(|R^\pi(s_{0,i})| + ||\phi^{\pi}(s_{0,i}')||\;||\bar{w}_0|| + ||\phi^{\pi}(s_{0,i})||\;||\bar{w}_0||\Big) \Bigr) \\
& \leq (1-\alpha_0)(C_r + 2C_w) + (\alpha_0)(C_r + 2C_w)\\
&= (C_r + 2C_w) \quad \text{(Assumption \ref{as:7})}
\end{split}    
\end{equation*}

Therefore the bound hold for t = 1. \\
Let the bound hold for t = k. We will prove that the bound will also hold for k+1 \\

\begin{equation*}
\begin{split}
\rho_{k+1} &= \rho_{k} + \alpha_{k}\Bigl(\frac{1}{M}\sum_{i=0}^{M-1} \Big(R_\pi(s_{k,i}) + \phi^{\pi}(s_{k,i}')^{\intercal}\bar{w}_k - \phi^{\pi}(s_{k,i})^{\intercal}\bar{w}_k\Big) - \rho_{k}\Bigr) \\
& = (1-\alpha_{k})\rho_{k} + \alpha_{k}\Bigl(\frac{1}{M}\sum_{i=0}^{M-1} \Big(R^\pi(s_{k,i}) + \phi^{\pi}(s_{k,i}')^{\intercal}\bar{w}_k - \phi^{\pi}(s_{k,i})^{\intercal}\bar{w}_k\Big) \Bigr) 
\end{split}
\end{equation*}

\begin{equation*}
\begin{split}
|\rho_{k+1}| &\leq (1-\alpha_{k})|\rho_{k}| + \alpha_{k}||\Bigl(\frac{1}{M}\sum_{i=0}^{M-1} \Big(R^\pi(s_{k,i}) + \phi^{\pi}(s_{k,i}')^{\intercal}\bar{w}_k - \phi^{\pi}(s_{k,i})^{\intercal}\bar{w}_k\Big) \Bigr)|| \\
& \leq (1-\alpha_{k})|\rho_{k}| + \alpha_{k}\Bigl(\frac{1}{M}\sum_{i=0}^{M-1} \Big(|R^\pi(s_{k,i})| + ||\phi^{\pi}(s_{k,i}')||\;||\bar{w}_k|| + ||\phi^{\pi}(s_{k,i})||\;||\bar{w}_k||\Big) \Bigr) \\
& \leq (1-\alpha_{k})(C_r + 2C_w) + (\alpha_{k})(C_r + 2C_w) = (C_r + 2C_w)
\end{split}    
\end{equation*}

The bound hold for t = k+1 as well. Hence by the principle of mathematical induction :

$$\forall t>0\;\; |\rho_t| \leq C_r + 2C_w$$

\end{proof}

\begin{lemma}\label{lm:a9}
The norm of target differential Q-value function parameter $\bar{w}_t$ is bounded
$$\forall t>0\;\; ||\bar{w}_t|| \leq C_w$$

Here, $C_w$ is the upper bound on differential Q-value function parameter $w_t$ (Algorithm \ref{alg:2}, step 8).
\end{lemma}

\begin{proof}
For t=1:
\begin{equation*}
\begin{split}
&\bar{w}_1 = (1 - \beta_0)\bar{w}_0 + \beta_0w_1 \\
&||\bar{w}_1|| \leq (1 - \beta_0)||\bar{w}_0|| + \beta_0||w_1|| \\
&||\bar{w}_1|| \leq (1 - \beta_0)C_w + \beta_0C_w \quad \text{(Assumption \ref{as:7}, Algorithm \ref{alg:2} - Step 8)}\\
&||\bar{w}_1|| \leq C_w
\end{split}    
\end{equation*}

The bound hold for t=1. \\
Let the bound hold for t = k. We will prove that the bound will also hold for k+1

\begin{equation*}
\begin{split}
&\bar{w}_{k+1} = (1 - \beta_k)\bar{w}_k + \beta_kw_{k+1} \\
&||\bar{w}_{k+1}|| \leq (1 - \beta_k)||\bar{w}_k|| + \beta_k||w_{k+1}|| \\
&||\bar{w}_{k+1}|| \leq (1 - \beta_k)C_w + \beta_kC_w \quad \text{(Assumption \ref{as:7}, Algorithm \ref{alg:2} - Step 8)}\\
&||\bar{w}_{k+1}|| \leq C_w
\end{split}    
\end{equation*}

The bound hold for t = k+1 as well. Hence by the principle of mathematical induction :
$$\forall t>0\;\; ||\bar{w}_t|| \leq C_w$$

\end{proof}

\begin{lemma}\label{lm:a10}
The norm of target average reward estimator $\bar{\rho}_t$ is bounded
$$\forall t>0\;\; ||\bar{\rho}_t|| \leq C_r + 2C_w$$
Here, $C_w$ is the upper bound on differential Q-value function parameter $w_t$ (Algorithm \ref{alg:2}, step 8), $C_r$ is the upper bound on rewards (Assumption \ref{as:4}).
\end{lemma}

\begin{proof}
For t=1:
\begin{equation*}
\begin{split}
&\bar{\rho}_1 = (1 - \beta_0)\bar{\rho}_0 + \beta_0\rho_1 \\
&||\bar{\rho}_1|| \leq (1 - \beta_0)||\bar{\rho}_0|| + \beta_0||\rho_1|| \\
&||\bar{\rho}_1|| \leq (1 - \beta_0)(C_r + 2C_w) + \beta_0(C_r + 2C_w) \quad \text{(Assumption \ref{as:7}, Lemma \ref{lm:a8})}\\
&||\bar{\rho}_1|| \leq C_r + 2C_w
\end{split}    
\end{equation*}

The bound hold for t=1. \\
Let the bound hold for t = k. We will prove that the bound will also hold for k+1

\begin{equation*}
\begin{split}
&\bar{\rho}_{k+1} = (1 - \beta_k)\bar{\rho}_k + \beta_k\rho_{k+1} \\
&||\bar{\rho}_{k+1}|| \leq (1 - \beta_k)||\bar{\rho}_k|| + \beta_k||\rho_{k+1}|| \\
&||\bar{\rho}_{k+1}|| \leq (1 - \beta_k)(C_r + 2C_w) + \beta_k(C_r + 2C_w) \quad \text{(Assumption \ref{as:7}, Lemma \ref{lm:a8})}\\
&||\bar{\rho}_{k+1}|| \leq C_r + 2C_w
\end{split}    
\end{equation*}

The bound hold for t = k+1 as well. Hence by the principle of mathematical induction :
$$\forall t>0\;\; ||\bar{\rho}_t|| \leq C_r + 2C_w$$
\end{proof}

\begin{lemma}\label{lm:a11}
The $A(\theta)$ matrix defined below is negative definite for all values of $\theta$ ($\theta$ is the policy parameter) when $\lambda = \eta - \lambda^{all}_{max}$.
$$  A(\theta) = \int d^{\pi}(s)(\phi^{\pi}(s)(\int P^{\pi}(s'|s) \phi^{\pi}(s') \,ds'-\phi^{\pi}(s))^{\intercal} - \eta I)\,ds$$

$$\forall x\;\;\;   x^{\intercal}A(\theta)x \leq -\lambda||x||^2, \;\;\; \lambda>0$$
$\eta$ is the l2-regularisation coefficient from Algorithm \ref{alg:2} and $\eta > \lambda_{max}^{all}$, where $\lambda_{max}^{all}$ is defined in the proof below.
\end{lemma}

\begin{proof}
Let:
\begin{equation}\label{eq:a20}
     A'(\theta) = \int d^{\pi}(s)(\phi^{\pi}(s)(\int P^{\pi}(s'|s) \phi^{\pi}(s') \,ds'-\phi^{\pi}(s))^{\intercal})\,ds = A(\theta) + \eta I 
\end{equation}

Here, $\eta$ is the l2-regularization coefficient from Algorithm \ref{alg:2}.

\begin{equation*}
x^{\intercal}A'(\theta)x = x^{\intercal}\Bigl(\frac{A'(\theta)^{\intercal}+A'(\theta)}{2}\Bigr)x \leq \lambda_{max}(\theta) ||x||^2    
\end{equation*}

Here, $\Bigl(\dfrac{A'(\theta)^{\intercal}+A'(\theta)}{2}\Bigr)$ is a symmetric matrix and $\lambda_{max}(\theta)$ is the maximum eigen value of the $\Bigl(\dfrac{A'(\theta)^{\intercal}+A'(\theta)}{2}\Bigr)$. Using $\lambda_{max}^{all}$ from Assumption \ref{as:13}:

\begin{equation*}
\begin{split}
&\implies x^{\intercal}A'(\theta)x \leq \lambda_{max}^{all} ||x||^2\\
& x^{\intercal}(A'(\theta)-\eta I)x \leq (\lambda_{max}^{all}-\eta) ||x||^2\\
& x^{\intercal}A(\theta)x \leq (\lambda_{max}^{all}-\eta) ||x||^2 \;\;(\text{using \ref{eq:a20})}\\
\end{split}    
\end{equation*}

Here, if we take $\eta > \lambda_{max}^{all}$ then we can set $\lambda = \eta - \lambda_{max}^{all}$.
$$\implies \forall x\;\;\;   x^{\intercal}A(\theta)x \leq -\lambda||x||^2, \;\;\; \lambda>0$$
\end{proof}

\begin{lemma}\label{lm:a12}
Let $\theta_{1}$ and $\theta_{2}$ be the policy parameter for $\pi_1$ and $\pi_2$ respectively. $d^{\pi_1}(\cdot)$ and $d^{\pi_2}(\cdot)$ be the stationary state distribution for $\pi_1$ and $\pi_2$ respectively. Here, $D_{TV}$ denotes the total variation distance between two probability distribution function. We have:
$$ \int |d^{\pi_1}(s)-d^{\pi_2}(s)|\, ds = 2D_{TV}(d^{\pi_1},d^{\pi_2}) \leq L_d||\theta_{1} -\theta_{2}|| $$

Here, $L_d = 2^{n+1}(\lceil{\log_{\kappa}a^{-1}\rceil} + 1\slash\kappa)L_{t} $. $L_{t}$ is the Lipchitz constant for the transition probability density function (Assumption \ref{as:9}). Constants a and $\kappa$ are from Assumption \ref{as:2}, n is the dimension of state space.  
\end{lemma}

\begin{proof}
\begin{equation*}
\int |d^{\pi_1}(s)-d^{\pi_2}(s)|\, ds = 2D_{TV}(d^{\pi_1},d^{\pi_2}) = 2D_{TV}(\mu_1,\mu_2)
\end{equation*}

Let $\mu_1$ and $\mu_2$ be the stationary state probability measure for $\pi_1$ and $\pi_2$ respectively. Then we have :

\begin{equation*}
\begin{split}
    &d\mu_1 = d^{\pi_1}(s)\,ds \\
    &d\mu_2 = d^{\pi_2}(s)\,ds \\
\end{split}
\end{equation*}

Using the result of Theorem 3.1 of \citet{22}:
\begin{equation}\label{eq:a22}
2D_{TV}(\mu_1,\mu_2) \leq 2 \Bigl(\lceil{\log_{\kappa}a^{-1}\rceil} + \frac{1}{\kappa}\Bigr) \|K_{1} - K_{2}\|_{TV}   
\end{equation}

where $K_{1}$ and $K_2$  are probability transition kernel for markov chain induced by policy $\pi_1$ and $\pi_2$.

From \eqref{eq:a22}:

\begin{equation*}
\begin{split}
||K_{1} - K_{2}|| & \leq \sup_{\|g\|_{TV} = 1} ||\int g(ds)(K_1(\cdot|s) - K_2(\cdot|s))||_{TV}
\end{split}    
\end{equation*}

\begin{equation*}
\begin{split}
||\int g(ds)(K_1(\cdot|s) - K_2(\cdot|s))||_{TV} & \leq \sup_{|f|\leq 1} |\iint f(s')(K_1 - K_2)(ds'|s)g(ds)| \\
&\leq \sup_{|f|\leq 1} |\iint f(s')(P^{\pi'}(s'|s) - P^{\pi}(s'|s))(s'|s)g(ds)ds'| \\
& \leq \sup_{|f|\leq 1} \iint |f(s')|\;|(P^{\pi'}(s'|s) - P^{\pi}(s'|s)|g(ds)ds' \\
& \leq L_t||\theta_1 - \theta_2|| \int g(ds) \int ds' \\
& \leq 2^{m}L_t||\theta_1 - \theta_2||
\end{split}
\end{equation*}

\begin{equation}\label{eq:a23}
\implies ||K_{1} - K_{2}|| \leq  2^{m}L_t||\theta_1 - \theta_2||      
\end{equation}

From \eqref{eq:a22} and \eqref{eq:a23}:

\begin{equation*}
\begin{split}
\int |d^{\pi'}(s)-d^{\pi}(s)|\, ds = 2D_{TV}(d^{\pi'},d^{\pi'}) & \leq  2^{n+1}(\lceil{\log_{\kappa}a^{-1}\rceil} + \frac{1}{\kappa})L_{t}||\theta_1 - \theta_2|| \\
& \leq L_d||\theta_{1} -\theta_{2}||
\end{split}
\end{equation*}

\end{proof}

\begin{lemma}\label{lm:a13}
The optimal differential Q-value function parameter $w_{\epsilon}^{*}$ according to compatible function approximation Lemma (\ref{lm:2}) is bounded by constant $C_{w_\epsilon^{*}}$.
$$||w_{\epsilon}^{*}|| \leq C_{w_\epsilon^{*}}$$
\end{lemma}

\begin{proof}
From Compatible Function Approximation Lemma \ref{lm:2}:
\begin{equation*}
\begin{split}
\nabla_\theta \rho(\pi) & =
\int_{S}d^{\pi}(s)\nabla_{a}Q_{diff}^{\pi}(s,a)|_{a =\pi(s)}\nabla_{\theta}\pi(s,\theta) \,ds \\
&= \int_{S}d^{\pi}(s)\nabla_{a}Q_{diff}^{w^{*}_{\epsilon}}(s,a)|_{a =\pi(s)}\nabla_{\theta}\pi(s,\theta) \,ds \\
& = \int_{S}d^{\pi}(s)\nabla_{\theta}\pi(s,\theta)\nabla_{\theta}\pi(s,\theta)^{\intercal}w_{\epsilon}^{*} \,ds \\
& = E[\nabla_{\theta}\pi(s,\theta)\nabla_{\theta}\pi(s,\theta)^{\intercal}]w_{\epsilon}^{*}
\end{split}    
\end{equation*}

Here,
\begin{equation*}
  H_\theta = E[\nabla_{\theta}\pi(s,\theta)\nabla_{\theta}\pi(s,\theta)^{\intercal}]     
\end{equation*}

\begin{equation*}
\begin{split}
 \nabla_\theta \rho(\pi) & = H_\theta w_{\epsilon}^{*} \\
 \implies w_{\epsilon}^{*} & = H_\theta^{-1} \nabla_\theta \rho(\pi) \\
 \implies ||w_{\epsilon}^{*}|| & \leq ||H_\theta^{-1}||\;|| \nabla_\theta \rho(\pi)||
\end{split}
\end{equation*}

By using Assumption \ref{as:14}, the lower bound on minimum eigenvalue of $H_\theta$ for all $\theta$ is $\lambda_{min}^{\epsilon}$ and using Assumption \ref{as:5} and \ref{as:6} :
\begin{equation*}
\begin{split}
  ||w_{\epsilon}^{*}|| & \leq \frac{L_aL_\pi}{\lambda_{min}^{\epsilon}} = C_{w_\epsilon^{*}}
\end{split}
\end{equation*}
\end{proof}

\begin{lemma}\label{lm:a14}
The average reward performance metric, defined in (\ref{eq:3}), $\rho(\pi)(\rho(\theta))$ is Lipchitz continuous wrt to the policy (actor) parameter $\theta$.
$$||\rho(\theta_1) - \rho(\theta_2) || \leq L_p ||\theta_1 - \theta_2||$$
\end{lemma}

\begin{proof}
Let $\theta_1$ and $\theta_2$ be the policy parameters of policy $\pi'$ and $\pi$.

\begin{equation*}
\begin{split}
||\rho(\theta_1) - \rho(\theta_2) || & = ||\rho(\pi') - \rho(\pi) || \\  
& = ||\int_{S} d^{\pi'}(s) R^{\pi'}(s) \,ds - \int_{S} d^{\pi}(s) R^{\pi}(s) \,ds || \\
& \leq ||\int_{S} (d^{\pi'}(s)-d^{\pi}(s)) R^{\pi'}(s) \,ds || + ||\int_{S} d^{\pi}(s) (R^{\pi'}(s)-R^{\pi}(s)) \,ds || \\
& \leq L_d||\theta_1 - \theta_2|| \quad \text{(Lemma \ref{lm:a12})} \\
& \quad + L_r||\theta_1 - \theta_2|| \quad \text{(Assumption \ref{as:10})} \\
& \leq (L_d+L_r)||\theta_1 - \theta_2|| = L_p||\theta_1 - \theta_2|| \quad(L_d+L_r = L_p) 
\end{split}    
\end{equation*}

\end{proof}

\begin{lemma}\label{lm:a15}
The optimal differential Q-value function parameter $w(\theta_t)^{*}$ as a function of actor parameter $\theta_t$ is Lipchitz continuous with constant $L_v$ for off-policy case. Note: $w_t^{*} = w(\theta_t)^{*}$. $\mu$ is the behaviour policy.
$$||w_t^{*} - w_{t+1}^{*} || \leq L_v||\theta_{t+1} - {\theta_t}||$$
\end{lemma}

\begin{proof}
$\eta$ is the l2-regularisation coefficient from Algorithm \ref{alg:3} and $\eta > \chi_{max}^{all}$, where $\chi_{max}^{all}$ is defined in Lemma \ref{lm:a15.1}. Because of carefully setting the value of $\eta$, $A(\theta_t)$ is negative definite. Thus, for off-policy TD(0) with l2-regularization the following condition holds true for optimal differential Q-value function parameter $w_t^{*}$:\begin{equation*}
    E[(R^{\mu}(s)-\rho_t^{*})\phi^{\pi}(s) + (\phi^{\pi}(s)(E[\phi^{\pi}(s')]-\phi^{\pi}(s))^{\intercal} - \eta I)w_t^{*}] = 0
\end{equation*}
\begin{equation*}
\begin{split}
    & b(\theta_t) := E[(R^{\mu}(s)-\rho_t^{*})\phi^{\pi}(s)] \\
    & A(\theta_t) := E[(\phi^{\pi}(s)(E[\phi^{\pi}(s')]-\phi^{\pi}(s))^{\intercal} - \eta I)]
\end{split}
\end{equation*}

\begin{equation*}
    \therefore b(\theta_t) + A(\theta_t)w_t^{*} = 0 \implies w_t^{*} = - A(\theta_t)^{-1}b(\theta_t)
\end{equation*}

Expectation above is with respect to stationary state distribution $d^{\mu}(\cdot)$ of policy $\mu$. Please note the abuse of notation here, $A(\theta_t)$ is actually same as $A_{off}^{\mu}(\theta_t)$ of Lemma \ref{lm:a15.1}.

\begin{equation}\label{eq:a29}
\begin{split}
    ||w_t^{*} - w_{t+1}^{*} || & = ||A(\theta_t)^{-1}b(\theta_t) - A(\theta_{t+1})^{-1}b(\theta_{t+1})|| \\
    &\leq  ||A(\theta_t)^{-1}b(\theta_t) -A(\theta_{t+1})^{-1}b(\theta_t) +A(\theta_{t+1})^{-1}b(\theta_t) - A(\theta_{t+1})^{-1}b(\theta_{t+1})|| \\
    & \leq ||A(\theta_t)^{-1} -A(\theta_{t+1})^{-1}||\;||b(\theta_t)|| \term{1} \\  
    & \;\;\;+ ||A(\theta_{t+1})^{-1}||\;||b(\theta_t) - b(\theta_{t+1})|| \term{2}
\end{split}    
\end{equation}

From \eqref{eq:a29}:

\termw{1}:
\begin{equation}\label{eq:a30}
\begin{split}
||A(\theta_t)^{-1} -A(\theta_{t+1})^{-1}|| & =||A(\theta_t)^{-1}A(\theta_{t+1})A(\theta_{t+1})^{-1} -A(\theta_t)^{-1}A(\theta_t)^A(\theta_{t+1})^{-1}|| \\
& \leq ||A(\theta_t)^{-1}||\;||A(\theta_t)-A(\theta_{t+1})||\;||A(\theta_{t+1})^{-1}||
\end{split}
\end{equation}

From \eqref{eq:a30}:

Here, $\pi'$ and $\pi$ represents the policy with parameter $\theta_{t+1}$ and $\theta_t$ respectively and $\mu$ be the behaviour policy . 

\begin{equation}\label{eq:a31}
\begin{split}
||A(\theta_t)-A(\theta_{t+1})|| & \leq ||\int d^{\mu}(s)(\phi^{\pi'}(s)(\int P^{\mu}(s'|s) \phi^{\pi'}(s') \,ds'-\phi^{\pi'}(s))^{\intercal} - \eta I)\,ds \\ 
&\qquad - \int d^{\mu}(s)(\phi^{\pi}(s)(\int P^{\mu}(s'|s) \phi^{\pi}(s') \,ds'-\phi^{\pi}(s))^{\intercal} - \eta I)\,ds || \\
&\leq ||\int d^{\mu}(s)(\phi^{\pi'}(s)(\int P^{\mu}(s'|s) \phi^{\pi'}(s') \,ds')^{\intercal})\,ds \\
& \qquad - \int d^{\mu}(s)(\phi^{\pi}(s)(\int P^{\mu}(s'|s) \phi^{\pi}(s') \,ds')^{\intercal})\,ds || \term{1} \\ 
& \qquad + ||\int d^{\mu}(s)(\phi^{\pi}(s)(\phi^{\pi}(s))^{\intercal} )\,ds - \int d^{\mu}(s)(\phi^{\pi'}(s)(\phi^{\pi'}(s))^{\intercal} )\,ds || \term{2}
\end{split}
\end{equation}

From \eqref{eq:a31}:\\

\termw{1}:
\begin{equation*}
\begin{split}
& ||\int d^{\mu}(s)(\phi^{\pi'}(s)(\int P^{\mu}(s'|s) \phi^{\pi'}(s') \,ds')^{\intercal})\,ds - \int d^{\mu}(s)(\phi^{\pi}(s)(\int P^{\mu}(s'|s) \phi^{\pi}(s') \,ds')^{\intercal})\,ds || \\
& \leq || \int (d^{\mu}(s)-d^{\mu}(s))\phi^{\pi'}(s)(\int P^{\mu}(s'|s) \phi^{\pi'}(s') \,ds')^{\intercal}\,ds || \\
& \qquad + ||\int d^{\mu}(s)(\phi^{\pi'}(s)-\phi^{\pi}(s))(\int P^{\mu}(s'|s) \phi^{\pi'}(s') \,ds')^{\intercal}\,ds|| \\
& \qquad + ||\int d^{\mu}(s)\phi^{\pi}(s)(\int (P^{\mu}(s'|s)-P^{\mu}(s'|s)) \phi^{\pi'}(s') \,ds')^{\intercal}\,ds|| \\
& \qquad + ||\int d^{\mu}(s)\phi^{\pi}(s)(\int P^{\mu}(s'|s)(\phi^{\pi'}(s')-\phi^{\pi}(s')) \,ds'^{\intercal})\,ds \\
& \leq 2L_\phi ||\theta_{t+1} - \theta_{t}|| \qquad \text{(Assumption \ref{as:8})} \\
\end{split}    
\end{equation*}

\begin{equation}\label{eq:a32}
\begin{split}
& ||\int d^{\mu}(s)(\phi^{\pi'}(s)(\int P^{\mu}(s'|s) \phi^{\pi'}(s') \,ds')^{\intercal})\,ds - \int d^{\mu}(s)(\phi^{\pi}(s)(\int P^{\mu}(s'|s) \phi^{\pi}(s') \,ds')^{\intercal})\,ds || \\
& \leq 2L_\phi ||\theta_{t+1} - \theta_{t}||
\end{split}    
\end{equation}
From \eqref{eq:a31}:\\
\termw{2}:
\begin{equation}\label{eq:a33}
\begin{split}    
||\int d^{\mu}(s)&(\phi^{\pi}(s)(\phi^{\pi}(s))^{\intercal} )\,ds - \int d^{\mu}(s)(\phi^{\pi'}(s)(\phi^{\pi'}(s))^{\intercal} )\,ds || \\
& \leq ||\int d^{\mu}(s)(\phi^{\pi}(s)-\phi^{\pi'}(s))(\phi^{\pi}(s))^{\intercal} \,ds|| \\
& \qquad + ||\int d^{\mu}(s)\phi^{\pi'}(s)(\phi^{\pi}(s)-\phi^{\pi'}(s))^{\intercal} \,ds|| \\
& \leq 2L_\phi ||\theta_{t+1} - \theta_{t}||
\end{split}
\end{equation}

Using \eqref{eq:a32} and \eqref{eq:a33} in \eqref{eq:a31}
\begin{equation}\label{eq:a34}
||A(\theta_t)-A(\theta_{t+1})|| \leq (4L_\phi) ||\theta_{t+1} - \theta_{t}||
\end{equation}

From \eqref{eq:a29}:\\

\termw{2}:
\begin{equation*}
\begin{split}
||b(\theta_t) - b(\theta_{t+1})|| &= ||\int d^{\mu}(s)(R^{\mu}(s)-\rho_{t+1}^{*})\phi^{\pi'}(s)\,ds - \int d^{\mu}(s)(R^{\mu}(s)-\rho_{t}^{*})\phi^{\pi}(s)\,ds||\\
& \leq ||\int d^{\mu}(s)R^{\mu}(s)\phi^{\pi'}(s)\,ds - \int d^{\mu}(s)R^{\mu}(s)\phi^{\pi}(s)\,ds|| \\
& \quad + ||\int d^{\mu}(s)\rho_{t+1}^{*}\phi^{\pi'}(s)\,ds - \int d^{\mu}(s)\rho_{t}^{*}\phi^{\pi}(s)\,ds|| \\
& \leq ||\int d^{\mu}(s)(R^{\mu}(s)-R^{\mu}(s))\phi^{\pi'}(s)\,ds|| \\
& \quad + ||\int d^{\mu}(s)R^{\mu}(s)(\phi^{\pi'}(s)-\phi^{\pi}(s))\,ds||\\
& \quad + ||\int d^{\mu}(s)(\rho_{t+1}^{*}-\rho_{t}^{*})\phi^{\pi'}(s)\,ds|| \\
& \quad + ||\int d^{\mu}(s)\rho_{t}^{*}(\phi^{\pi'}(s)-\phi^{\pi}(s))\,ds||\\
& \leq C_rL_\phi||\theta_{t+1} - \theta_{t}||\qquad \text{(Assumption \ref{as:4})} \\
& \quad+ L_p||\theta_{t+1} - \theta_{t}|| \qquad \text{(Lemma \ref{lm:a14})} \\
& \quad+ C_rL_\phi||\theta_{t+1} - \theta_{t}||\qquad \text{(Assumption \ref{as:4})}
\end{split}
\end{equation*}

\begin{equation}\label{eq:a35}
 \implies ||b(\theta_t) - b(\theta_{t+1})|| \leq (2C_rL_\phi +L_p)||\theta_{t+1} - \theta_{t}||
\end{equation}

Using \eqref{eq:a30}, \eqref{eq:a34} and \eqref{eq:a35} in \eqref{eq:a29}:

\begin{equation*}
\begin{split}
||w_t^{*} - w_{t+1}^{*} || & \leq ||A(\theta_t)^{-1} -A(\theta_{t+1})^{-1}||\;||b(\theta_t)||  + ||A(\theta_{t+1})^{-1}||\;||b(\theta_t) - b(\theta_{t+1})|| \\
& \leq ||A(\theta_t)^{-1}||\;||A(\theta_t)-A(\theta_{t+1})||\;||A(\theta_{t+1})^{-1}||\;||b(\theta_t)|| \\
& \quad+ ||A(\theta_{t+1})^{-1}||\;||b(\theta_t) - b(\theta_{t+1})|| \\
&\leq 4L_\phi||A(\theta_t)^{-1}||\;||A(\theta_{t+1})^{-1}||\;||b(\theta_t)||\;||\theta_{t+1} - \theta_{t}|| \\
& \quad+ (2C_rL_\phi +L_p)||A(\theta_{t+1})^{-1}||\;||\theta_{t+1} - \theta_{t}||
\end{split}    
\end{equation*}

Note:
\begin{itemize}
    \item $||b(\theta_t)|| = || \int d^{\mu}(s)(R^{\mu}(s)-\rho_{t}^{*})\phi^{\pi}(s)\,ds || \leq 2C_r$ \quad\text{(Assumption \ref{as:4})}
    \item Let $\lambda_{min}$ is the lower bound on eigen values of $A(\theta)$ for all $\theta$.
\end{itemize}

\begin{equation*}
\begin{split}
\therefore ||w_t^{*} - w_{t+1}^{*} || & \leq \frac{8C_r L_\phi}{\lambda_{min}^2} ||\theta_{t+1} - \theta_{t}|| \\
& +\frac{(2C_rL_\phi +L_p)}{\lambda_{min}} ||\theta_{t+1} - \theta_{t}|| \\
& \leq L_v||\theta_{t+1} - \theta_{t}||
\end{split}    
\end{equation*}

where,
\begin{equation*}
L_v = \frac{8C_rL_\phi}{\lambda_{min}^2} + \frac{(2C_rL_\phi +L_p)}{\lambda_{min}} 
\end{equation*}
\end{proof}

\begin{lemma}\label{lm:a15.1}
The $A_{off}^{\mu}(\theta)$ matrix defined below is negative definite for all values of $\theta$ ($\theta$ is the policy parameter) when $\lambda = \eta - \chi_{max}^{all}$. Here, $\theta^{\mu}$ is the policy parameter for behaviour policy $\mu$. 
$$  A_{off}^{\mu}(\theta) := \int d^{\mu}(s)(\phi^{\pi}(s)(\int P^{\pi}(s'|s) \phi^{\pi}(s') \,ds'-\phi^{\pi}(s))^{\intercal} - \eta I)\,ds$$

$$\forall x\;\;\;   x^{\intercal}A_{off}^{\mu}(\theta)x \leq -\lambda||x||^2, \;\;\; \lambda>0$$
$\eta$ is the l2-regularisation coefficient from Algorithm \ref{alg:3} and $\eta > \chi_{max}^{all}$, where $\chi_{max}^{all}$ is defined in the proof below.
\end{lemma}

\begin{proof}
Let:
\begin{equation}\label{eq:a36}
     A_{off}^{{\mu}'}(\theta) = \int d^{\mu}(s)(\phi^{\pi}(s)(\int P^{\pi}(s'|s) \phi^{\pi}(s') \,ds'-\phi^{\pi}(s))^{\intercal})\,ds = A_{off}^{\mu}(\theta) + \eta I 
\end{equation}

Here, $\eta$ is the l2-regularization coefficient from Algorithm \ref{alg:2}.

\begin{equation*}
x^{\intercal}A_{off}^{{\mu}'}(\theta)x = x^{\intercal}\Bigl(\frac{A_{off}^{{\mu}'}(\theta)^{\intercal}+A_{off}^{{\mu}'}(\theta)}{2}\Bigr)x \leq \chi_{max}(\theta) ||x||^2    
\end{equation*}

Here, $\Bigl(\dfrac{A_{off}^{{\mu}'}(\theta)^{\intercal}+A_{off}^{{\mu}'}(\theta)}{2}\Bigr)$ is a symmetric matrix and $\chi_{max}(\theta)$ is the maximum eigenvalue of the $\Bigl(\dfrac{A_{off}^{{\mu}'}(\theta)^{\intercal}+A_{off}^{{\mu}'}(\theta)}{2}\Bigr)$. Using $\chi_{max}^{all}$ from Assumption \ref{as:15}:

\begin{equation*}
\begin{split}
&\implies x^{\intercal}A_{off}^{{\mu}'}(\theta)x \leq \chi_{max}^{all} ||x||^2\\
& x^{\intercal}(A_{off}^{{\mu}'}(\theta)-\eta I)x \leq (\chi_{max}^{all}-\eta) ||x||^2\\
& x^{\intercal}A_{off}^{\mu}(\theta)x \leq (\chi_{max}^{all}-\eta) ||x||^2 \;\;(\text{using \ref{eq:a36})}\\
\end{split}    
\end{equation*}

Here, if we take $\eta > \chi_{max}^{all}$ then we can set $\lambda = \eta - \chi_{max}^{all}$.
$$\implies \forall x\;\;\;   x^{\intercal}A_{off}^{\mu}(\theta)x \leq -\lambda||x||^2, \;\;\; \lambda>0$$
\end{proof}

\subsection{Asymptotic Convergence Analysis}
\begin{theorem}\label{th:a5}
In Algorithm \ref{alg:4}, let policy parameter $\theta_t$ be kept constant at $\theta$. The differential Q-value function parameter $w_t$ and the target differential Q-value function parameter $\bar{w}_t$ converges to $w(\theta)^*$. Also, average reward estimator $\rho_t$ and target average reward estimator $\bar{\rho}_t$ converges to $\rho(\theta)^*$. 
\end{theorem}

\begin{proof}
For simplicity of proof we are assuming the batch size M to be 1. Differential Q-value function parameter $w_t \in \mathbb{R}^k$, $\phi^{\pi}(s) \in \mathbb{R}^k$ and $\rho_t$ is a scalar. Let the update rules used for differential Q-value function parameter and average reward estimator be as follows:

\begin{equation}
\begin{split}
w_{t+1} &= w_t + \alpha_t\Bigl(R^\pi(s_{t}) - \bar{\rho_t} + \phi^\pi(s_{t}')^{\intercal}\bar{w_t} - \phi^\pi(s_{t})^{\intercal}w_t\Bigr)\phi^\pi(s_{t}) - \alpha_t\eta w_t\\
\rho_{t+1} &= \rho_t+\alpha_t\Big(R^\pi(s_{t})-\rho_t+\phi^\pi(s_{t}')^{\intercal}\bar{w_t}-\phi^\pi(s_{t})^\intercal\bar{w_t}\Big)\\
\overline{w}_{t+1} &= \overline{w}_{t} + \beta_t(w_{t+1} - \overline{w}_{t})\\
\overline{\rho}_{t+1} &= \overline{\rho}_{t} + \beta_t(\rho_{t+1} - \overline{\rho}_{t})
\end{split}    
\end{equation}

Let us define $z_t$ as $[w_t \;\; \rho_t]^{\intercal}$ and $\bar{z}_t$ as $[\bar{w}_t\;\; \bar{\rho}_t]^{\intercal}$. \textbf{0} is a vector in $\mathbb{R}^k$ and $I_{0}$ is an identity matrix in $\mathbb{R}^{(k+1)\times(k+1)}$ with $I_{0}[k][k] = 0$ (assuming indexing starts from 0). 

\begin{equation}
\begin{split}
\begin{bmatrix} w_{t+1}\\ \rho_{t+1}  \end{bmatrix} =& \begin{bmatrix} w_{t}\\ \rho_{t}  \end{bmatrix} 
+ \alpha_t\Bigg(R^\pi(s_{t})\begin{bmatrix}\phi^\pi(s_{t}) \\ 1 \end{bmatrix} 
+ \begin{bmatrix}\phi^\pi(s_{t})\phi^\pi(s_{t}')^{\intercal} & -\phi^\pi(s_{t}) \\ \phi^\pi(s_{t}')^{\intercal} - \phi^\pi(s_{t})^{\intercal} & 0\end{bmatrix}\begin{bmatrix} \bar{w}_{t}\\ \bar{\rho}_{t}  \end{bmatrix} \\
& - \begin{bmatrix}\phi^\pi(s_{t})\phi^\pi(s_{t})^{\intercal} & \textbf{0}\\ \textbf{0}^{\intercal} &1 \end{bmatrix}\begin{bmatrix} w_{t}\\ \rho_{t}  \end{bmatrix} -\eta I_{0}\begin{bmatrix} w_{t}\\ \rho_{t}  \end{bmatrix}\Bigg) \\
\begin{bmatrix} \bar{w}_{t+1}\\ \bar{\rho}_{t+1}  \end{bmatrix} = & \begin{bmatrix} \bar{w}_{t}\\ \bar{\rho}_{t}  \end{bmatrix} + \beta_t\Big(\begin{bmatrix} w_{t+1}\\ \rho_{t+1}  \end{bmatrix}-\begin{bmatrix} \bar{w}_{t}\\ \bar{\rho}_{t}  \end{bmatrix}\Big)
\end{split}
\end{equation}

Here, $R^\pi(s_{t})\begin{bmatrix}\phi^\pi(s_{t}) \\ 1 \end{bmatrix} = R_{\phi}^\pi(s_{t})$,  $A_{\phi}(s_t, s_t') = \begin{bmatrix}\phi^\pi(s_{t})\phi^\pi(s_{t}')^{\intercal} & -\phi^\pi(s_{t}) \\ \phi^\pi(s_{t}')^{\intercal} - \phi^\pi(s_{t})^{\intercal} & 0\end{bmatrix}$ and $B_{\phi}(s_t) = \begin{bmatrix}\phi^\pi(s_{t})\phi^\pi(s_{t})^{\intercal} & \textbf{0}\\ \textbf{0}^{\intercal} &1 \end{bmatrix}$

\begin{equation}\label{eq:a40}
\begin{split}
z_{t+1} &= z_{t} + \alpha_t\big(R_{\phi}^{\pi}(s_t) + A_{\phi}(s_t, s_t')\bar{z}_t - (B_{\phi}(s_t)+ \eta I_{0})z_t\big)  \\
\bar{z}_{t+1} &= \bar{z}_t + \beta_t(z_{t+1} - \bar{z}_t)
\end{split}
\end{equation}

Now, we will use the extension of stability criteria for iterates given \citet{34} to two timescale stochastic approximation scheme \citep{35} to show the convergence of the differential Q-value function parameter and average reward estimator together. Let us write \eqref{eq:a40} in the standard form of stochastic approximation scheme.

\begin{equation*}
\begin{split}
z_{t+1} &= z_{t} + \alpha_t\big(h(z_t, \bar{z}_t) + \mathcal{M}^{1}_{t+1}\big) 
\end{split}
\end{equation*}
Let, $\bar{R}_{\phi}^{\pi} = \int_{S} d^{\pi}(s_t)R_{\phi}^{\pi}(s_t)\,ds_t$, $\bar{A}_{\phi} =\int_{S} d^{\pi}(s_t)\int_{S}P^{\pi}(s_t'|s_t)A_{\phi}(s_t, s_t')\,ds_t'\,ds_t$, $\bar{B}_{\phi} = \int_{S} d^{\pi}(s_t)B_{\phi}(s_t)\,s_t$

Here,
\begin{equation*}
\begin{split}
h(z_t, \bar{z}_t) =& \int_{S} d^{\pi}(s_t)\big( R_{\phi}^{\pi}(s_t) + A_{\phi}(s_t, s_t')\bar{z}_t - (B_{\phi}(s_t)+ \eta I_{0})z_t\big)\,ds_t \\
=& \bar{R}_{\phi}^{\pi} + \bar{A}_{\phi}\bar{z}_t - (\bar{B}_{\phi}+ \eta I_{0})z_t \\
\mathcal{M}^{1}_{t+1} =& R_{\phi}^{\pi}(s_t) + A_{\phi}(s_t, s_t')\bar{z}_t - (B_{\phi}(s_t)+ \eta I_{0})z_t - h(z_t, \bar{z}_t)
\end{split}
\end{equation*}

\begin{equation*}
\begin{split}
\bar{z}_{t+1} &= \bar{z}_{t} + \beta_t\big(g(z_t, \bar{z}_t) + \mathcal{M}^{2}_{t+1} + \epsilon(n)\big) 
\end{split}
\end{equation*}
Here,
\begin{equation*}
\begin{split}
g(z_t, \bar{z}_t) =& \lambda(\bar{z}_t) - \bar{z}_t \\
\mathcal{M}^{2}_{t+1} =& 0 \\
\lambda(\bar{z}_t) =& (\bar{B}_{\phi} + \eta I_{0})^{-1}(\bar{R}^{\pi}_{\phi} + \bar{A}_{\phi}\bar{z}_t) \\
\epsilon(n) =& z_{t+1} - \lambda(\bar{z}_t) 
\end{split}
\end{equation*}

$\lambda(\bar{z})$ is the unique globally asymptotically stable equilibrium point of the ODE $\dot{z} = h(z(t),\bar{z})$. $\lambda$ used here has no relation to usage of $\lambda$ in any other section of the paper. Using Lemma 1 of Chapter 6 of \citep{36}, we have  $\|z_{t+1} - \lambda(\bar{z}_t) \| \to 0$. Hence $\epsilon(n) = o(1)$. Therefore we can use the conclusion of \citep{35}. 

We will now satisfy condition A1 till condition A5 of \citep{35} to show the convergence of parameters:

\textbf{Condition A1}:
\begin{equation}\label{eq:a41}
\begin{split}
\|h(z_1, \bar{z}_1) - h(z_2, \bar{z}_2)\| &= \| \bar{A}_{\phi}(\bar{z}_1 - \bar{z}_2) - (\bar{B}_\phi+ \eta I_{0})(z_1 - z_2) \|  \\
&\leq \|\bar{A}_{\phi}\|\|\bar{z}_1 - \bar{z}_2\| + \|\bar{B}_\phi+ \eta I_{0}\|\|z_1 - z_2\| \\
&\leq \max(\|\bar{A}_{\phi}\|,\|\bar{B}_\phi+ \eta I_{0}\|)(\|\bar{z}_1 - \bar{z}_2\| + \|z_1 - z_2\|) \\
&= L_{h}(\|\bar{z}_1 - \bar{z}_2\| + \|z_1 - z_2\|)\;\; (L_{h} = \max(\|\bar{A}_{\phi}\|,\|\bar{B}_\phi+ \eta I_{0}\|))
\end{split}    
\end{equation}

Therefore, $h(z,\bar{z})$ is Lipchitz continuous with constant $L_h$.

\begin{equation}\label{eq:42}
\begin{split}
\|g(z_1, \bar{z}_1) - g(z_2, \bar{z}_2)\| &= \|((\bar{B}_\phi+ \eta I_{0})^{-1}\bar{A}_{\phi}-I)(\bar{z}_1 - \bar{z}_2)\| \\
&\leq \|((\bar{B}_\phi+ \eta I_{0})^{-1}\bar{A}_{\phi}-I)\|\|\bar{z}_1 - \bar{z}_2\| \\
&= L_{g}\|\bar{z}_1 - \bar{z}_2\| \;\;(L_{g} = \|((\bar{B}_\phi+ \eta I_{0})^{-1}\bar{A}_{\phi}-I)\| )
\end{split}
\end{equation}

Therefore, $g(z, \bar{z})$ is Lipchitz continuous with constant $L_g$.

Using \ref{eq:a41} and \ref{eq:42}, condition A1 is satisfied.

\textbf{Condition A2:}\\
Let us define an increasing sequence of $\sigma-$fields $\{\mathcal{F}_t\}$ as $\{z_m,\bar{z}_m,\mathcal{M}_{m}^{1},\mathcal{M}_{m}^{2}, m\leq t\}$.

\begin{equation*}
\begin{split}
E[\mathcal{M}^{1}_{t+1}|\mathcal{F}_t] &= E[R_{\phi}^{\pi}(s_t) + A_{\phi}(s_t, s_t')\bar{z}_t - (B_{\phi}(s_t)+ \eta I_{0})z_t - h(z_t, \bar{z}_t)|\mathcal{F}_t] \\
&= \int_{S} d^{\pi}(s_t)\big( R_{\phi}^{\pi}(s_t) + \int_{S}P^{\pi}(s_t'|s_t)A_{\phi}(s_t, s_t')\bar{z}_t\,ds_t' - (B_{\phi}(s_t)+ \eta I_{0})z_t\big)\,ds_t - h(z_t, \bar{z}_t) \\
&= 0
\end{split}
\end{equation*}

\begin{equation*}
E[\mathcal{M}^{2}_{t+1}|\mathcal{F}_t] = 0    
\end{equation*}

Hence, $\{\mathcal{M}^{1}_{t}\}$ and $\{\mathcal{M}^{2}_{t}\}$ are martingale difference sequence.
\begin{equation*}
\begin{split}
\|\mathcal{M}^{1}_{t+1}\|^{2} &= \| (R_{\phi}^{\pi}(s_t)-\bar{R}_{\phi}^{\pi}) + (A_{\phi}(s_t, s_t')-\bar{A}_{\phi})\bar{z}_t - (B_{\phi}(s_t)-\bar{B}_{\phi})z_t \|^{2} \\
&\leq 3(\|R_{\phi}^{\pi}(s_t)-\bar{R}_{\phi}^{\pi}\|^{2} + \|A_{\phi}(s_t, s_t')-\bar{A}_{\phi}\|^{2}\|\bar{z}_t\|^{2} + \|B_{\phi}(s_t)-\bar{B}_{\phi}\|^{2}\|z_t\|^{2}) \\
&\leq K_1(1 + \|z_t\|^{2} + \|\bar{z}_t\|^{2})
\end{split}
\end{equation*}

Here, $K_1 = 6\max(\|R_{\phi}^{\pi}(s_t)\|,\|(A_{\phi}(s_t, s_t')\|,\|B_{\phi}(s_t)\|)$ and $K_1$ is guaranteed to be finite from Assumption \ref{as:3} and \ref{as:4}.
We have, $E[\|\mathcal{M}^{1}_{t+1}\|^{2}||\mathcal{F}_t] \leq K_1(1 + \|z_t\|^{2} + \|\bar{z}_t\|^{2})$ and $E[\|\mathcal{M}^{2}_{t+1}\|^{2}||\mathcal{F}_t] \leq K_2(1 + \|z_t\|^{2} + \|\bar{z}_t\|^{2})$. $K_2$ can be any positive constant. Hence condition A2 is satisfied.\\

\textbf{Condition A3:}

We have, $\sum_t \alpha_t = \sum_t \frac{C_\alpha}{(1+t)^{\sigma}} = \infty$, $\sum_t \beta_t = \sum_t \frac{C_\beta}{(1+t)^{u}} = \infty$ and $\sum_t (\alpha_{t}^{2} + \beta_{t}^{2}) = \sum_t \big((\frac{C_\alpha}{(1+t)^{\sigma}})^{2} +(\frac{C_\beta}{(1+t)^{u}})^{2}\big) < \infty$. We can carefully set the value of $\sigma$ and $u$ to satisfy the conditions on step sizes. Further if $\sigma<u$ then $\beta_t = o(\alpha_t)$.\\

\textbf{Condition A4:}
\begin{equation*}
\begin{split}
h_{c}(z,\bar{z}) &:= \frac{h(cz,c\bar{z})}{c} \\
h_{c}(z,\bar{z}) & = \frac{\bar{R}_{\phi}^{\pi} + c\bar{A}_{\phi}\bar{z}_t - c(\bar{B}_{\phi}+ \eta I_{0})z_t}{c}\\
\lim_{c\to \infty} h_{c}(z,\bar{z}) & = \lim_{c\to \infty} \frac{\bar{R}_{\phi}^{\pi} + c\bar{A}_{\phi}\bar{z}_t - c(\bar{B}_{\phi}+ \eta I_{0})z_t}{c}\\
&= \bar{A}_{\phi}\bar{z}_t - (\bar{B}_{\phi}+ \eta I_{0})z_t
\end{split}
\end{equation*}

Let us define $h_{\infty}(z_t, \bar{z}_t):= \bar{A}_{\phi}\bar{z}_t - (\bar{B}_{\phi}+ \eta I_{0})z_t$. The ODE $\dot{z}(t):= h_{\infty}(z(t), \bar{z})$ has a unique globally asymptotically stable equilibrium point $\lambda_{\infty}(\bar{z}) = (\bar{B}_{\phi}+ \eta I_{0})^{-1}\bar{A}_{\phi}\bar{z}$ if $(\bar{B}_{\phi}+ \eta I_{0})$ is positive definite matrix. Let $C_{\phi} = \int_{S} d^{\pi}(s_{t})\phi^\pi(s_{t})\phi^\pi(s_{t})^{\intercal}\,ds_t$.

\begin{equation*}
\begin{split}
&\bar{B}_{\phi}+ \eta I_{0} = \begin{bmatrix} C_{\phi}+\eta I& \textbf{0} \\ \textbf{0}^{\intercal} & 1 \end{bmatrix}    \\
&\begin{bmatrix} w^{\intercal} & \rho \end{bmatrix}\begin{bmatrix} C_{\phi}+\eta I& \textbf{0} \\ \textbf{0}^{\intercal} & 1 \end{bmatrix} \begin{bmatrix} w \\ \rho \end{bmatrix} = w^{\intercal}(C_{\phi}+\eta I)w + \rho^2
\end{split}
\end{equation*}

If $\eta$ is strictly greater than negative of the minimum eigenvalue of $C_{\phi}$ then, 
\begin{equation}
\begin{split}
\forall \begin{bmatrix} w \\ p \end{bmatrix} \not = \begin{bmatrix} 0 \\ 0 \end{bmatrix} \quad \begin{bmatrix} w^{\intercal} & \rho \end{bmatrix}\begin{bmatrix} C_{\phi}+\eta I& \textbf{0} \\ \textbf{0}^{\intercal} & 1 \end{bmatrix} \begin{bmatrix} w \\ \rho \end{bmatrix} > 0 \\
\forall \begin{bmatrix} w \\ p \end{bmatrix} \not = \begin{bmatrix} 0 \\ 0 \end{bmatrix} \quad \begin{bmatrix} w^{\intercal} & \rho \end{bmatrix}\begin{bmatrix}\bar{B}_{\phi}+ \eta I_{0} \end{bmatrix} \begin{bmatrix} w \\ \rho \end{bmatrix} > 0 \\
\end{split}
\end{equation}

Hence, for $\eta+\lambda_{min}(C_{\phi}) > 0$, $\bar{B}_{\phi}+ \eta I_{0}$ is positive definite matrix. Further, $\lambda_{\infty}(z)$ is Lipchitz continuous. Therefore, the ODE $\dot{z}(t):= h_{\infty}(z(t), \bar{z})$ has a unique globally asymptotically stable equilibrium point $\lambda_{\infty}(\bar{z})$ and $\lambda_{\infty}(0) = 0$. Condition A4 is satisfied.\\

\textbf{Condition A5:}

\begin{equation}
\begin{split}
g_c(\bar{z}) &:= \frac{g(c\lambda_{\infty}(\bar{z}), c\bar{z})}{c} \\    
g_c(\bar{z}) &= \frac{(\bar{B}_{\phi} + \eta I_{0})^{-1}(\bar{R}^{\pi}_{\phi} + c\bar{A}_{\phi}\bar{z}) - c\bar{z}}{c} \\
\lim_{c \to \infty} g_c(\bar{z}) &= \lim_{c \to \infty} \frac{(\bar{B}_{\phi} + \eta I_{0})^{-1}(\bar{R}^{\pi}_{\phi} + c\bar{A}_{\phi}\bar{z}) - c\bar{z}}{c} \\
& = (\bar{B}_{\phi} + \eta I_{0})^{-1}\bar{A}_{\phi}\bar{z} - \bar{z}
\end{split}
\end{equation}

Let us define $g_{\infty}(\bar{z}):=((\bar{B}_{\phi} + \eta I_{0})^{-1}\bar{A}_{\phi}-I)\bar{z}$. The ODE $\dot{\bar{z}}(t) = g_{\infty}(\bar{z}(t))$ has origin as its unique globally asymptotically stable equilibrium if $I - (\bar{B}_{\phi} + \eta I_{0})^{-1}\bar{A}_{\phi}$ is positive definite matrix.

$\|\cdot\|$ refers to L2-norm. $\lambda_i$ are the eigenvalues of the matrix $C_{\phi}$. Let us assume the following:
\begin{equation}
\begin{split}
&\max(1,\max_i(\frac{1}{\lambda_i + \eta})) = \|(\bar{B}_{\phi} + \eta I_{0})^{-1}\| < \frac{1}{\|\bar{A}_{\phi}\|} \\
&\implies \|(\bar{B}_{\phi} + \eta I_{0})^{-1}\|\|\bar{A}_{\phi}\| < 1 \\
&\implies \|x\|\|(\bar{B}_{\phi} + \eta I_{0})^{-1}\|\|\bar{A}_{\phi}\|\|x\| < \|x\|^2 \\
&\implies \|x^{\intercal}(\bar{B}_{\phi} + \eta I_{0})^{-1}\bar{A}_{\phi}x\| < \|x\|^2 \\
&\implies x^{\intercal}(\bar{B}_{\phi} + \eta I_{0})^{-1}\bar{A}_{\phi}x < \|x\|^2 \\
&\implies x^{\intercal}(I - (\bar{B}_{\phi} + \eta I_{0})^{-1}\bar{A}_{\phi})x > 0  \\
\end{split}
\end{equation}

Hence, if $\max(1,\max_i(\dfrac{1}{\lambda_i + \eta})) < \dfrac{1}{\|\bar{A}_{\phi}\|}$, then $I - (\bar{B}_{\phi} + \eta I_{0})^{-1}\bar{A}_{\phi}$ is positive definite matrix. Therefore, the ODE $\dot{\bar{z}}(t) = g_{\infty}(\bar{z}(t))$ has origin as its unique globally asymptotically stable. Condition A5 is satisfied.

Let us the consider the ODE $\dot{z}(t) = h(z(t),\bar{z})$. Here, $h(z(t),\bar{z}) = \bar{R}_{\phi}^{\pi} + \bar{A}_{\phi}\bar{z} - (\bar{B}_{\phi}+ \eta I_{0})z(t)$  As earlier, for $\eta+\lambda_{min}(C_{\phi}) > 0$, $\bar{B}_{\phi}+ \eta I_{0}$ is positive definite matrix. Therefore, the ODE $\dot{z}(t):= h(z(t), \bar{z})$ has a unique globally asymptotically stable equilibrium point $\lambda(\bar{z}) = (\bar{B}_{\phi} + \eta I_{0})^{-1}(\bar{R}^{\pi}_{\phi} + \bar{A}_{\phi}\bar{z})$.

Now, consider the ODE $\dot{\bar{z}}(t) = g(\lambda(\bar{z}(t)),\bar{z}(t))$. Here, $g(\lambda(\bar{z}(t)),\bar{z}(t)) = \lambda(\bar{z}(t)) - \bar{z}(t) = (\bar{B}_{\phi} + \eta I_{0})^{-1}\bar{R}^{\pi}_{\phi} - (I - (\bar{B}_{\phi} + \eta I_{0})^{-1}\bar{A}_{\phi})\bar{z}(t)$. For $\max(1,\max_i(\dfrac{1}{\lambda(C_\phi)_i + \eta})) < \dfrac{1}{\|\bar{A}_{\phi}\|}$, $I - (\bar{B}_{\phi} + \eta I_{0})^{-1}\bar{A}_{\phi}$ is positive definite matrix. Therefor the ODE $\dot{\bar{z}}(t) = g(\lambda(\bar{z}(t)),\bar{z}(t))$ has a unique globally asymptotically stable equilibrium point $(\bar{B}_{\phi} + \eta I_{0} - \bar{A}_{\phi})^{-1}\bar{R}^{\pi}_{\phi}$. 

Since condition A1-A5 are satisfied, using the conclusion of \citep{35}, $z_t(=[w_t \;\rho_t]^{\intercal})$ and $\bar{z}_t(=[\bar{w}_t \; \bar{\rho}_t]^{\intercal})$ converges to $(\bar{B}_{\phi} + \eta I_{0} - \bar{A}_{\phi})^{-1}\bar{R}^{\pi}_{\phi} (=[w(\theta)^* \;\rho(\theta)^*]^{\intercal} = [\bar{w}(\theta)^* \;\bar{\rho}(\theta)^*]^{\intercal})$.
\end{proof}

\begin{lemma}\label{lm:a20}
For policy $\pi$ parameterized by $\theta$ and optimal differential Q-value function parameter $w^*(=w(\theta_t)^*)$ according to Theorem \ref{th:a5}, the following condition holds true:
\begin{equation*}
\mathbb{E}[\nabla_a Q_{diff}^{w^*}(s,a)|_{a=\pi(s)}\nabla_{\theta}\pi(s)|\theta] = \nabla_{\theta}\rho(\pi) + e^{\pi}
\end{equation*}
Here, $e^{\pi}$ denotes the error in gradient due to function approximation.
\begin{equation*}
e^{\pi} = \int_{S}d^{\pi}(s)\big((\nabla_a Q_{diff}^{w^*}(s,a)-\nabla_a Q_{diff}^{\pi}(s,a))|_{a=\pi(s)}\big)\nabla_{\theta}\pi(s)\;ds
\end{equation*}
\end{lemma}
\begin{proof}
\begin{equation*}
\begin{split}
\mathbb{E}[\nabla_a Q_{diff}^{w}(s,a)|_{a=\pi(s)}\nabla_{\theta}\pi(s)|\theta] & = \int_{S}d^{\pi}(s)\nabla_a Q_{diff}^{w}(s,a)\nabla_{\theta}\big)\pi(s)\;ds \\
 =& \int_{S}d^{\pi}(s)\big((\nabla_a Q_{diff}^{w}(s,a)-\nabla_a Q_{diff}^{\pi}(s,a))|_{a=\pi(s)}\big)\nabla_{\theta}\pi(s)\;ds \\
&+\int_{S}d^{\pi}(s)\nabla_a Q_{diff}^{\pi}(s,a)|_{a=\pi(s)}\nabla_{\theta}\pi(s)\;ds \\
=& \nabla_{\theta}\rho(\pi) + e^{\pi} \quad(\text{Using Theorem \ref{th:1}})
\end{split}
\end{equation*}
\end{proof}

We will now prove the convergence of policy parameter $\theta_t (\in \mathbb{R}^d)$ using the following update rule (M = 1):
\begin{equation*}
\begin{split}
\theta_{t+1} =& \Gamma_{C_\theta}\Big(\theta_t + \gamma_t\nabla_{a}Q_{diff}^{w}(s_{t},a)|_{a = \pi(s_{t})}\nabla_\theta\pi(s_{t})\Big) \\
\end{split}
\end{equation*}
Here, $s_t$ is the state sampled from the buffer at time step $t$. $\Gamma_{C_\theta}: \mathbb{R}^d \to C_\theta$ is a projection operator, where $C_\theta$ is compact convex set.

\begin{theorem}\label{th:a6}
$\Gamma_{C_\theta}: \mathbb{R}^d \to C_\theta$ is a projection operator, where $C_\theta$ is compact convex set and $\hat{\Gamma}_{C_\theta}(\theta)\nabla_{\theta}\rho(\theta)$ refers to directional derivative of $\Gamma_{C_\theta}(\cdot)$ in the direction $\nabla_{\theta}\rho(\theta)$ at $\theta$. Let $K = \{\theta \in C_\theta | \hat{\Gamma}_{C_\theta}(\theta)\nabla_{\theta}\rho(\theta) = 0 \}$ and $K^{\epsilon} = \{\theta' \in C_\theta | \exists\; \theta \in K \; \|\theta'-\theta\|<\epsilon \}$. $\forall \epsilon > 0 \;\exists \delta$ such that if $\sup_{\pi}\|e^{\pi}\| < \delta$ then $\theta_t$ converges to $K^{\epsilon}$ as $t \to \infty$ with probability one. $e^{\pi}$ is the function approximation error defined in Lemma \ref{lm:a20}.
\end{theorem}
\begin{proof}
Let the $\sigma$-field $\mathcal{F}^{2}_{t}$ for actor iterate be defined as $\sigma(s_m, m<t;\theta_n, n\leq t)$. We will use the Theorem 5.3.1 from Chapter 5 of \citet{38} to prove convergence.
\begin{equation}\label{eq:a52}
\begin{split}
\theta_{t+1} =& \Gamma_{C_\theta}\Big(\theta_t + \gamma_t\nabla_{a}Q_{diff}^{w}(s_{t},a)|_{a = \pi(s_{t})}\nabla_\theta\pi(s_{t})\Big) \\
\theta_{t+1} =& \Gamma_{C_\theta}\Big(\theta_t + \gamma_t\big( h_2(\theta_t) + \mathcal{N}_{t+1} + \mathcal{M}_{t+1}^{3}\big)\Big) \\
\end{split}
\end{equation}

Here,
\begin{equation*}
\begin{split}
h_2(\theta_t) =& \mathbb{E}[\nabla_a Q_{diff}^{w^*}(s_t,a)|_{a=\pi(s_t)}\nabla_{\theta}\pi(s_t)|\mathcal{F}^{2}_{t}] \\
\mathcal{N}_{t+1} =& \mathbb{E}[\nabla_a Q_{diff}^{w}(s_t,a)|_{a=\pi(s_t)}\nabla_{\theta}\pi(s_t)|\mathcal{F}^{2}_{t}] -\mathbb{E}[\nabla_a Q_{diff}^{w^*}(s_t,a)|_{a=\pi(s_t)}\nabla_{\theta}\pi(s_t)|\mathcal{F}^{2}_{t}] \\
\mathcal{M}^{3}_{t+1} =& \nabla_a Q_{diff}^{w}(s_t,a)|_{a=\pi(s_t)}\nabla_{\theta}\pi(s_t) - \mathbb{E}[\nabla_a Q_{diff}^{w}(s_t,a)|_{a=\pi(s_t)}\nabla_{\theta}\pi(s_t)|\mathcal{F}^{2}_{t}]\\
\end{split}
\end{equation*}

\textbf{Condition B1}:

We have, $\sum_t \alpha_t = \sum_t \frac{C_\alpha}{(1+t)^{\sigma}} = \infty$, $\sum_t \beta_t = \sum_t \frac{C_\beta}{(1+t)^{u}} = \infty$, $\sum_t \gamma_t = \sum_t \frac{C_\gamma}{(1+t)^v} = \infty $ and $\sum_t (\alpha_{t}^{2} + \beta_{t}^{2} + \gamma_t^{2}) = \sum_t \Big(\big(\frac{C_\alpha}{(1+t)^{\sigma}}\big)^{2} +\big(\frac{C_\beta}{(1+t)^{u}}\big)^{2} + \big(\frac{C_\gamma}{(1+t)^{v}}\big)^2\Big) < \infty$. We can carefully set the value of $\sigma$, $u$, and $v$ to satisfy the conditions on step sizes. Further if $\sigma<u<v$ then $\beta_t = o(\alpha_t)$ and $\gamma_t = o(\beta_t)$.\\

\textbf{Condition B2}:
We will now prove that $h_2(\theta)$ is Lipchitz continuous in $\theta$.\\ 

\begin{equation}\label{eq:a56}
\begin{split}
\nabla_{\theta}h_2(\theta) =& \nabla_{\theta}\int_{S}d^{\pi}(s)(w(\theta)^*)^{\intercal}\nabla_a \phi(s,a)|_{a=\pi(s)}\nabla_{\theta}\pi(s)\;ds \\
=&\int_{S}\nabla_{\theta}d^{\pi}(s)(w(\theta)^*)^{\intercal}\nabla_a \phi(s,a)|_{a=\pi(s)}\nabla_{\theta}\pi(s)\;ds \term{1}\\
&+\int_{S}d^{\pi}(s)(\nabla_{\theta}w(\theta)^*)^{\intercal}\nabla_a \phi(s,a)|_{a=\pi(s)}\nabla_{\theta}\pi(s)\;ds \term{2}\\
&+\int_{S}d^{\pi}(s)(w(\theta)^*)^{\intercal}(\nabla_{\theta}\nabla_a \phi(s,a)|_{a=\pi(s)})\nabla_{\theta}\pi(s)\;ds \term{3}\\
&+\int_{S}d^{\pi}(s)(w(\theta)^*)^{\intercal}\nabla_a \phi(s,a)|_{a=\pi(s)}\nabla_{\theta}^2\pi(s)\;ds \term{4}
\end{split}
\end{equation}

Using Assumption \ref{as:6}, $\pi(s,\theta)$ is Lipchitz continuous in $\theta$ and hence $\nabla_\theta\pi(s)$ is bounded. By Assumption \ref{as:17}, $\phi(s,a)$ is Lipchitz continuous in $a$ and therefore $\nabla_{a}\phi(s,a)$ is bounded. $\nabla_{\theta}d^{\pi}(s)$ is bounded by application of Theorem 2.1 of \citet{37}. Further, $\nabla_{\theta}w(\theta)^*$ is bounded because $w(\theta)^*$ is Lipchitz continuous in $\theta$\;(Lemma \ref{lm:a6}). By Assumption \ref{as:16}, $\nabla_{\theta}^2\pi(s)$ exists and is bounded because $\theta \in C_\theta$. Further, $\nabla_{\theta}\nabla_a \phi(s,a)|_{a=\pi(s)}$ is bounded by application of Assumption \ref{as:19} with Assumption \ref{as:6}. All the terms in \eqref{eq:a56} are bounded. Consequently, $\nabla_{\theta}h_2(\theta)$ is bounded and Lipchtiz continuous in $\theta$.

\textbf{Condition B3}: Now, we will prove the noise terms $\mathcal{N}_{t+1}$ and $\mathcal{M}_{t+1}^{3}$ converges asymptotically. $\mathcal{N}_{t+1}$ is $o(1)$ term because $w_t$ converges $w(\theta_t)^*$ according to Theorem \ref{th:a5}. Further,
\begin{equation*}
\begin{split}  
\xi_{T} =& \sum_{t=0}^{T-1}\gamma_t\mathcal{M}^{3}_{t+1} \\
=&\sum_{t=0}^{T-1}\gamma_t\Big(\nabla_a Q_{diff}^{w}(s_t,a)|_{a=\pi(s_t)}\nabla_{\theta}\pi(s_t) - \mathbb{E}[\nabla_a Q_{diff}^{w}(s_t,a)|_{a=\pi(s_t)}\nabla_{\theta}\pi(s_t)|\mathcal{F}^{2}_{t}]\Big)
\end{split}
\end{equation*}

We will now prove that ${\xi_t}$ is a martingale process.

\begin{equation}\label{eq:a58}
\begin{split}
&\mathbb{E}\big[\mathcal{M}^{3}_{t+1}|\mathcal{F}_t^{2}\big]\\
&=\mathbb{E}\big[\big(\nabla_a Q_{diff}^{w}(s_t,a)|_{a=\pi(s_t)}\nabla_{\theta}\pi(s_t) - \mathbb{E}[\nabla_a Q_{diff}^{w}(s_t,a)|_{a=\pi(s_t)}\nabla_{\theta}\pi(s_t)|\mathcal{F}^{2}_{t}]\big)|\mathcal{F}_t^{2}\big]\\
&=\mathbb{E}\big[\big(\nabla_a Q_{diff}^{w}(s_t,a)|_{a=\pi(s_t)}\nabla_{\theta}\pi(s_t)\big)|\mathcal{F}_t^{2}\big] - \mathbb{E}[\nabla_a Q_{diff}^{w}(s_t,a)|_{a=\pi(s_t)}\nabla_{\theta}\pi(s_t)|\mathcal{F}^{2}_{t}]\\
&= 0
\end{split}
\end{equation}

\begin{equation}\label{eq:a59}
\begin{split}
\mathbb{E}\big[\xi_T|\mathcal{F}^{2}_{T-1}\big] =& \mathbb{E}\big[\sum_{t=0}^{T-1}\gamma_t\mathcal{M}^{3}_{t+1}|\mathcal{F}^{2}_{T-1}\big] \\
=&\sum_{t=0}^{T-2}\gamma_t\mathcal{M}^{3}_{t+1} + \mathbb{E}\big[\mathcal{M}^{3}_{T}|\mathcal{F}_{T-1}^{2}\big]\\
=& \xi_{T-1} \quad (\text{Using \ref{eq:a58}})
\end{split}
\end{equation}

\begin{equation}\label{eq:a60}
\begin{split}
\mathbb{E}\big[\|\xi_{T}\|^2\big] =& \mathbb{E}\bigg[\Big(\sum_{n=0}^{T-1}\gamma_n \mathcal{M}^{3}_{n+1}\Big)^{\intercal}\Big(\sum_{m=0}^{T-1}\gamma_m\mathcal{M}^{3}_{m+1}\Big)\bigg] \\
=&\mathbb{E}\bigg[\sum_{n=0}^{T-1}\Big\|\gamma_n \mathcal{M}^{3}_{n+1}\Big\|^2\bigg] \\
\Big(\because \text{For }n>m \;&\mathbb{E}\Big[(\mathcal{M}^{3}_{n})^{\intercal}\mathcal{M}^{3}_{m}\Big]=\mathbb{E}\Big[(\mathcal{M}^{3}_{n})^{\intercal}E[\mathcal{M}^{3}_{m}|\mathcal{F}^{2}_{m-1}]\Big]=0\Big) \\
&\leq \bigg(\sum_{n=0}^{T-1}\gamma_n^{2}\bigg)\sup_n\mathbb{E}\bigg[\Big\| \mathcal{M}^{3}_{n+1}\Big\|^2\bigg] <\infty
\end{split}
\end{equation}

We have $\sum_{n}\gamma_n^{2}< \infty$ from condition B1. From Assumption \ref{as:6} and \ref{as:8} it can be proved that $\|\mathcal{M}^{3}_t\|$ is bounded. Therefore $\mathbb{E}\big[\|\xi_{T}\|^2\big]< \infty$. Using \eqref{eq:a59} and \eqref{eq:a60} we have $\xi_t$ is martingale process. 

Now,

\begin{equation}\label{eq:a61}
\begin{split}
\sum_{t}\mathbb{E}\bigg[\|\xi_{t+1}-\xi_{t}\|^2\Big|\mathcal{F}^{2}_t\bigg] &=\sum_{t}\mathbb{E}\bigg[\gamma_t^2\|\mathcal{M}^{3}_{t+1}\|^2\Big|\mathcal{F}^{2}_t\bigg] \\
&\leq \bigg(\sum_{t=0}\gamma_t^{2}\bigg)\sup_n\mathbb{E}\bigg[\Big\|\mathcal{M}^{3}_{n+1}\Big\|^2\Big|\mathcal{F}^{2}_n\bigg] <\infty
\end{split}
\end{equation}

By martingale convergence theorem of Chapter 11 of \citet{36} and using \ref{eq:a60} and \ref{eq:a61} it can proved that martingale $\xi_t$ converges and $\sum_{n=t}^{\infty}\gamma_n\mathcal{M}^{3}_{n+1} \to 0$ as $t \to \infty$. 

Hence the noise terms $\mathcal{N}_{t+1}$ and $\mathcal{M}_{t+1}^{3}$ converge asymptotically. 

\textbf{Condition B4}: $\|\theta_t\|$ is bounded because of projection operator $\Gamma_{C_\theta}$.

Using Theorem 5.3.1 of \citet{38} with the satisfaction of condition B1-B4 ensures that \ref{eq:a52} tracks the ODE given in \ref{ode} and  $\theta_t$ converges to $K^{\epsilon}$ as $t \to \infty$ where $\hat{\Gamma}_{C_\theta}(y)(x) = \lim_{\delta \to \infty}\dfrac{\Gamma_{C_\theta}(x+\delta y) - \Gamma_{C_\theta}(x)}{\delta}$.

\begin{equation}\label{ode}
\begin{split}
\dot{\theta}(t) = \hat{\Gamma}_{C_\theta}(\theta(t))h_2(\theta(t)) = \hat{\Gamma}_{C_\theta}(\theta(t))(\nabla_{\theta}\rho(\theta(t)) + e^{\pi(t)}) \quad(\text{Using Lemma \ref{lm:a20}})
\end{split}
\end{equation}

Further, as $\sup_{\pi}\|e^{\pi}\| \to 0$, \ref{eq:a52} tracks the ODE given in \ref{ode2} and  $\theta_t$ converges to $K$ as $t \to \infty$.

\begin{equation}\label{ode2}
\begin{split}
\dot{\theta}(t) = \hat{\Gamma}_{C_\theta}(\theta(t))(\nabla_{\theta}\rho(\theta(t))) \quad(\text{Using Lemma \ref{lm:a20}})
\end{split}
\end{equation}
\end{proof}

\section{Algorithm and Hyperparameters}

\subsection{(Off-Policy) ARO-DDPG Practical Algorithm}

\begin{algorithm}[H]
\caption{(Off-Policy) ARO-DDPG Practical Algorithm}
\label{alg:1}
\text{Initialize actor parameter $\theta$ and differential Q-value function parameters $w_{1},w_{2}$. Initialize actor target parameter $\theta \to \overline{\theta}$}
\text{Initialize differential Q-value function target parameters $ w_{1} \to \overline{w_{1}}, w_{2} \to \overline{w_{2}}$. Initialize average reward parameter $\rho$.}
\text{Initialize target average reward parameter $\rho \to \overline{\rho}$. Initialize Replay buffer = \{\}}
\begin{algorithmic}[1] 
\STATE $t=0$, $s_{0}$ = env.reset()
\WHILE{$t \leq$ total steps}
\STATE $a_t = \pi(s_{t}) + \epsilon$ \algorithmiccomment{$\epsilon$ denotes the noise}
\STATE $s_{t+1} \sim P(\cdot | s_t, a_t)$ and $r_t = R(s_{t},a_{t})$
\STATE Store $\{s_{t},a_{t},s_{t+1}\}$ in the Replay Buffer
\IF { $t\;\%\;eval\_freq == 0$}
\STATE Evaluate(agent)
\ENDIF
\IF { $t\;\%\;critic\_update\_freq == 0$}
\STATE Update critic according to (\ref{eq:23}) - (\ref{eq:26})
\ENDIF
\IF { $t\;\%\;actor\_update\_freq == 0$}
\STATE Update actor according to (\ref{eq:27}) - (\ref{eq:28})
\STATE Update target estimators according to (\ref{eq:29}) - (\ref{eq:31})
\ENDIF
\IF { $s_{t+1}$ is terminal}
\STATE $s_{t} = $ env.reset()
\ELSE
\STATE $s_{t} = s_{t+1}$ 
\ENDIF
\ENDWHILE
\end{algorithmic}
\end{algorithm}

\subsection{Finite time analysis algorithm}
Here we present the algorithm with linear function approximator for which finite time analysis was done. $\mathcal{B}_t$ denotes the batch of tuple of the form $\{s_{i}, a_{i}, s_{i}'\}$ sampled from the buffer at timestep $t$. $\Gamma_{C_w}$ is a projection operator defined as  $\Gamma_{C_w}: \mathbb{R}^k \to B$, where $B(\subset\mathbb{R}^k)$ is a compact convex set. Here, the differential Q-value function parameter $w \in \mathbb{R}^k$.

\begin{algorithm}[H]
\caption{ On-policy AR-DPG with Linear FA}
\label{alg:2}
\text{Initialize actor parameter $\theta$ and differential Q-value function parameters $w$. Initialize actor target parameter $\theta \to \overline{\theta}$.}\\
\text{Initialize differential Q-value function target parameters $ w \to \overline{w}$.Initialize average reward parameter $\rho$} \\ 
\text{Initialize target average reward parameter $\rho \to \overline{\rho} $}\\
\text{Initialize buffer = \{\}}
\begin{algorithmic}[1] 
\STATE $t=0$, $s_{0}$ = env.reset()
\WHILE{$t \leq$ total steps}
\STATE $a_t = \pi(s_{t}) + \epsilon$ \algorithmiccomment{$\epsilon$ is the noise}
\STATE $s_{t+1} \sim P(\cdot | s_t, a_t)$ and $r_t = R(s_{t},a_{t})$
\STATE Store $\{s_{t},a_{t},s_{t+1}\}$ in the Buffer
\IF { $t \;\%\; critic\_update\_freq == 0$}
\STATE Sample $\mathcal{B}_t = \{s_{i}, a_{i}, s_{i}'\}_{i=0}^{M-1}$ from the Replay Buffer 
\STATE $w_{t+1} =\Gamma_{C_w}\Big( w_t + \dfrac{\alpha_t}{M}\sum_{i=0}^{M-1}\Bigl(R^\pi(s_{i}) - \bar{\rho_t} + \phi^\pi(s_{i}')^{\intercal}\bar{w_t} - \phi^\pi(s_{i})^{\intercal}w_t\Bigr)\phi^\pi(s_{i}) - \alpha_t\eta w_t\Big)$
\STATE $\rho_{t+1} = \rho_t+\dfrac{\alpha_t}{M}\sum_{i=0}^{M-1}\Big(R^\pi(s_{i})-\rho_t+\phi^\pi(s_{i}')^{\intercal}\bar{w_t}-\phi^\pi(s_{i})^\intercal\bar{w_t}\Big)$
\STATE $\overline{w}_{t+1} = \overline{w}_{t} + \beta_t(w_{t+1} - \overline{w}_{t+1})$
\STATE $\overline{\rho}_{t+1} = \overline{\rho}_{t} + \beta_t(\rho_{t+1} - \overline{\rho}_{t+1})$
\STATE $\theta_{t+1} = \theta_t + \dfrac{\gamma_t}{M}\sum_{i=0}^{M-1}\nabla_{a}Q_{diff}^{w}(s_{i},a)|_{a = \pi(s_{i})}\nabla_\theta\pi(s_{i})$
\STATE buffer = \{\}
\ENDIF
\IF { $s_{t+1}$ is terminal}
\STATE $s_{t} = $ env.reset()
\ELSE
\STATE $s_{t} = s_{t+1}$ 
\ENDIF
\ENDWHILE
\end{algorithmic}
\end{algorithm}

\begin{algorithm}[H]
\caption{ Off-policy AR-DPG with Linear FA}
\label{alg:3}
\text{Initialize actor parameter $\theta$ and differential Q-value function parameters $w$. Initialize actor target parameter $\theta \to \overline{\theta}$ } \\ 
\text{Initialize differential Q-value function target parameters $ w \to \overline{w}$. Initialize average reward parameter $\rho$ }\\
\text{Initialize target average reward parameter $\rho \to \overline{\rho}$. $\mu$ is the behavior policy}\\
\text{Initialize Replay buffer = \{\}}
\begin{algorithmic}[1] 
\STATE $t=0$, $s_{0}$ = env.reset()
\WHILE{$t \leq$ total steps}
\STATE $a_t = \mu(s_{t}) + \epsilon$ \algorithmiccomment{$\epsilon$ is the noise}
\STATE $s_{t+1} \sim P(\cdot | s_t, a_t)$ and $r_t = R(s_{t},a_{t})$
\STATE Store $\{s_{t},a_{t},s_{t+1}\}$ in the Replay Buffer
\STATE Sample $\mathbb{B}_t = \{s_{i}, a_{i}, s_{i}'\}_{i=0}^{M-1}$ from the Replay Buffer \STATE $w_{t+1} = \Gamma_{C_w}\Big(w_t + \dfrac{\alpha_t}{M}\sum_{i=0}^{M-1}\Bigl(R^\mu(s_{i}) - \bar{\rho_t} + \phi^\pi(s_{i}')^{\intercal}\bar{w_t} - \phi^\pi(s_{i})^{\intercal}w_t\Bigr)\phi^\pi(s_{i}) - \alpha_t\eta w_t\Big)$
\STATE $\rho_{t+1} = \rho_t+\dfrac{\alpha_t}{M}\sum_{i=0}^{M-1}\Big(R^\mu(s_{i})-\rho_t+\phi^\pi(s_{i}')^{\intercal}\bar{w_t}-\phi^\pi(s_{i})^\intercal\bar{w_t}\Big)$
\STATE $\overline{w}_{t+1} = \overline{w}_{t} + \beta_t(w_{t+1} - \overline{w}_{t+1})$
\STATE $\overline{\rho}_{t+1} = \overline{\rho}_{t} + \beta_t(\rho_{t+1} - \overline{\rho}_{t+1})$
\STATE $\theta_{t+1} = \theta_t + \dfrac{\gamma_t}{M}\sum_{i=0}^{M-1}\nabla_{a}Q_{diff}^{w}(s_{i},a)|_{a = \pi(s_{i})}\nabla_\theta\pi(s_{i})$
\IF { $s_{t+1}$ is terminal}
\STATE $s_{t} = $ env.reset()
\ELSE
\STATE $s_{t} = s_{t+1}$ 
\ENDIF
\ENDWHILE
\end{algorithmic}
\end{algorithm}

\subsection{Asymptotic analysis algorithm}
Here we present the algorithm with linear function approximator for which asymptotic analysis was done. $\mathcal{B}_t$ denotes the batch of tuple of the form $\{s_{i}, a_{i}, s_{i}'\}$ sampled from the buffer at timestep $t$. $\Gamma_{C_\theta}$ is a projection operator defined as  $\Gamma_{C_\theta}: \mathbb{R}^d \to C_\theta$, where $C_\theta(\subset\mathbb{R}^d)$ is a compact convex set. Here, the actor parameter $\theta \in \mathbb{R}^d$.

\begin{algorithm}[H]
\caption{ On-policy AR-DPG with Linear FA}
\label{alg:4}
\text{Initialize actor parameter $\theta$ and differential Q-value function parameters $w$. Initialize actor target parameter $\theta \to \overline{\theta}$.}\\
\text{Initialize differential Q-value function target parameters $ w \to \overline{w}$.Initialize average reward parameter $\rho$} \\ 
\text{Initialize target average reward parameter $\rho \to \overline{\rho} $}\\
\text{Initialize buffer = \{\}}
\begin{algorithmic}[1] 
\STATE $t=0$, $s_{0}$ = env.reset()
\WHILE{$t \leq$ total steps}
\STATE $a_t = \pi(s_{t}) + \epsilon$ \algorithmiccomment{$\epsilon$ is the noise}
\STATE $s_{t+1} \sim P(\cdot | s_t, a_t)$ and $r_t = R(s_{t},a_{t})$
\STATE Store $\{s_{t},a_{t},s_{t+1}\}$ in the Buffer
\IF { $t \;\%\; critic\_update\_freq == 0$}
\STATE Sample $\mathcal{B}_t = \{s_{i}, a_{i}, s_{i}'\}_{i=0}^{M-1}$ from the Replay Buffer 
\STATE $w_{t+1} =w_t + \dfrac{\alpha_t}{M}\sum_{i=0}^{M-1}\Bigl(R^\pi(s_{i}) - \bar{\rho_t} + \phi^\pi(s_{i}')^{\intercal}\bar{w_t} - \phi^\pi(s_{i})^{\intercal}w_t\Bigr)\phi^\pi(s_{i}) - \alpha_t\eta w_t$
\STATE $\rho_{t+1} = \rho_t+\dfrac{\alpha_t}{M}\sum_{i=0}^{M-1}\Big(R^\pi(s_{i})-\rho_t+\phi^\pi(s_{i}')^{\intercal}\bar{w_t}-\phi^\pi(s_{i})^\intercal\bar{w_t}\Big)$
\STATE $\overline{w}_{t+1} = \overline{w}_{t} + \beta_t(w_{t+1} - \overline{w}_{t+1})$
\STATE $\overline{\rho}_{t+1} = \overline{\rho}_{t} + \beta_t(\rho_{t+1} - \overline{\rho}_{t+1})$
\STATE $\theta_{t+1} = \Gamma_{C_\theta}\Big(\theta_t + \dfrac{\gamma_t}{M}\sum_{i=0}^{M-1}\nabla_{a}Q_{diff}^{w}(s_{i},a)|_{a = \pi(s_{i})}\nabla_\theta\pi(s_{i})\Big)$
\STATE buffer = \{\}
\ENDIF
\IF { $s_{t+1}$ is terminal}
\STATE $s_{t} = $ env.reset()
\ELSE
\STATE $s_{t} = s_{t+1}$ 
\ENDIF
\ENDWHILE
\end{algorithmic}
\end{algorithm}

\begin{algorithm}[H]
\caption{ Off-policy AR-DPG with Linear FA}
\label{alg:5}
\text{Initialize actor parameter $\theta$ and differential Q-value function parameters $w$ } \\ 
\text{Initialize actor target parameter $\theta \to \overline{\theta}$ and } \\
\text{Initialize differential Q-value function target parameters $ w \to \overline{w}$}\\
\text{Initialize average reward parameter $\rho$ and}\\
\text{Initialize target average reward parameter $\rho \to \overline{\rho} $}\\
\text{$\mu$ is the behavior policy}\\
\text{Initialize Replay buffer = \{\}}
\begin{algorithmic}[1] 
\STATE $t=0$, $s_{0}$ = env.reset()
\WHILE{$t \leq$ total steps}
\STATE $a_t = \mu(s_{t}) + \epsilon$ \algorithmiccomment{$\epsilon$ is the noise}
\STATE $s_{t+1} \sim P(\cdot | s_t, a_t)$ and $r_t = R(s_{t},a_{t})$
\STATE Store $\{s_{t},a_{t},s_{t+1}\}$ in the Replay Buffer
\STATE Sample $\mathbb{B}_t = \{s_{i}, a_{i}, s_{i}'\}_{i=0}^{M-1}$ from the Replay Buffer \STATE $w_{t+1} = w_t + \dfrac{\alpha_t}{M}\sum_{i=0}^{M-1}\Bigl(R^\mu(s_{i}) - \bar{\rho_t} + \phi^\pi(s_{i}')^{\intercal}\bar{w_t} - \phi^\pi(s_{i})^{\intercal}w_t\Bigr)\phi^\pi(s_{i}) - \alpha_t\eta w_t$
\STATE $\rho_{t+1} = \rho_t+\dfrac{\alpha_t}{M}\sum_{i=0}^{M-1}\Big(R^\mu(s_{i})-\rho_t+\phi^\pi(s_{i}')^{\intercal}\bar{w_t}-\phi^\pi(s_{i})^\intercal\bar{w_t}\Big)$
\STATE $\overline{w}_{t+1} = \overline{w}_{t} + \beta_t(w_{t+1} - \overline{w}_{t+1})$
\STATE $\overline{\rho}_{t+1} = \overline{\rho}_{t} + \beta_t(\rho_{t+1} - \overline{\rho}_{t+1})$
\STATE $\theta_{t+1} =\Gamma_{C_\theta}\Big(\theta_t + \dfrac{\gamma_t}{M}\sum_{i=0}^{M-1}\nabla_{a}Q_{diff}^{w}(s_{i},a)|_{a = \pi(s_{i})}\nabla_\theta\pi(s_{i})\Big)$
\IF { $s_{t+1}$ is terminal}
\STATE $s_{t} = $ env.reset()
\ELSE
\STATE $s_{t} = s_{t+1}$ 
\ENDIF
\ENDWHILE
\end{algorithmic}
\end{algorithm}

\subsection{Hyperparameters}
The hyper-parameters mentioned in this section produces good performance for all the environment save for "fish-upright" where we used GeLU activation function.

\begin{center}
\begin{tabular}{ c|c } 
 \hline
 \bf{Hyperparameter} & \bf{Value}  \\ 
 \hline
 Buffer Size & 1e6  \\ 
 Total Environment Steps & 1e6  \\ 
 Batch size & 256\\
 Evaluation Frequency & 5000 \\
 Training Episode Length & 1000 \\
 Evaluation Episode Length & 10000 \\
 Activation Function & ReLU \\
 Learning rate Actor & 3e-4 \\
 Learning rate Differential Q-value function & 3e-4 \\
 Learning rate Average reward parameter & 3e-4 \\
 No. of Hidden Layers & 2\\
 No. of Nodes in Hidden Layer & 128\\
 Update frequency & 10 steps\\
 No. of Critic updates & 10 \\
 No. of Actor updates & 5 \\
 Polyak averaging constant & 0.995\\
 \hline
\end{tabular}
\end{center}
\end{document}